%% file: cvpr.tex
\makeatletter
\@namedef{ver@everyshi.sty}{}
\makeatother

\documentclass[final]{cvpr}

\usepackage{times}
\usepackage{epsfig}
\usepackage{graphicx}
\usepackage{amsmath}
\usepackage{amssymb}
\usepackage[table]{xcolor}
\usepackage{booktabs}
\usepackage{makecell}
\usepackage{array}
\usepackage{overpic}

\usepackage{nopageno} 

\usepackage{pgfplots} 
\pgfplotsset{compat=1.7} 
\usepackage{multirow}
\usepackage{pdflscape} 
\usepackage{titling} 
\predate{}
\postdate{} 
\definecolor{orange_our_bright}{HTML}{FF7F0E}
\definecolor{orange_our}{HTML}{ae5a21}
\definecolor{blue_our}{HTML}{4060c4}
\definecolor{green_our}{HTML}{507e50}

\newcommand{\ra}[1]{\renewcommand{\arraystretch}{#1}}

\usepackage[pagebackref=true,breaklinks=true,colorlinks,bookmarks=false]{hyperref}

\def\cvprPaperID{8649} 
\def\confYear{CVPR 2021}

\begin{document}

\title{Fostering Generalization in Single-view 3D Reconstruction by Learning a Hierarchy of Local and Global Shape Priors}
\date{} 

\author{Jan Bechtold$^{12}$\\
\and
Maxim Tatarchenko$^1$\\
\and
Volker Fischer$^1$\\
\and
Thomas Brox$^2$\\
\and 
{$^1$Bosch Center for Artificial Intelligence}
\and
{$^2$University of Freiburg}
}

\maketitle

\input{01_abstract}

\input{02_introduction}

\input{03_related_work}

\input{04_method}

\input{05_experiments}

\input{06_conclusion}


\input{cvpr.bbl}
\appendix
\input{11_appendix}

\end{document}

%% file: 01_abstract.tex
\begin{abstract}

Single-view 3D object reconstruction has seen much progress, yet methods still struggle generalizing to novel shapes unseen during training.
Common approaches predominantly rely on learned global shape priors and, hence, disregard detailed local observations.
In this work, we address this issue by learning a hierarchy of priors at different levels of locality from ground truth input depth maps.
We argue that exploiting local priors allows our method to efficiently use input observations, thus improving generalization in visible areas of novel shapes.
At the same time, the combination of local and global priors enables meaningful hallucination of unobserved parts resulting in consistent 3D shapes.
We show that the hierarchical approach generalizes much better than the global approach. It generalizes not only between different instances of a class but also across classes and to unseen arrangements of objects.

\end{abstract}

%% file: 02_introduction.tex
\section{Introduction}
\label{sec:introduction}

The usual problem setting of single-view 3D reconstruction assumes an input image with a single dominant object, where the geometry of both the visible and the invisible part of this object shall be reconstructed.
For the invisible parts, reconstruction must rely on shape priors, which can be based on the object class, symmetry, or smoothness. The geometry of the visible parts can be obtained, at least partially, from sensing data (e.g., depth, texture, shading).

\begin{figure}
  \begin{center}
    \begin{tabular}{c c}
    \includegraphics[width=0.4\linewidth]{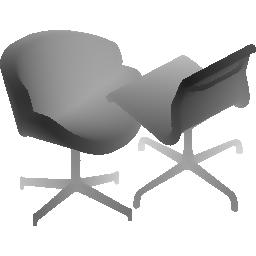} &
    \includegraphics[width=0.4\linewidth]{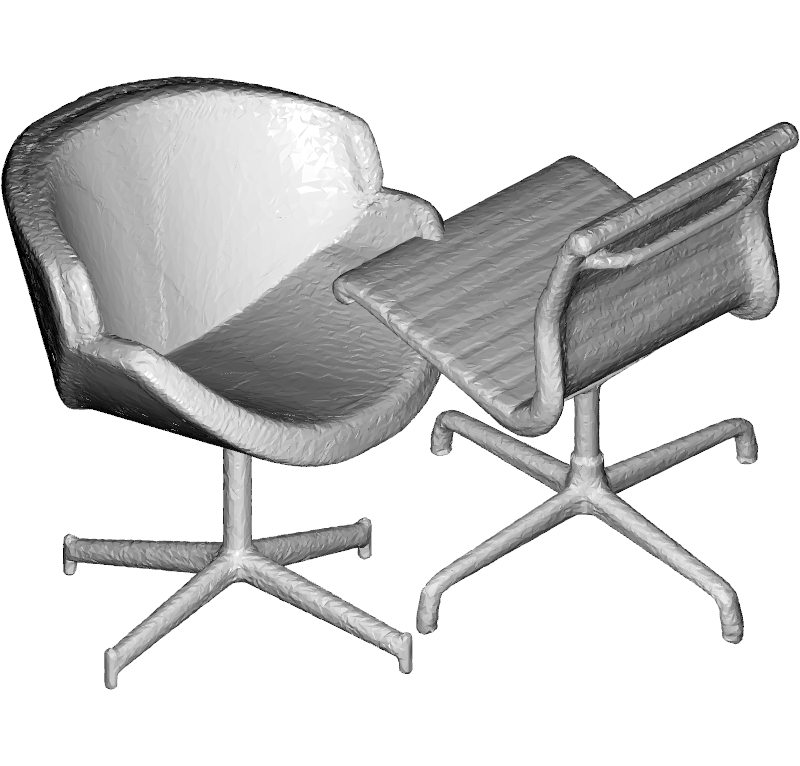} \\
    Input depth & Ground truth\\
    \includegraphics[width=0.35\linewidth]{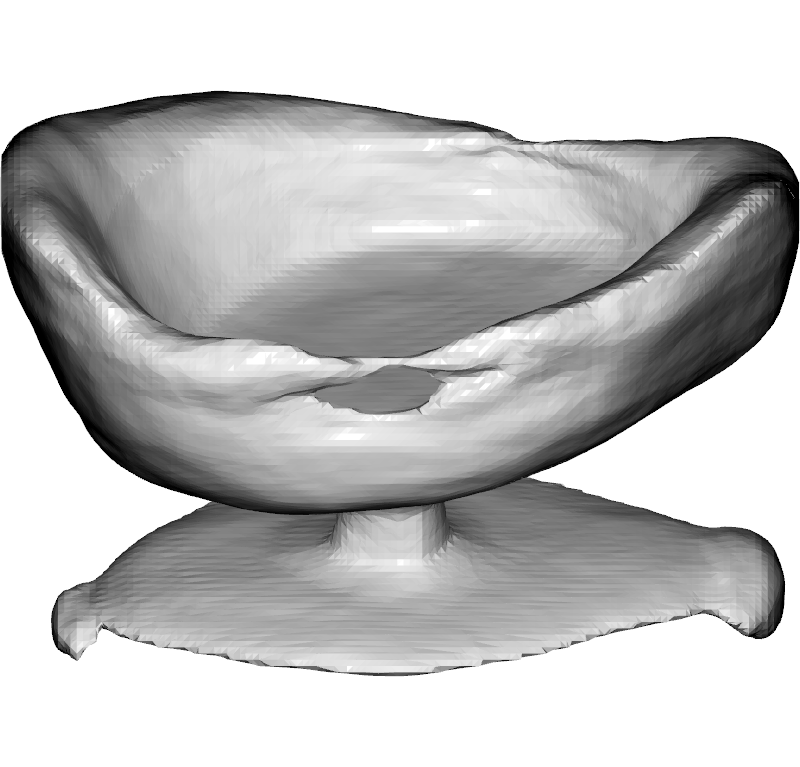} &
    \includegraphics[width=0.4\linewidth]{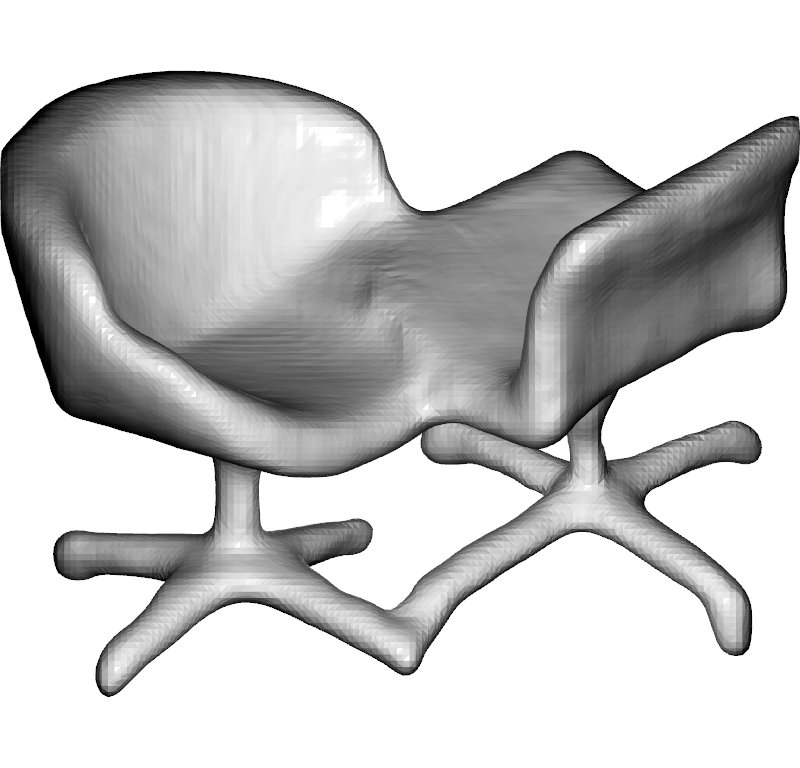} \\ 
    Global prior (ONet) & Hierarchical prior (HPN, ours)\\
   \end{tabular}
  \end{center}
\caption{We employ a hierarchical shape prior to enable recombination of partial shapes observed during training. This significantly improves generalization compared to conventional global shape priors.}
\label{fig:teaser}
\end{figure}

Most existing approaches are encoder-decoder networks~\cite{choy20163d,fan17cvpr,groueix2018papier,mescheder2019occupancy,richter18cvpr,tatarchenko2017octree,wang18eccv} and have been shown to barely generalize to novel shape categories~\cite{zhang2018learninggenre}.
Only few works have targeted generalization explicitly~\cite{bautista2020generalization, thai20203dsdfnet,zhang2018learninggenre}.
They argue that, for better generalization, the problem should be split into two parts: (1) prediction of a geometric representation of the visible parts from a single RGB image and (2) prediction of the final shape from the geometric representation.
In this paper, we focus on the prediction of the object shape and assume the ground truth depth map to be already given as input. This reflects the argument that an intermediate depth map helps generalization~\cite{zhang2018learninggenre} and should make the reconstruction of the visible parts almost trivial.

Surprisingly, however, existing approaches fail to generalize even in the visible areas, despite the perfect input.
Consider the example in Fig.~\ref{fig:teaser}: ONet~\cite{mescheder2019occupancy} trained on single chairs uses its learned prior to reconstruct the shape for an input with two chairs. Although the required shape prior (chairs) has been seen during training, the approach cannot use this knowledge to explain the clean observation of two chairs (Fig.~\ref{fig:teaser} top left), which leads to an unresolved competition between observation and prior (Fig.~\ref{fig:teaser} bottom left). This reveals a general problem of existing approaches: not only do they not generalize to new object classes, they even do not generalize to new combinations of the same training classes. Even if we would train these networks on pairs of chairs, they must see all possible configurations of pairs -- a combinatorial explosion.  

\begin{figure*}[htb!]
\begin{center}
   \begin{overpic}[width=1.0\textwidth]{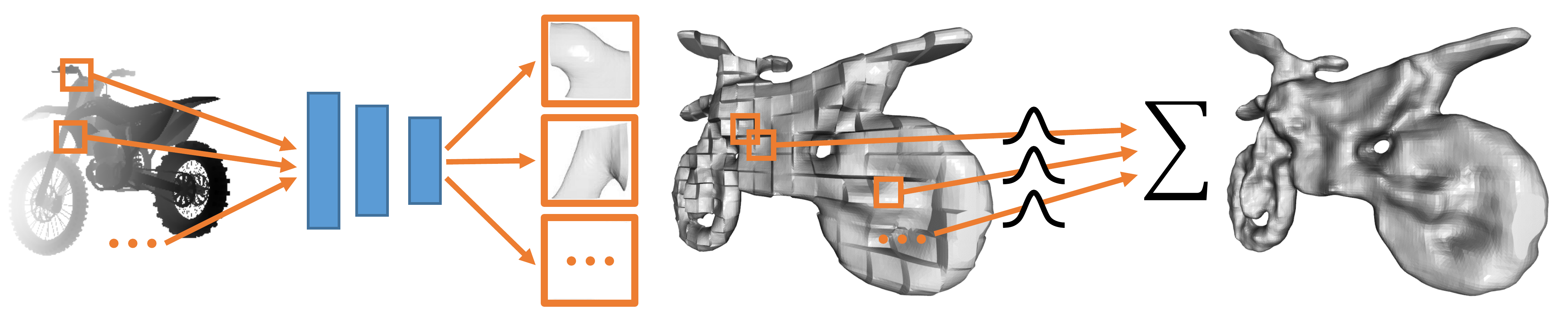}
\put(4,18){$p^1_{i,j}$}
\put(19.7,16){Local@32}
\put(35.5,21){$r^1_{x,y,z}$}
\end{overpic}
\end{center}
   \caption{The proposed local reconstruction module independently reconstructs the shapes of individual patches of the input in a sliding window fashion. The resulting overlapping 3D parts are aggregated with Gaussian-weighted averaging into the final shape estimate.}
\label{fig:local_reconstruction}
\end{figure*}

In this paper, we propose to foster the recombination of previously seen partial shapes by a hierarchical approach.
It consists of two main building blocks: (1) a local reconstruction module that reconstructs the shape at a certain level of locality (Fig.~\ref{fig:local_reconstruction}), and (2) fusion of the beliefs from various levels of locality (Fig.~\ref{fig:hierarchical_reconstruction}).
The reconstruction module is effectively an implicit surface network (e.g. ONet) which performs shape estimation from patches of the input image. If the patch size covers the whole image, it comes down to the original global surface network. 
Intuitively, instead of reconstructing the full shape with a single prediction effort, local versions of the network learn to estimate geometry of individual object parts and put those together to obtain the whole shape.
Since similar shape parts are likely to repeat between different categories, this strategy offers effective recombination of parts from various training samples and, hence, much better generalization potential.

Since local patches have a limited view of the overall shape, the reconstructed global shape may not look consistent, especially in large occluded areas. Therefore, we combine multiple patch sizes (including the global one based on the full image) to form a hierarchy of such local networks. The combination is possible by simple averaging of the logit outputs. 

We demonstrate the intriguing effect of the new hierarchical reconstruction concept on various generalization tasks derived from the ShapeNet \cite{chang15shapenet} dataset. This includes tasks that require inter-class generalization and generalization from single to multiple objects. The results show the huge effect of the ability to recombine parts, which is missing in all previous learning-based reconstruction approaches.
This ability also improves the data efficiency: in contrast to existing global methods, the performance of our local networks does not noticeably degrade even when training on as little as 1\% of the original data.
Since the choice of the base reconstruction module is flexible, the hierarchy of local networks acts as a working principle that can be applied to enhance the generalization of effectively any method based on implicit functions.
We refer to this as Hierarchical Prior Network (HPN).

%% file: 03_related_work.tex
\section{Related Work}
\label{sec:related_work}

\noindent \textbf{3D representations.} A large portion of single-view 3D reconstruction research has dealt with developing methods that operate on different 3D representations.
Those include voxels~\cite{choy20163d}, octrees~\cite{tatarchenko2017octree}, patch-based~\cite{groueix2018papier} or deformable~\cite{wang18eccv} meshes, point clouds~\cite{fan17cvpr}, nested depth maps~\cite{richter18cvpr} and implicit functions~\cite{mescheder2019occupancy, genova2020local}.
All these pipelines effectively follow the same design: a 2D encoder which compresses the input image into a single global latent vector and a 3D decoder which regresses the output 3D representation from it.

\noindent \textbf{3D parts.}
Multiple works reconstruct the output shape as a collection of 3D parts which can come in form of cuboids~\cite{li17grass, niu2018im2struct, abstractionTulsiani17, zou17iccv}, superquadrics~\cite{paschalidou2020learning, paschalidou2019superquadrics}, convex elements~\cite{deng2020cvxnet} or actual semantic parts~\cite{li2019learning, Wu_2020_CVPR}.
All these approaches use parts solely as an alternative 3D representation and do not provide a mechanism for attending to local patches of the input image.
This is different for our method: we directly consider the relationship between local input patches and their 3D counterparts.
Note also that we do not make any assumptions about shape parts being semantically meaningful, which makes our approach general and prevents the need for having semantic annotations similar to \cite{mo2019partnet}.

\noindent \textbf{Generalization.} Only few methods explicitly touch the matter of generalization to shape categories unseen during training.
Shin \etal~\cite{shin2018pixels} and Tatarchenko \etal~\cite{tatarchenko2019single} analyze the conventional setup and conclude that working in the viewer-centered mode is a necessary (though not sufficient) condition for generalization.
Zhang \etal~\cite{zhang2018learninggenre}, Wu \etal~\cite{shapehd} and Thai \etal~\cite{thai20203dsdfnet} propose to predict intermediate geometric representations in the pipeline and show that this improves generalization.
In our work, we use a similar setting but further simplify it by starting from a ground truth depth map.
Surprisingly, we find that even then the actual generalization achieved by existing methods is still limited.
Thai \etal \cite{thai20203dsdfnet} show that using three-degree-of-freedom camera poses and SDFs as a 3D representation, while keeping the architecture from \cite{mescheder2019occupancy}, helps generalize to a new dataset.

\noindent \textbf{Local encoding.}
Several existing works proposed to include local encoding modules into the pipeline.
Xu \etal~\cite{xu2019disn} combine local and global features with the aim of improving the reconstruction details. However, their method is not forced to use local information and could in principle ignore it, plus they never explicitly target the generalization setting. For a special case of reconstructing human clothing, Saito \etal~\cite{saito2019pifu} propose to align local per-pixel features to the global shape context, thus explicitly leveraging the 2D-3D relationship.
Peng \etal~\cite{peng2020convolutional} combine a local encoder with an implicit function decoder for a task of point-cloud-based surface reconstruction. Similarly, multiple works \cite{Badki20meshlet, Jiang20locimplgrid, Chabra20deeplocalsdf} target a setting
where surfaces are locally reconstructed from sparse multi-view observations.
Similar in spirit to our approach, Chibane \etal~\cite{chibane2020implicit} propose to extract a hierarchy of features for solving several 3D-to-3D tasks.
Bautista~\etal~\cite{bautista2020generalization} locally assign features and 3D points in order to get a more expressive intermediate shape representation. Most similar to ours is the work from Genova~\etal~\cite{genova2020local}. Local Gaussian regions of the input depth map are encoded and decoded independently. The global 3D shape results from the sum of the deep local implicit functions. However, the module that distributes the Gaussian regions requires a global context and can break if major dataset priors, like that of having a single object, are violated.

%% file: 04_method.tex
\section{Method}
\label{sec:method}

The core idea of our approach is based on two observations: 
(1) effective generalization to new classes and new configurations requires the recombination of partial shapes seen during training; (2) recombination of such partial shapes requires (local) support regions of different sizes in the input image.

Although the regular encoder-decoder networks consider a hierarchy of multiple receptive field sizes when observing the input, they do not learn local \emph{priors} during training. This is because their loss function only considers the whole object reconstruction, for which all of the input image and all of the ground truth shape is observed. While all the information for recombination is available, there is nothing in the training procedure that requires and fosters recombination. 

For this reason, we combine multiple local reconstruction networks that only observe a cut-out part of the image and the corresponding cut-out part of the ground truth shape during training. The different levels of locality yield networks that have learned more specialized (global) or less specialized (local) priors. In their combination, they enable part recombination at all locality levels and consistency of the global shape at the same time. 

\subsection{Local Reconstruction}
\label{sec:local_reconstruction}
Consider a single-channel input depth map \(d \in \mathbb{R}^{W\times H}\) of width \(W\) and height \(H\) pixels, and its corresponding ground truth 3D model $D$ represented as a mesh with vertices $V_D$ and faces $F_D$.
Following the conventional setup in literature, we assume that $D$ is normalized such that it fits into a unit cube.

For the \(\ell\)-th hierarchy level, we denote with \(N^{\ell} \in \mathbb{N}\) the width and height in pixels of a square patch \(p^{\ell}_{i,j}\subset d\) centered at pixel position \((i,j)\).
These patches are positioned across the input \(d\) using a stride of \(s^{\ell}_{\text{train}}\) pixels.
For each $p^{\ell}_{i, j}$ there is a corresponding 3D volume $r^{\ell}_{x, y, z}$ centered at position $(x, y, z)$ in the ground-truth 3D model.
In the general case, the shape of $r^{\ell}_{x, y, z}$ is a frustum determined by the internal camera parameters,
and the 3D position \(x,y,z\) depends on the patch location ${i,j}$ and the camera model.
For simplicity, we assume an orthographic camera model which results in $r^{\ell}_{x, y, z}$ being a cuboid with $x=y=M \in (0, 1)$ and $z=1$.
However, the whole setup could be extended to support perspective cameras.

Our local reconstruction module is an implicit function \(f^{\ell}\), for example an Occupancy Network (ONet), which takes as input a patch $p^{\ell}_{i, j}$ and some points $\mathbb{SP}^{K\times 3}$ in \(r^{\ell}_{x, y, z}\) and outputs 3D predictions for $r^{\ell}_{x, y, z}$ in form of an occupancy logit or signed distance value for every input point. ONet could be replaced by any other network that implements an implicit function in 3D.

We extract a mesh from the occupancy logits by using Marching Cubes~\cite{lorensen1987marching} with an empirically determined threshold $\tau$ as described in Occupancy Networks. We use the same procedure if the backbone network predicts SDF values, but determine a new threshold.

At training time, each 3D part is effectively treated as an independent sample, i.e. the only difference to the original ONet is in the training data. Therefore we normalize the training points from the 3D part \(r^{\ell}_{x, y, z}\) to lie within $[-0.5,0.5]$ in all three dimensions.
Similar to Occupancy Networks, during training, we only provide a randomly sampled subset of training points to the network.

During inference, the network is applied in a sliding window fashion with a 2D stride $s^{\ell}_{\text{infer}}$, such that each 3D region of the prediction gets updated by multiple parts.
This enables smoother transitions between adjacent parts.
We fuse predictions from multiple parts together by Gaussian-weighted averaging of the outputs of all contributing parts in the overlapping regions.

Since we assume that the camera model is known, there is a deterministic assignment between the predicted 3D parts and their absolute locations within the unit cube of the full shape.
We use it to assemble a full reconstructed shape from individual predicted parts. An example of such a reconstruction for patches of size $N=32$ and stride $s=16$ is shown in Fig.~\ref{fig:local_reconstruction}.

\subsection{Hierarchical fusion}
We train multiple local reconstruction networks, each operating on different patch sizes $N$ (including $N=W=H=256$, i.e. the full image case).
In case of non-square input images we suggest to use zero-padding in order to convert them into a square shape.
Together the local reconstruction networks form a hierarchy of $K$ independent predictions relying on priors of different locality levels \(\ell\in\{1,\ldots,K\}\) which we then fuse into a single final prediction.

\begin{figure}[t]
\begin{center}
   \includegraphics[width=1.0\linewidth]{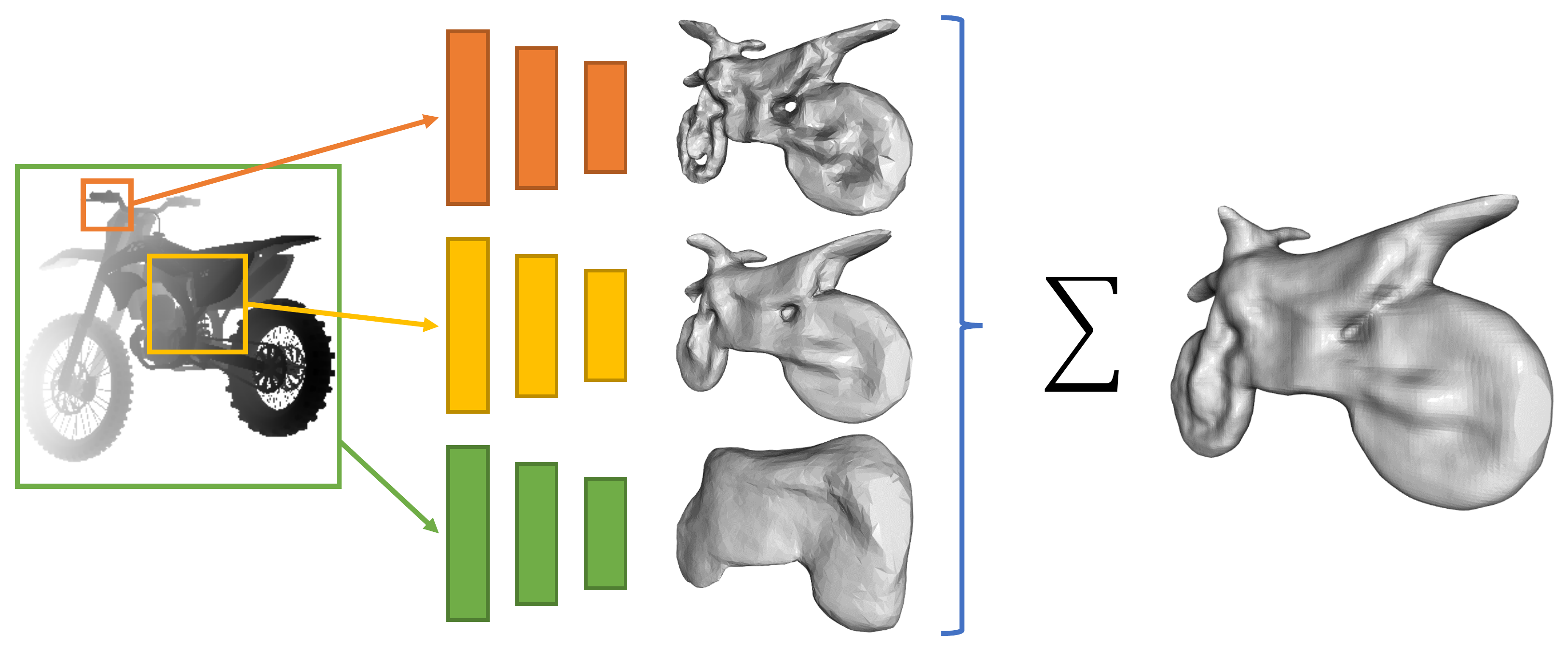}
\end{center}
   \caption{We use a hierarchy of networks operating on input patches of different resolutions (including the global one) to produce multiple shape reconstruction variants. Those are fused by simple averaging to yield the final reconstruction.}
\label{fig:hierarchical_reconstruction}
\end{figure}

Similarly to averaging softmax outputs of overlapping parts in the previous section, we combine predictions from different hierarchy levels by averaging their corresponding softmax outputs. 
Since individual output values correspond to pseudo-probabilities that a certain 3D region is occupied, averaging them already provides an automatic mechanism to weigh the contribution of each level onto the final fused reconstruction.
For example, in areas of the shape which are visible in the input image where the local reconstruction is usually more confident, local occupancy scores dominate those of the global one, and vice versa for invisible shape regions.
The fusion of hierarchy levels is illustrated in Fig.~\ref{fig:hierarchical_reconstruction} where three hierarchy levels of differing local patch sizes are fused to produce a single reconstruction. We call this combination of networks acting at multiple levels of locality Hierarchical Prior Network (HPN).  

More sophisticated (learned) averaging schemes are conceivable, but come with the risk of overfitting to the training configurations. As we show in the experiments, already simple averaging leads to consistent shapes and is free from a bias to the training set. 

%% file: 05_experiments.tex
\section{Experiments}
\label{sec:experiments}

\begin{figure*}
\begin{overpic}[width=1.0\textwidth]{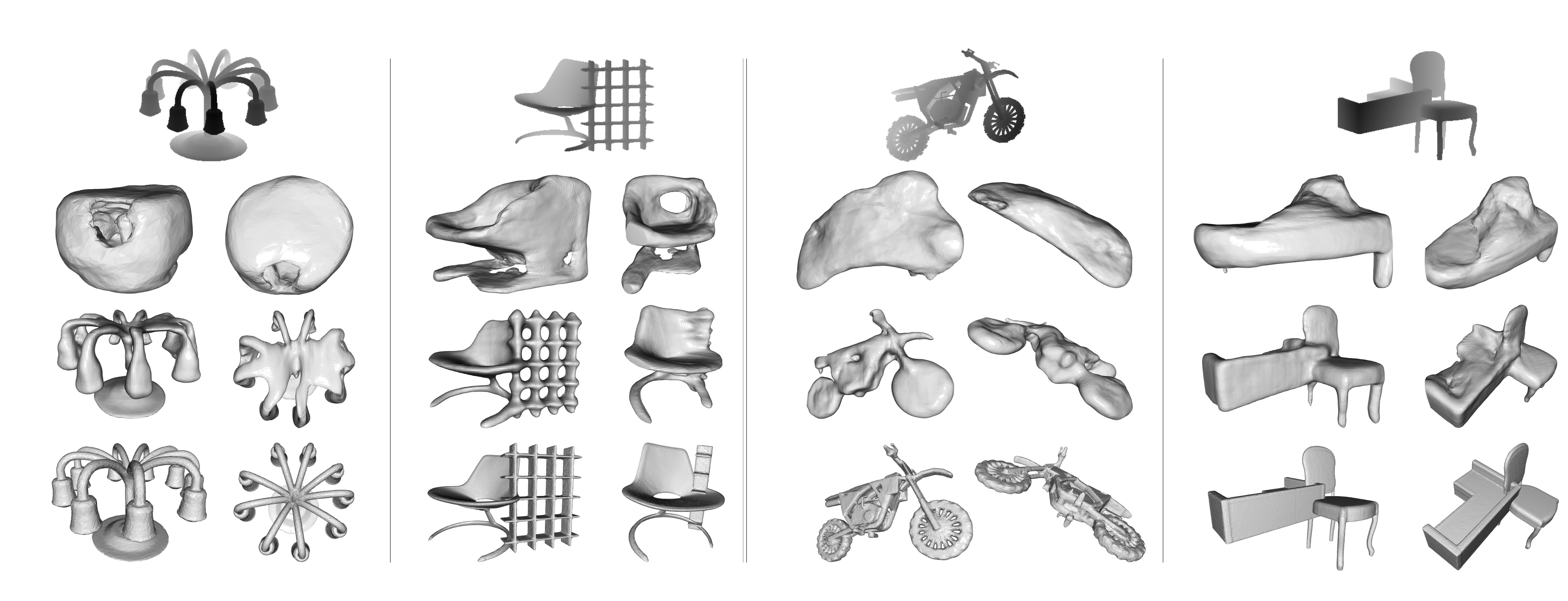}
\put(0.5,30){\rotatebox{90}{Input}}
\put(48.5,30){\rotatebox{90}{Input}}
\put(0.5,21){\rotatebox{90}{ONet}}
\put(48.5,21){\rotatebox{90}{ONet}}
\put(0.5,11){\rotatebox{90}{HPN-SDF}}
\put(48.5,13){\rotatebox{90}{HPN}}
\put(0.5,5){\rotatebox{90}{GT}}
\put(48.5,5){\rotatebox{90}{GT}}
\put(16,38){Trained on \textit{multi-class}}
\put(68,38){Trained on \textit{lamps}}
\end{overpic}
\caption{Reconstruction results for unseen classes in the different generalization settings. \textbf{Left:} Networks trained in the \textit{multi-class} setting (on planes, cars and chairs). \textbf{Right:} Networks trained on lamps. More examples are provided in supplemental~\ref{sec:apx_qualitative}.}
\label{fig:main_results}
\end{figure*}
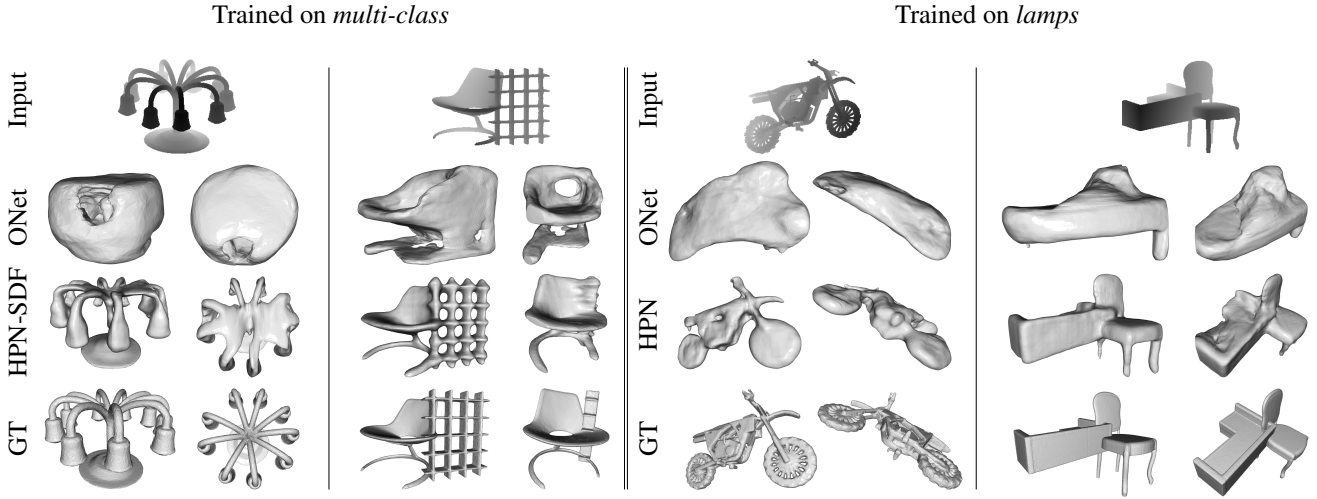

\input{tables/main_results}

Existing approaches generalize to a certain degree to novel instances of a category seen during training. We target the more difficult generalization to novel categories and novel object assemblies. 

\subsection{Datasets}

We train our method on two different subsets of the ShapeNet dataset~\cite{chang15shapenet}.
(1) We report on the train split from Zhang \etal~\cite{zhang2018learninggenre} referred to as \textit{multi-class}, where networks are trained on planes, cars, and chairs. (2) We train on shapes from only a single category \textit{(single class)}. These training categories are \textit{chair} or \textit{lamp}.

We evaluate our method on individual shape categories as suggested by Zhang \etal~\cite{zhang2018learninggenre}, both on the ones seen during training, corresponding to generalization across instances, and on those not seen during training, corresponding to generalization across classes.

In addition, we propose a new test set referred to as \textit{Composition}, which allows us to explicitly evaluate generalization to novel object arrangements.
We create it by placing up to three objects into one image. We exclusively use shape instances from the ShapeNet test set. For each compositional image, we randomly select the shape categories. Then, we pick objects of the selected categories and modify elevation and azimuth of their pose. Before rendering the image with PyTorch3D~\cite{ravi2020pytorch3d}, we shift the objects along the x-axis to reduce their overlap. 

\subsection{Models}
\noindent \textbf{ONet.}
We train the original occupancy network~\cite{mescheder2019occupancy} on the ground truth depth images. 

\noindent \textbf{GenRe.}
GenRe~\cite{zhang2018learninggenre} is the pioneer work for generalization to novel categories. The GenRe network architecture consists of two parts. The first one estimates a depth map for a given RGB image. The second one reconstructs the 3D shape, given the depth image. We report the Chamfer distance from their paper for reconstructions from ground truth depth maps.
For a comparison of all 13 test classes please see Tbl.~\ref{tbl:apx_results}.

\noindent \textbf{LDIF.}
LDIF~\cite{genova2020local} represents 3D shapes as multiple local implicit functions and improves over ONet and GenRe, thus being the state-of-the-art method. We use their custom data preprocessing pipeline to train LDIF networks on single-view perspective depth images.

\noindent \textbf{ONet-SDF.}
We use the same training points as for Occupancy Networks, but replace the binary occupancy label with the signed distance (SDF) of each point to the mesh surface. Points within the mesh have a negative distance. This also changes the training of the network from binary classification to regression. Instead of using the binary cross entropy as loss, we now use the $L_1$ loss. In order to extract a mesh from the SDF-values predicted by the network, we empirically determine the new threshold $\tau_{sdf} = -0.02$. Therefore, we pick $\tau_{sdf}$ from the interval $[-1,1]$ with a stepsize of $0.1$ and a smaller stepsize of $0.01$ in the interval $[-0.1,0.1]$. 

\noindent \textbf{HPN and Local@N.}
As described in Sec.~\ref{sec:method} we design local variants of the ONet and a fused variant for which different hierarchy levels are combined.
In general, we refer to the fused variant as hierarchical prior networks (HPN) and to its local variants as Local@N where \(N\) is the width and height of a local patch in pixels, \eg, Local@64 for patches of size \(64\times64\) pixels.
HPN is the fused version of Global@256, Local@64 and Local@32.
HPN-SDF is the fused version of Global@256-SDF, Local@64 and Local@32, i.e. the SDF representation is used for the global but not for the local networks.

\subsection{Setup}
\label{subsec:exp_setup}

\noindent \textbf{Training.}
All networks were trained using the ADAM optimizer~\cite{kingma2014adam} with the same optimization settings as used for the Occupancy Network~\cite{mescheder2019occupancy}. 
We trained all networks until convergence.
Similar to Occupancy Networks,
during training, we only provide a randomly sampled subset
of \(1500\) training points to our local networks.

\noindent \textbf{Evaluation metrics.}
We report quantitative results for two widely used 3D reconstruction metrics: F-score~\cite{Knapitsch2017fscore} and Chamfer distance (CD)~\cite{barrow1977chamfer}.
The two scores highlight different aspects of the reconstruction, as the F-score is robust to outliers (large deviations) and CD is not.
We further discuss this point in Sec.~\ref{sec:analysis}.
For completeness, we list the IoU values in \ref{tbl:apx_results_iou}.

As part of our analysis, we additionally report F-score and Chamfer distance for the parts of the 3D shape that are visible from the input image and the parts that are invisible (self occluded) from the input image.
In order to determine the visibility label, we project a set of points from the ground truth mesh into the depth image and check, whether they coincide with the respective depth value (visible) or are larger than the respective depth value (invisible). We do this for all test shapes, s.t. during evaluation we can look up the visibility label and compute the metrics separately.

\noindent \textbf{Implementation.} All the networks are implemented in PyTorch~\cite{NEURIPS2019_9015}. For visualizing qualitative examples, we used the Open3D~\cite{Zhou2018open3d} framework.

\subsection{Results}
Fig.~\ref{fig:main_results} shows the drastically improved generalization to new shape classes and shape configurations compared to the state of the art. None of the networks has seen such categories during training, but thanks to the ability to flexibly recombine training parts, the hierarchical prior can also reconstruct completely new shapes in a reasonable quality. 
This also includes the composition of two objects, which was never observed during training. In contrast, the plain ONet model is bound to the most similar global shapes during training, which is insufficient in all these examples. Remembering the nice-looking reconstructions from literature, one should be aware that these were obtained via largely overlapping training and test sets. 

Although the effect of the local recombination principle is already evident and indisputable from just the visual impression, Tbl.~\ref{tbl:results} also quantifies this effect. 
In all train-test configurations HPN outperforms the baselines and the previous state of the art in generalization.
The performance almost doubles on unseen classes, both in terms of F-Score and Chamfer distance, in comparison to ONet. It also significantly improves over LDIF in terms of F-Score. For Chamfer distance, LDIF is competitive with HPN. We hypothesize that this happens because for some shapes our local networks produce outliers which have a large impact on the mean distance. Interestingly, LDIF represents the training classes better than all other methods but completely fails on compositional shapes. This indicates that LDIF is capable of nicely fitting the training data which is not useful when generalization is required.
The use of signed distance functions yields more detailed reconstructions in conjunction with our hierarchical prior network (HPN-SDF), leading to best scores in the \textit{multi-class} setting.

All approaches achieve consistently better scores on the unseen categories than on the new compositional test dataset. We conclude that the compositional setting is more difficult. One reason might be that one shape occludes the other, which requires to reconstruct the front side of the occluding shape (bookshelf), and the backside of the occluded shape (chair); see Fig.~\ref{fig:main_results}.

\subsection{Analysis}
\label{sec:analysis}

\subsubsection{Different hierarchy levels}
\begin{figure*}[ht]
  \begin{center}
    \begin{tabular}{c c c c c c c c}
        Input & 16 & 32 & 64 & 128 & 256 & Fused & GT\\
        \includegraphics[height=0.11\linewidth]{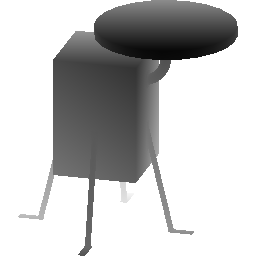} &
        \includegraphics[height=0.11\linewidth]{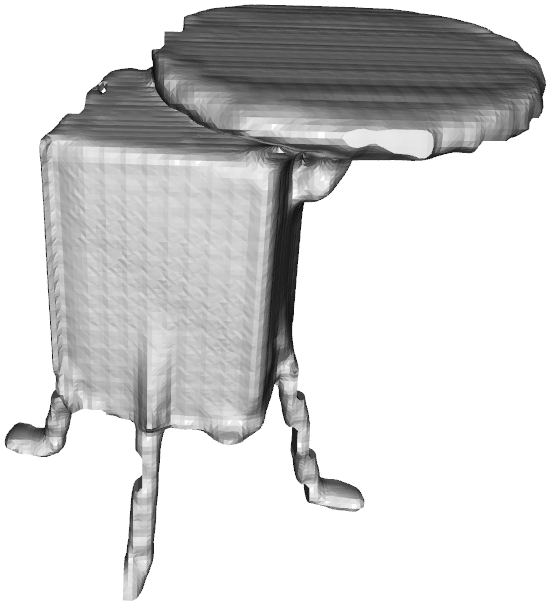} &
        \includegraphics[height=0.11\linewidth]{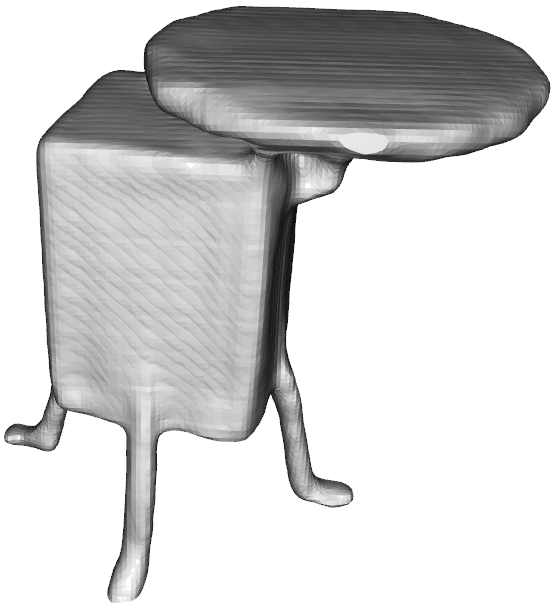} &
        \includegraphics[height=0.11\linewidth]{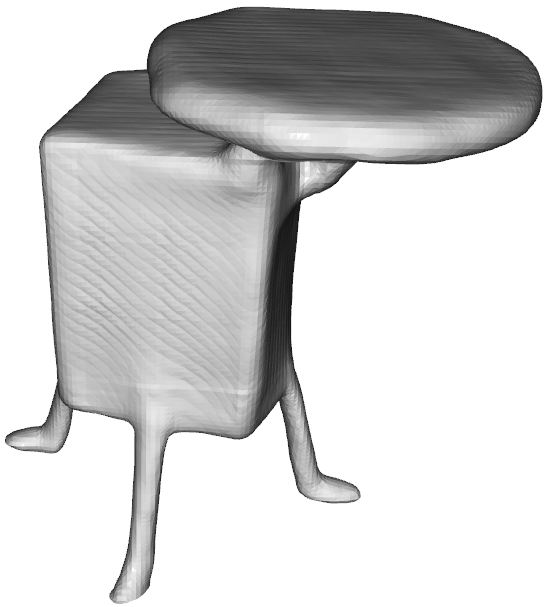} &
        \includegraphics[height=0.11\linewidth]{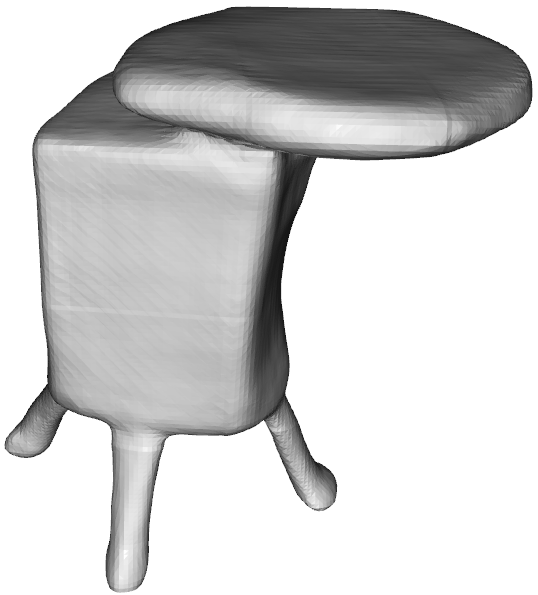} &
        \includegraphics[height=0.11\linewidth]{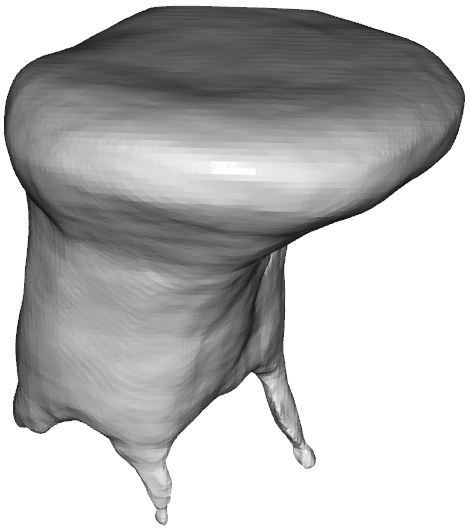} &
        \includegraphics[height=0.11\linewidth]{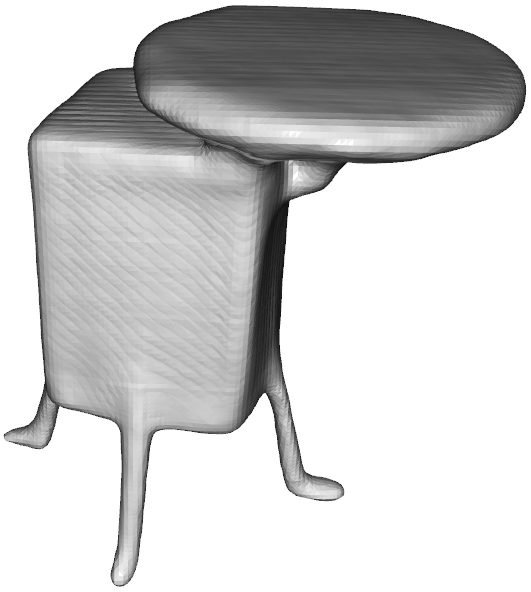} &
        \includegraphics[height=0.11\linewidth]{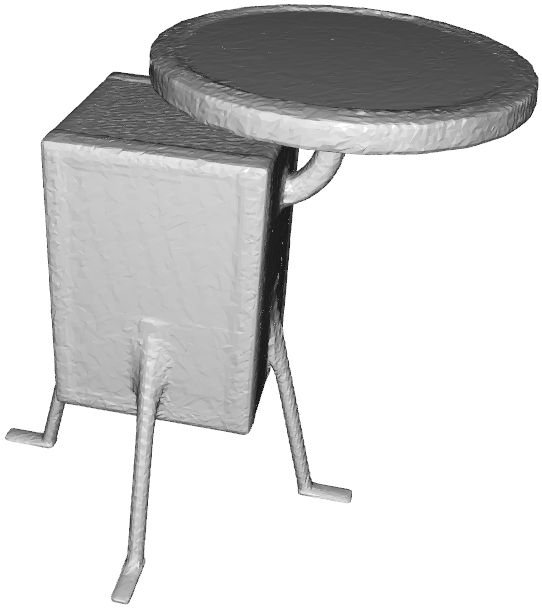} \\
        \includegraphics[height=0.11\linewidth]{figures/different_levels/input_white} &
        \includegraphics[height=0.11\linewidth]{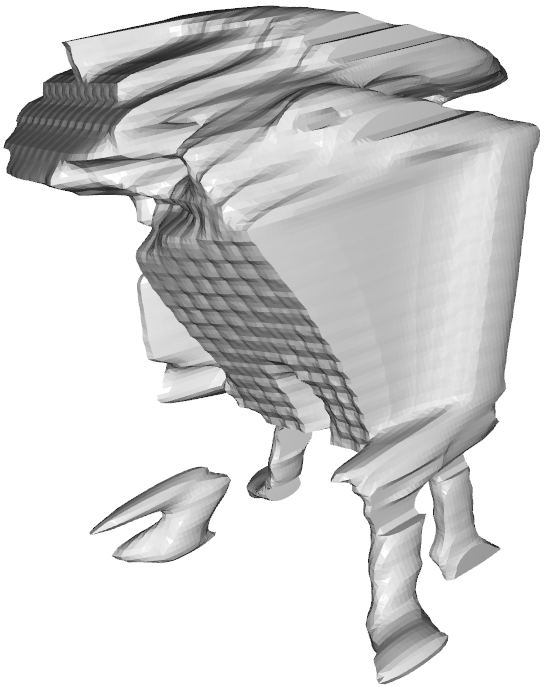} &
        \includegraphics[height=0.11\linewidth]{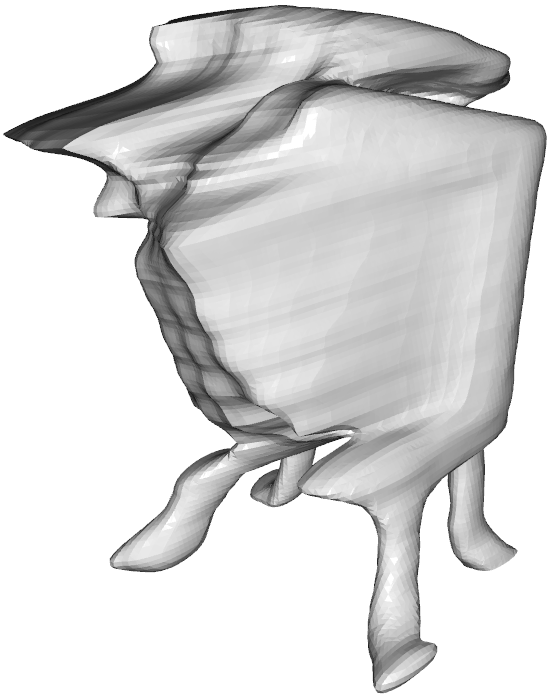} &
        \includegraphics[height=0.11\linewidth]{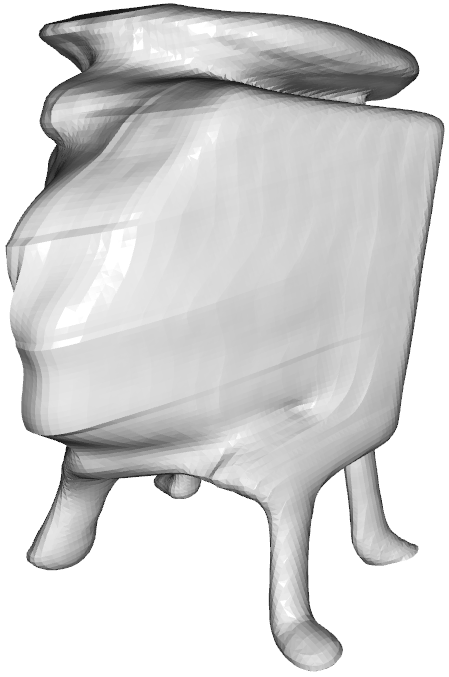} &
        \includegraphics[height=0.11\linewidth]{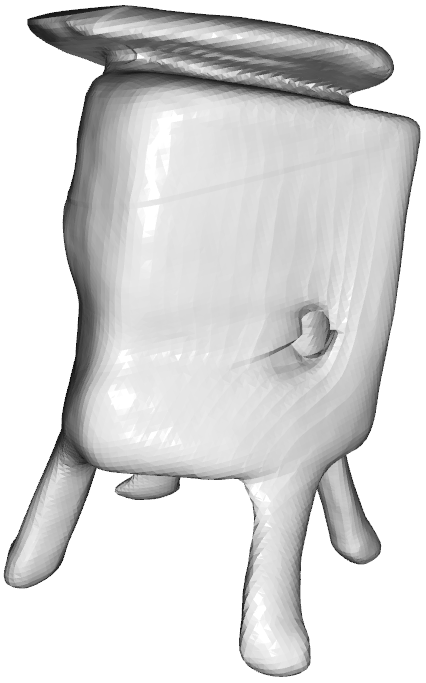} &
        \includegraphics[height=0.11\linewidth]{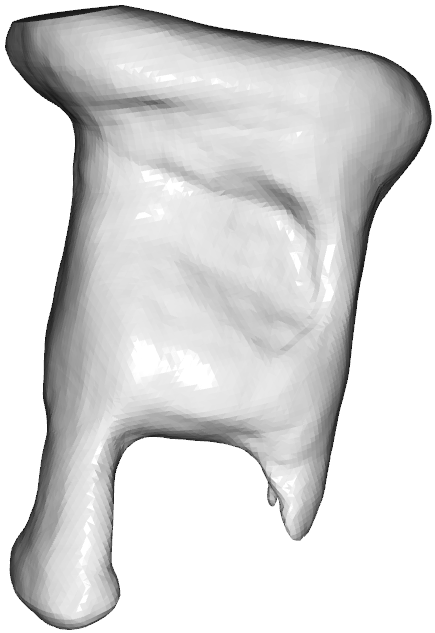} &
        \includegraphics[height=0.11\linewidth]{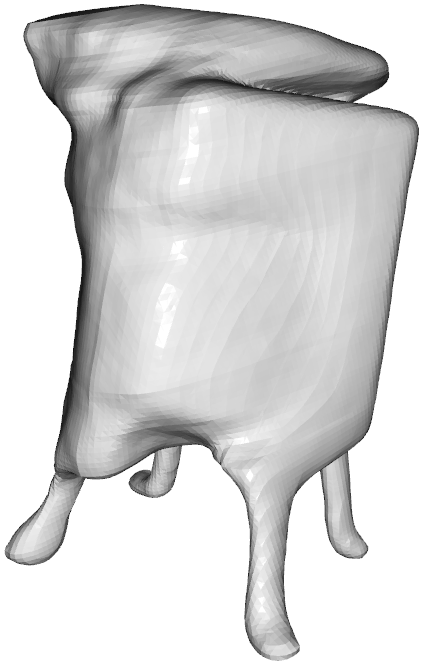} &
        \includegraphics[height=0.11\linewidth]{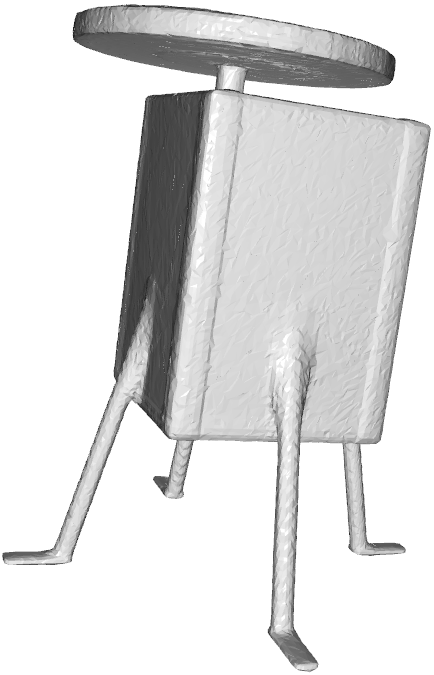} \\
    \end{tabular}
  \end{center}
\caption{Reconstructions produced by different hierarchy levels. Numbers correspond to different patch sizes N. \textbf{Top row:} Same viewpoint. \textbf{Bottom row:} Opposite viewpoint. Reconstructions for very small patches are particularly noisy, since they see little context and the smaller overlap area reduces the spatial smoothing effect. However, due to the aggregation with other levels, this has no negative effect on the fused reconstruction.}
\label{fig:analysis_qualitative}
\end{figure*}

We investigated the reconstruction by individual local networks and how they contribute to the full hierarchical reconstruction.
Fig.~\ref{fig:analysis_qualitative} shows an example and Tbl.~\ref{tbl:analysis_results} reports test set scores on the full shape, as well as the visible and invisible parts of it. All models are trained on chairs and evaluated on the other categories.
In visible areas, the local networks reconstruct details much better than the global network, which highlights the problem that global priors interfere with the measurements in these areas.
Local networks with the smallest patch size (16 and 32) are particularly noisy in the invisible areas.
Surprisingly, local models with larger patch size perform a bit better (on average) than the global network in the invisible areas.
This supports our recombination idea and indicates that explaining even the invisible shape regions with a collection of local priors may have advantages over using a single global one.

\input{tables/analysis_results}

The best reconstruction scores are achieved when combining all available sources of information: HPN@(256+128+64+32+16) works better than any other configuration. This result is slightly unintuitive given that the most local reconstruction levels (32 and 16) on their own do not provide any quantitative improvement over the more global ones (128 and 64). We hypothesize that this can be attributed to the full set of networks acting like an ensemble, which averages out the errors of individual levels thus improving the final score. Note that more hierarchy levels also means higher computational cost. As a trade-off between efficiency and accuracy, in the rest of the paper we only use three levels, i.e. HPN@(256+64+32).

\begin{figure}
  \begin{center}
  \begin{overpic}[width=0.48\textwidth]{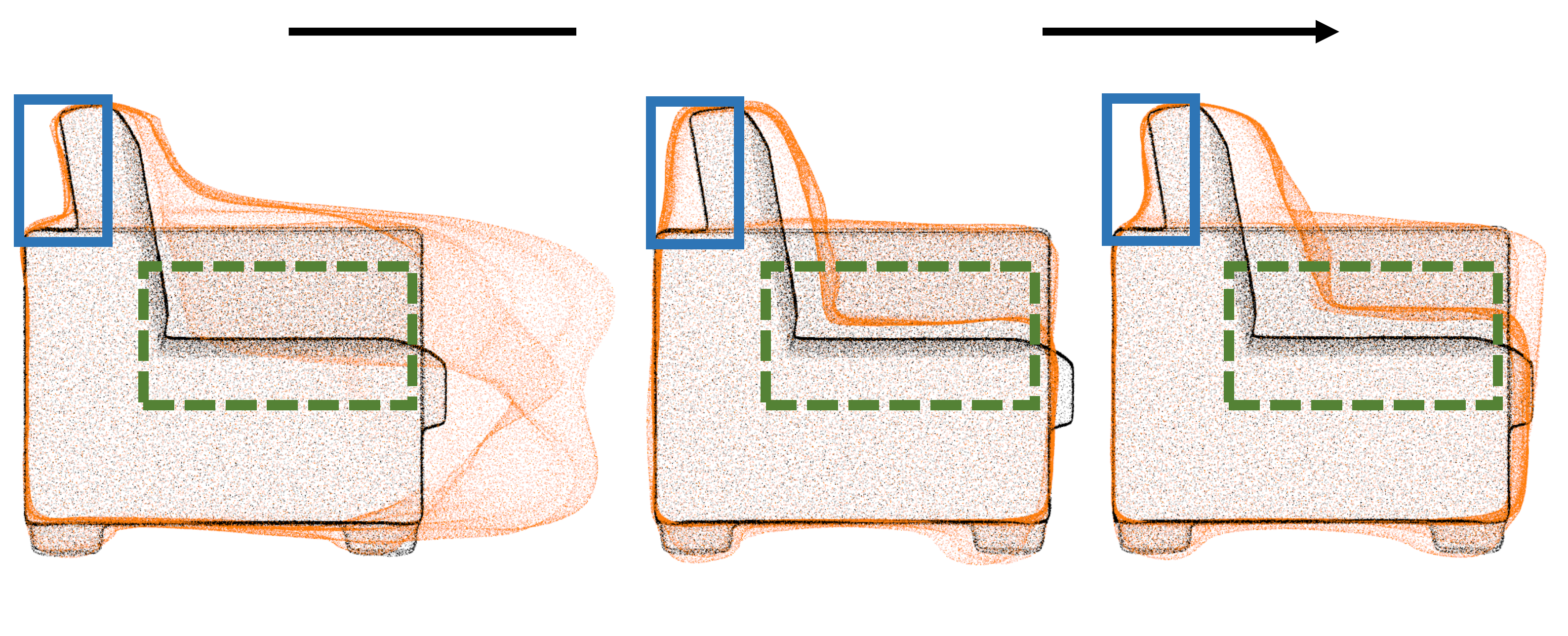}
    \put(38.4,37.6){\small viewing direction}
    \put(12,0){\small Local@64}
    \put(51,0){\small ONet}
    \put(81,0){\small HPN}
  \end{overpic}
  \end{center}
\caption{Point cloud of a couch (side-view). \textbf{Black:} Ground truth. \textcolor{orange_our_bright}{\bf Orange:} Predicted shape. Marked boxes indicate regions where the local and global predictions differ significantly. Best viewed in color.}
\label{fig:metrics}
\end{figure}
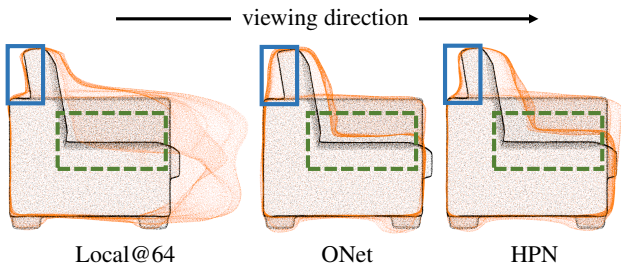

We can better understand the properties captured by individual quantitative metrics by looking at Fig.~\ref{fig:metrics}. The local reconstruction is very precise in the visible area (blue box) but completely wrong in the invisible part. The global reconstruction acts the other way around: it is off in the visible area but provides a plausible prior for the invisible part. The fused version gets much better on average - this is captured well by the CD. However, certain parts are still not perfect, e.g. the HPN reconstruction within the dashed green box is a bit off. Because of that, the F-score, being a robust metric, may not react so strongly to such changes. One should be aware of this when interpreting the quantitative results.

\subsubsection{Data efficiency}
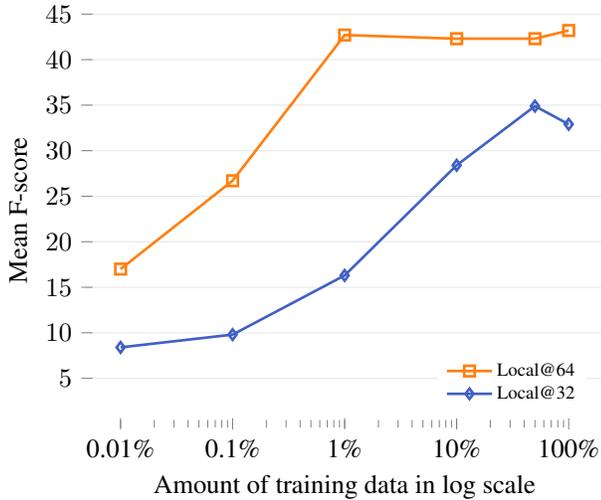
\begin{figure}[ht]
\begin{center}
    \begin{tikzpicture}
    \begin{semilogxaxis}[
        xlabel={Amount of training data in log scale},
        ylabel={Mean F-score},
        xmin=0.005, xmax=200,
        ymin=0, ymax=47,
        xticklabels={0.01\%,0.1\%,1\%,10\%,100\%},
        xtick={0.01,0.1,1,10,100},
        ytick={5,10,15,20,25,30,35,40,45},
        legend pos=south east,
        legend style={draw=none,nodes={scale=0.7}},
        ymajorgrids=true,
        grid style={line width=.1pt, draw=gray!20},
        axis lines=left,
        axis line style={draw=none}
    ]
    
    \addplot[ 
        color=orange_our_bright,
        mark=square,
        line width=1pt,
        ]
        coordinates {
            (0.01,17.0)
            (0.1,26.7)
            (1,42.7)
            (10,42.3)
            (50,42.3)
            (100,43.2) 
        };
    \addplot[ 
        line width=1pt,
        color=blue_our,
        mark=diamond, 
        ]
        coordinates {
            (0.01,8.4)
            (0.1,9.8)
            (1,16.3)
            (10,28.4)
            (50,34.9)
            (100,32.9) 
        };
    \legend{Local@64,Local@32,HPN,ONet}
        
    \end{semilogxaxis}
    \end{tikzpicture}
\end{center}
\caption{Reconstruction quality in dependence of the number of training samples. Local reconstruction reaches its full performance already with as little as 1\% of the training data.}
\label{fig:lessdata}
\end{figure}

Another expression of improved generalization is the required use of training data. 
Because the patch-based networks can effectively learn to recombine, hence reuse, parts seen during training to reconstruct novel shapes, we expect to need less training data for the patch-based networks to reach the same performance as the global variant. We evaluated this claim, comparing our Local@64 with the global ONet on the unseen categories of the multi-class setting for different amounts of training examples. Results are summarized in Fig.~\ref{fig:lessdata}.

As expected, we see a significantly higher mean F-score for the patch-based network (bright orange) compared to the global network (dark blue) for all training dataset sizes.
The local network reaches its full performance already with just 1\% of the training data. 
Both networks converge for large amounts of training data. 
Two effects cause this data efficiency: (1) The local parts are less complex than a global shape, i.e., they require less data to be represented.  (2) Each training sample comprises many local parts, which increases the effective training set size. 

\subsubsection{Failure cases}

\begin{figure}[t]
  \begin{center}
    \begin{tabular}{c c c c} 
    Input & GT & ONet & HPN (Ours) \\
    \includegraphics[width=0.17\linewidth]{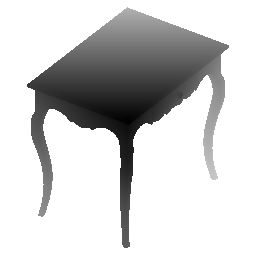} &
    \includegraphics[width=0.2\linewidth]{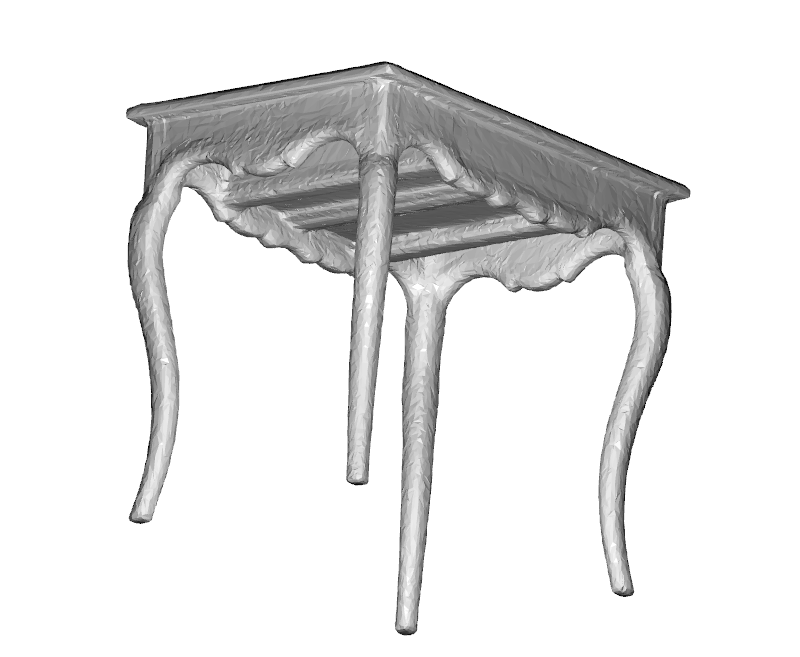} &
    \includegraphics[width=0.2\linewidth]{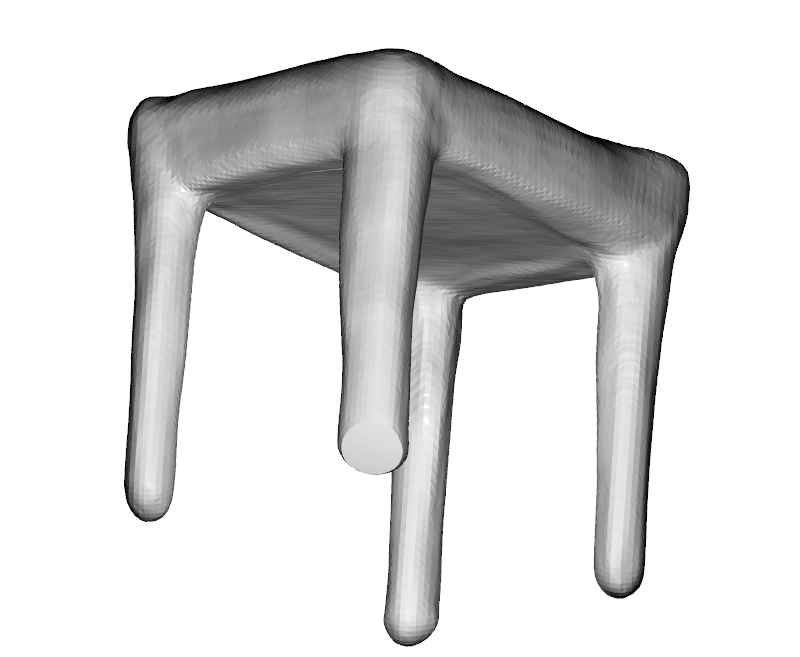} &    \includegraphics[width=0.2\linewidth]{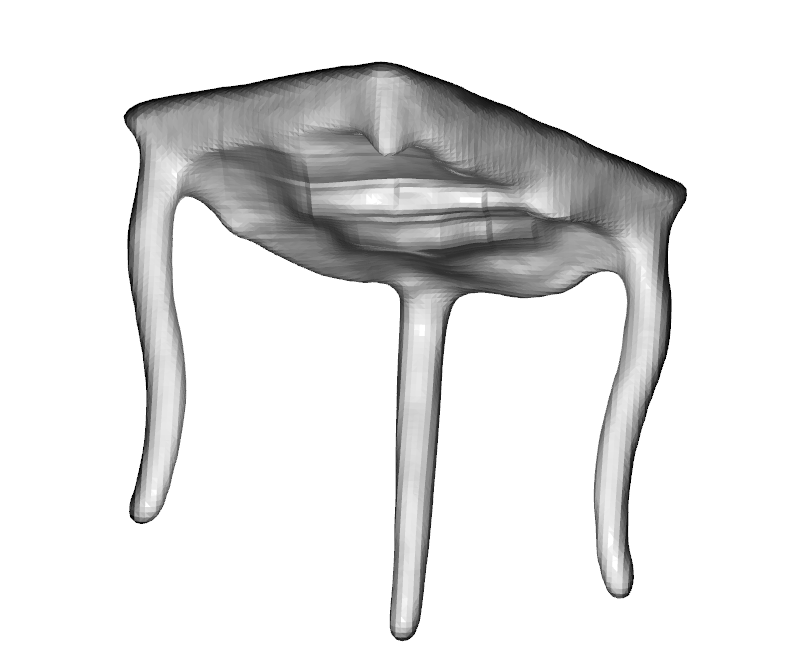}\\ 
    \includegraphics[width=0.16\linewidth]{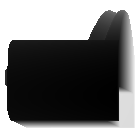} &
    \includegraphics[width=0.2\linewidth]{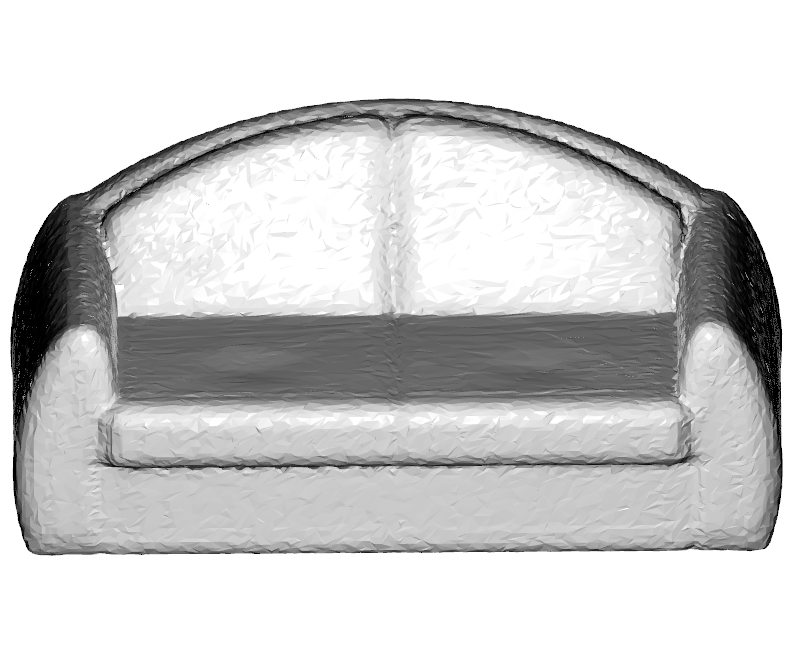} &
    \includegraphics[width=0.2\linewidth]{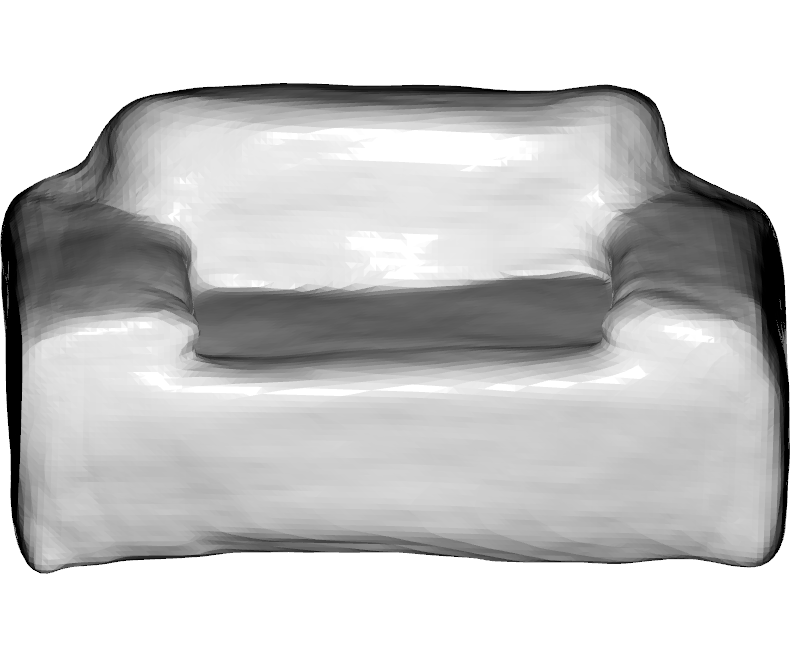} &    \includegraphics[width=0.2\linewidth]{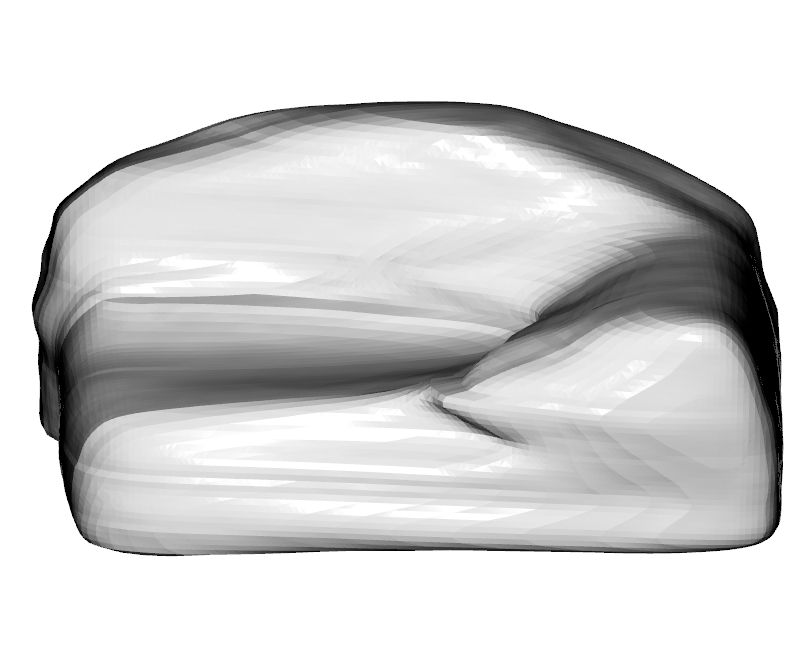}\\ 
    Input & GT & Local@64 & HPN (Ours) \\
    \includegraphics[width=0.17\linewidth]{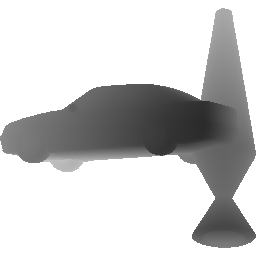} &
    \includegraphics[width=0.2\linewidth]{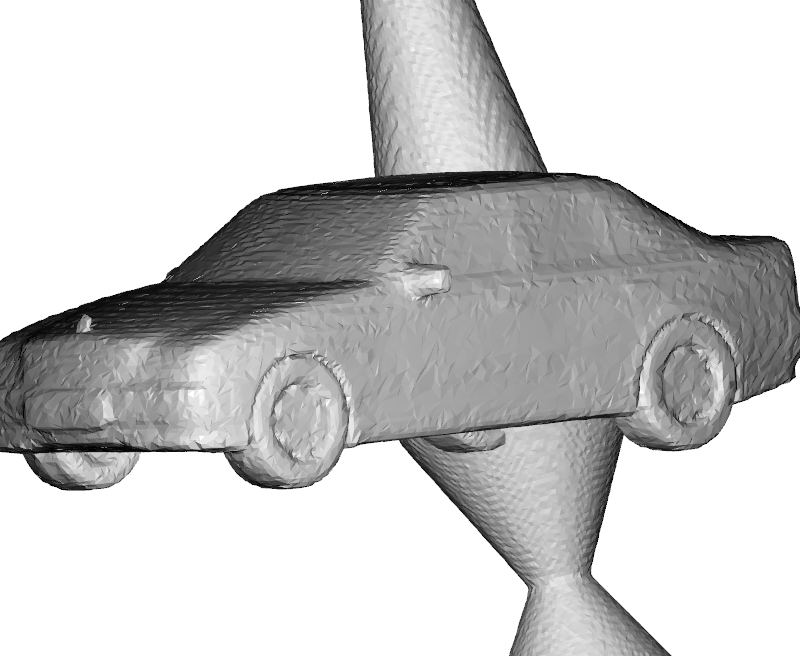} &
    \includegraphics[width=0.2\linewidth]{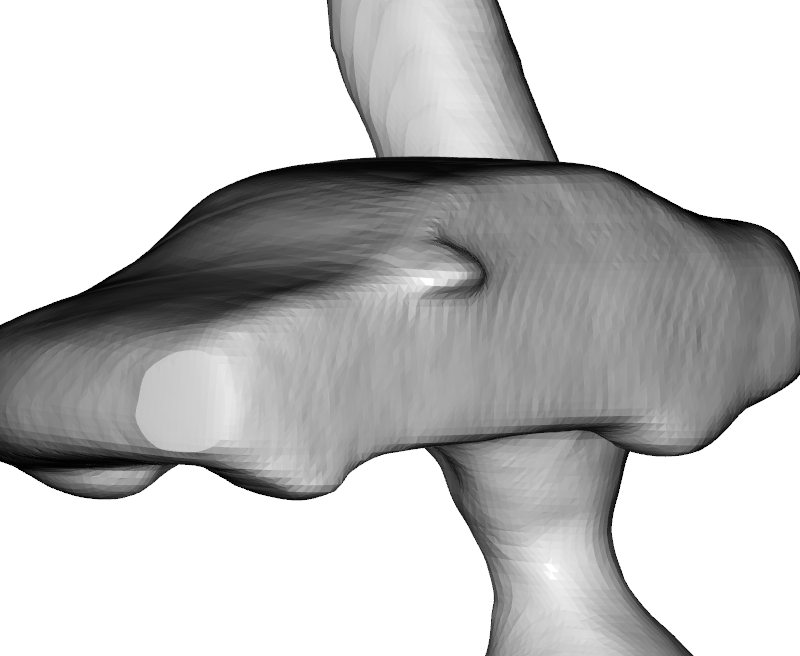} &
    \includegraphics[width=0.2\linewidth]{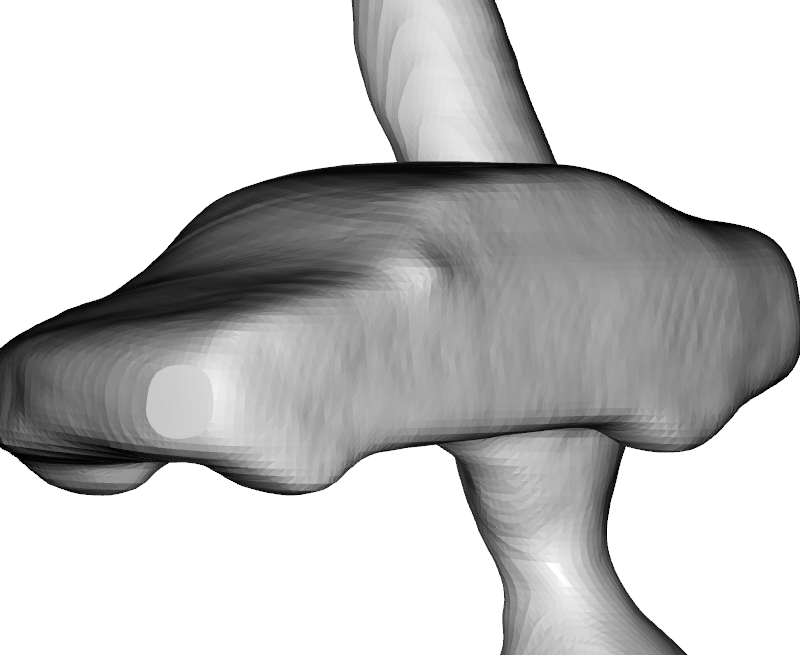}
  \end{tabular}
  \end{center}
\caption{Failure cases of our approach. All models are trained on \textit{multi-class} First row: ONet correctly reconstructs the  leg of the table that is invisible in the input. HPN misses this leg. Second row: difficult view of a couch. ONet correctly reconstructs the invisible arm rest, while HPN does not. Third row: detailed reconstructions, like the mirror in the local reconstruction, can disappear due to global aggregation in HPN.}
\label{fig:failure}
\end{figure}

Fig.~\ref{fig:failure} shows some failure cases. Since the local reconstruction emphasizes the visible areas more, transfer of the global layout from examples seen during training is less pronounced than with the purely global baseline. Conversely, some details reconstructed correctly with the purely local network can be washed out due to the aggregation with the more global hierarchy levels.

\subsection{Real data}
In order to verify that our conclusions are not limited to the case of perfect ground truth input depth maps, we run the evaluation on selected depth maps from the ScanNet~\cite{dai2017scannet} dataset.
Both the baseline network and our approach were only trained in the synthetic multi-class setting.

\begin{figure}
  \begin{center}
    \begin{tabular}{c c c}
    Input & ONet & HPN\\
    \includegraphics[width=0.3\linewidth]{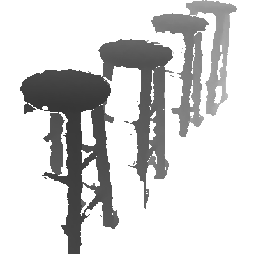} &
    \includegraphics[width=0.3\linewidth]{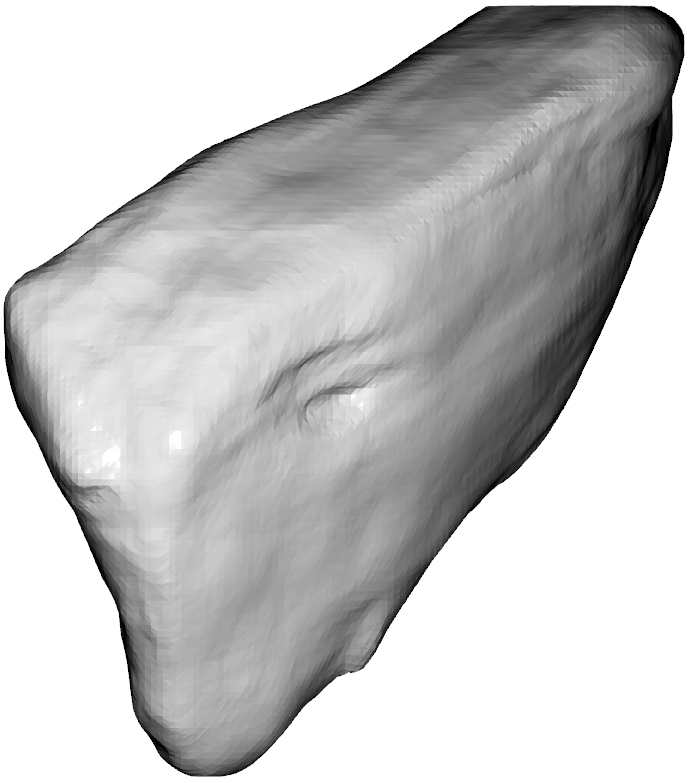} &
    \includegraphics[width=0.3\linewidth]{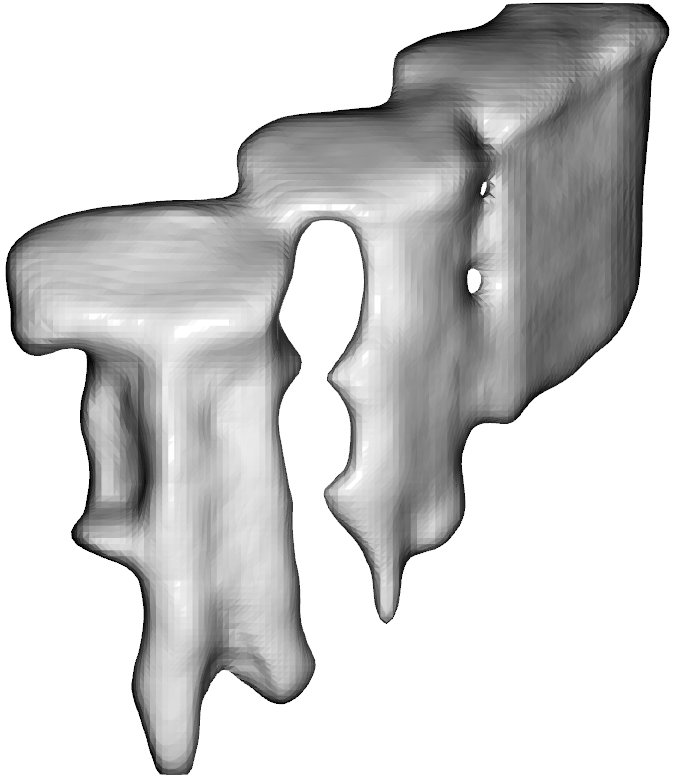} \\
    \includegraphics[width=0.3\linewidth]{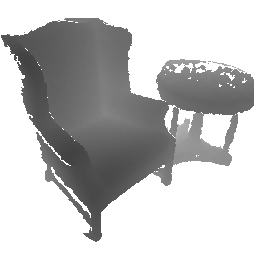} &
    \includegraphics[width=0.3\linewidth]{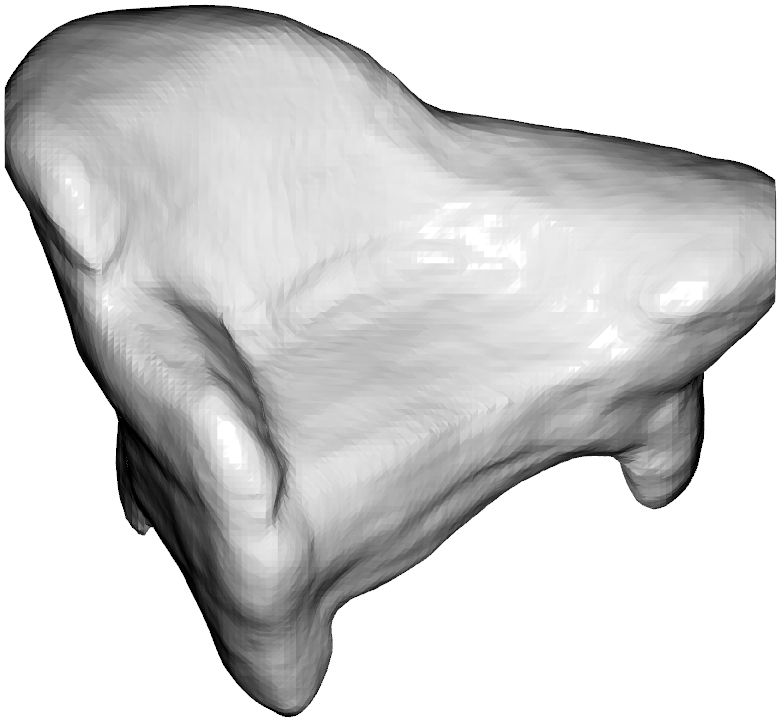} &
    \includegraphics[width=0.3\linewidth]{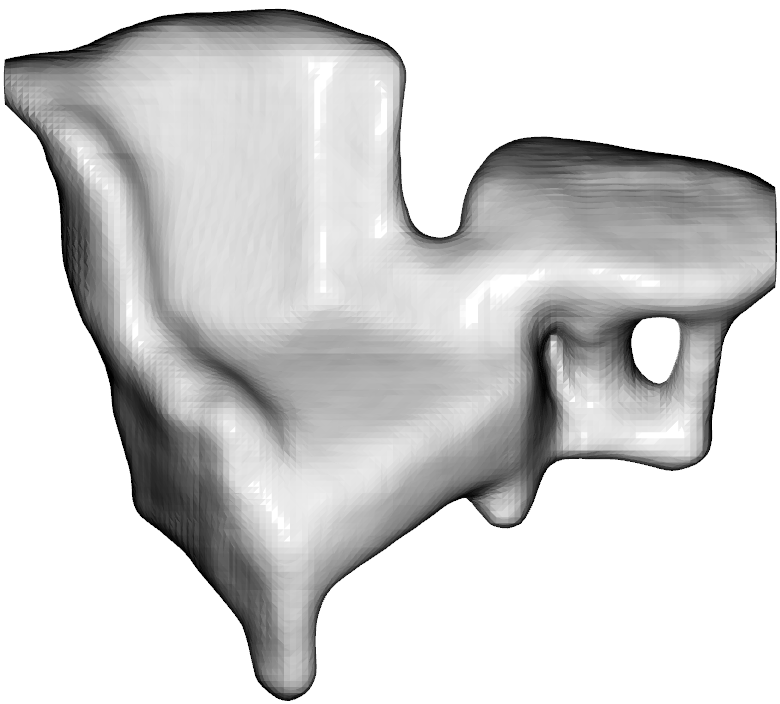} \\
    \includegraphics[width=0.18\linewidth]{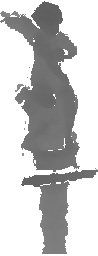} &
    \includegraphics[width=0.18\linewidth]{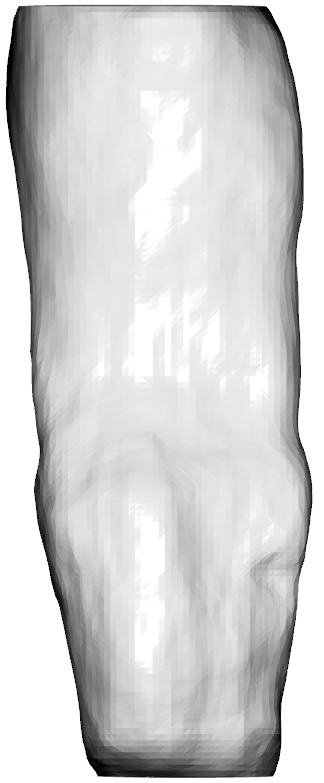} &
    \includegraphics[width=0.18\linewidth]{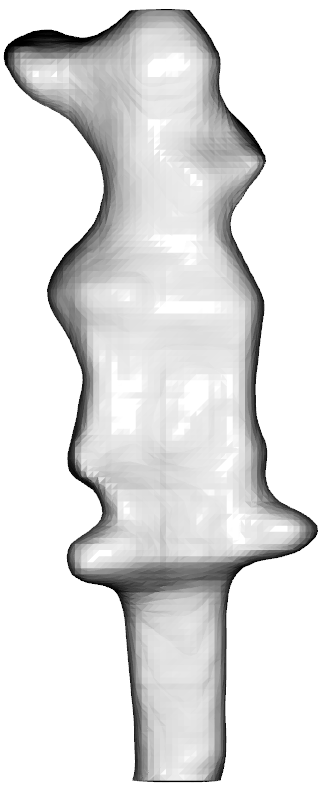} \\
   \end{tabular}
  \end{center}
\caption{When provided with real-world noisy depth images from ScanNet, our multi-class-trained HPN yields more detailed reconstructions than the ONet baseline.}
\label{fig:real_data}
\end{figure}

Fig.~\ref{fig:real_data} shows that HPN produces reasonable results both in case of multiple objects per scene and in case of a complex novel object from the statue class, although it has never seen noisy real-world data during training.
In contrast, the ONet baseline only captures global blob-like structures.

%% file: tables/main_results.tex
\begin{table*}[tb!]
\centering
\resizebox*{\hsize}{!}{
	\setlength{\tabcolsep}{2mm}
	\begin{tabular}{l c
	c >{\columncolor{lightgray}}c 
	c >{\columncolor{lightgray}}c 
	c >{\columncolor{lightgray}}c 
	c >{\columncolor{lightgray}}c 
	c >{\columncolor{lightgray}}c 
	|| c >{\columncolor{lightgray}}c 
	|| c >{\columncolor{lightgray}}c} 
	\toprule
	& &	
	\multicolumn{2}{c}{Chair} &
	\multicolumn{2}{c}{Lamp} &
	\multicolumn{2}{c}{Speaker} &
	\multicolumn{2}{c}{Sofa} &
	\multicolumn{2}{c||}{Table} &
	\multicolumn{2}{c||}{Mean (unseen)} & \multicolumn{2}{c}{Composition}\\
	& &
	F$\uparrow$ & CD$\downarrow$ & 
	F$\uparrow$ & CD$\downarrow$ & 
	F$\uparrow$ & CD$\downarrow$ & 
	F$\uparrow$ & CD$\downarrow$ & 
	F$\uparrow$ & CD$\downarrow$ & 
	F$\uparrow$ & CD$\downarrow$ & 
	F$\uparrow$ & CD$\downarrow$\\ 
	\midrule
	\multirow{6}{*}{\rotatebox{90}{\textit{plane,car,chair}}} &
	ONet~\cite{mescheder2019occupancy} &
	\textcolor{blue_our}{40.8} & \textcolor{blue_our}{4.1} &
	18.8 & 9.3 &
	38.4 & 6.0 & 
	43.2 & 4.7 &
	35.2 & 5.3 &
	\textcolor{orange_our}{29.3} & \textcolor{orange_our}{6.8} &
	\textcolor{green_our}{18.3} & \textcolor{green_our}{8.7}\\
	& ONet-SDF~\cite{mescheder2019occupancy} &
	\textcolor{blue_our}{35.9} & \textcolor{blue_our}{4.6} &
	19.9 & 8.5 &
	37.6 & 5.8 &
	38.3 & 5.1 &
	33.0 & 5.6 &
	\textcolor{orange_our}{28.6} & \textcolor{orange_our}{6.6} &
	\textcolor{green_our}{19.3} & \textcolor{green_our}{8.0}\\
	& GenRe~\cite{zhang2018learninggenre} &
	\textcolor{blue_our}{-} & \textcolor{blue_our}{-} &
	- & 6.0* &
	- & 7.7* &
	- & 5.9* &
	- & 5.7* &
	\textcolor{orange_our}{-} & \textcolor{orange_our}{5.7}* &
	\textcolor{green_our}{-} & \textcolor{green_our}{-}\\
	& LDIF$_{svim1d}$~\cite{genova2020local} & 
	\textcolor{blue_our}{\textbf{62.1}} & \textcolor{blue_our}{\textbf{0.9}} &
	20.8 & 9.4 & 
	22.9 & 5.2 &
	52.7 & \textbf{1.3} &
	33.0 & \textbf{3.3} &
	\textcolor{orange_our}{32.1} & \textcolor{orange_our}{\textbf{3.5}} &
	\textcolor{green_our}{16.4} & \textcolor{green_our}{10.9}\\
	\cmidrule{2-16} 
	& HPN (ours) &
	\textcolor{blue_our}{44.3} & \textcolor{blue_our}{3.8} &
	38.4 & 4.8 &
	\textbf{49.7} & \textbf{4.8} &
	46.6 & 4.5 &
	43.7 & 4.4 &
	\textcolor{orange_our}{42.9} & \textcolor{orange_our}{4.9} &
	\textcolor{green_our}{30.2} & \textcolor{green_our}{5.7}\\
	& HPN-SDF (ours) &
	\textcolor{blue_our}{53.6} & \textcolor{blue_our}{3.3} &
	\textbf{56.5} & \textbf{3.5} &
	49.4 & 5.0 &
	\textbf{54.4} & 3.9 &
	\textbf{53.1} & 3.7 &
	\textcolor{orange_our}{\textbf{48.2}} &
	\textcolor{orange_our}{4.6} &
	\textcolor{green_our}{\textbf{42.4}} &
	\textcolor{green_our}{\textbf{3.9}}\\
	\midrule
	\midrule
	\multirow{5}{*}{\rotatebox{90}{\textit{chair}}} &
	ONet~\cite{mescheder2019occupancy} &
	\textcolor{blue_our}{36.2} & \textcolor{blue_our}{4.6} &
	16.6 & 10.3 &
	34.2 & 6.5 &
	35.3 & 5.4 &
	31.6 & 5.9 &
	\textcolor{orange_our}{24.3} & \textcolor{orange_our}{7.9} &
	\textcolor{green_our}{16.5} & \textcolor{green_our}{9.3}\\
	& LDIF$_{svim1d}$~\cite{genova2020local} &
	\textcolor{blue_our}{\textbf{59.2}} & \textcolor{blue_our}{\textbf{1.0}} &
    17.8 & 10.6 &
	21.6 & 5.6 &
	\textbf{44.4} & \textbf{1.4} &
	31.4 & \textbf{3.9} &
	\textcolor{orange_our}{27.7} & \textcolor{orange_our}{\textbf{4.2}} &
	\textcolor{green_our}{14.9} & \textcolor{green_our}{13.0} \\
	& HPN (ours) &
	\textcolor{blue_our}{43.0} & \textcolor{blue_our}{3.9} &
	40.2 & 4.6 &
	48.6 & \textbf{4.8} &
	\textbf{44.4} & 4.6 &
	\textbf{44.2} & 4.3 &
	\textcolor{orange_our}{43.1} &
	\textcolor{orange_our}{4.7} &
	\textcolor{green_our}{31.2} &
	\textcolor{green_our}{5.3}\\
	& HPN-SDF (ours) &
	\textcolor{blue_our}{41.2} & \textcolor{blue_our}{4.2} &
    \textbf{43.6} & \textbf{4.3} &
	\textbf{48.8} & 5.0 &
	43.8 & 5.0 &
	\textbf{44.2} & 4.5 &
	\textcolor{orange_our}{\textbf{45.3}} & \textcolor{orange_our}{4.7} &
	\textcolor{green_our}{\textbf{31.7}} & \textcolor{green_our}{\textbf{5.2}} \\
	\midrule
	\multirow{5}{*}{\rotatebox{90}{\textit{lamp}}} &
	ONet~\cite{mescheder2019occupancy} &
	20.4 & 8.1 &
	\textcolor{blue_our}{42.0} & \textcolor{blue_our}{4.7} &
	37.8 & 5.6 &
	24.2 & 7.2 &
	29.1 & 7.1 &
	\textcolor{orange_our}{26.8} & \textcolor{orange_our}{6.8} &
	\textcolor{green_our}{18.1} & \textcolor{green_our}{8.5}\\
	& LDIF$_{svim1d}$~\cite{genova2020local} & 
	12.4 & 12.2 &
	\textcolor{blue_our}{48.1} & \textcolor{blue_our}{\textbf{2.5}} &
	21.6 & 5.1 &
	11.8 & 7.4 &
	17.1 & 10.5 &
	\textcolor{orange_our}{21.1} & \textcolor{orange_our}{5.6} &
	\textcolor{green_our}{12.5} & \textcolor{green_our}{14.0}\\
	& HPN (ours) &
	\textbf{42.4} & \textbf{4.7} &
	\textcolor{blue_our}{\textbf{50.3}} & \textcolor{blue_our}{3.6} &
	\textbf{53.2} & \textbf{4.6} &
	\textbf{45.2} & \textbf{5.0} &
	\textbf{47.1} & \textbf{4.7} &
	\textcolor{orange_our}{\textbf{47.1}} &
	\textcolor{orange_our}{\textbf{4.7}} &
	\textcolor{green_our}{\textbf{35.8}} &
	\textcolor{green_our}{\textbf{5.0}}\\
	& HPN-SDF (ours) &
	41.1 & 4.8 &
    \textcolor{blue_our}{48.4} & \textcolor{blue_our}{3.6} &
	51.5 & \textbf{4.6} &
	44.7 & \textbf{5.0} &
	44.8 & 4.8 &
	\textcolor{orange_our}{46.1} & \textcolor{orange_our}{4.8} &
	\textcolor{green_our}{33.9} & \textcolor{green_our}{5.2} \\
	\bottomrule
	\end{tabular}
}
\vspace{1mm}
\caption{Comparison of the hierarchical prior network (HPN) to the state of the art in terms of generalization. The top part of the table shows training in the \textit{multi-class} setting, the lower part shows training on a single class. We report two metrics: F-score (F, shown in \%) and \colorbox{lightgray}{Chamfer distance} (CD, multiplied by 100 for better readability). * denotes results taken from the original paper. \textit{svim1d} denotes \cite{genova2020local}'s data generation - a single perspective ground truth depth map. \textcolor{blue_our}{Results on categories seen during training are marked in blue}. \textcolor{orange_our}{\emph{Mean (unseen)} shows the average of per-class scores over all 13 unseen categories.} \textcolor{green_our}{\emph{Composition} shows results on the composition of two objects per image.} On \textcolor{green_our}{compositions}, HPN is more than twice as accurate as the state of the art and generally better on \textcolor{orange_our}{unseen classes}, while LDIF is better on \textcolor{blue_our}{seen classes}. See supplemental~\ref{sec:apx_quantitative} for more results. Best viewed in color.}
\label{tbl:results}
\vspace{3mm}
\end{table*}

%% file: tables/analysis_results.tex
\begin{table}[ht]
\centering
\resizebox*{\linewidth}{!}{
	\setlength{\tabcolsep}{2mm}
	\ra{1}
	\begin{tabular}{@{} l @{\ \ \ } l @{\ \ \ } c  >{\columncolor{lightgray}} c c >{\columncolor{lightgray}} c c >{\columncolor{lightgray}}c }
	\toprule
	& & \multicolumn{2}{c}{Full} & \multicolumn{2}{c}{Visible} & \multicolumn{2}{c}{Invisible} \\
	& & F$\uparrow$ & CD$\downarrow$ & F$\uparrow$ & CD$\downarrow$ & F$\uparrow$ & CD$\downarrow$ \\
	\midrule
	\multirow{6}{*}{\rotatebox{90}{\textit{chair}}} &
	Global@256 (ONet) & 23.3 & 7.7 & 28.6 & 6.4 & 22.7 & 7.7 \\
	& Local@128 & 38.5 & 4.9 & 57.8 & 2.8 & 27.1 & 6.0 \\
	& Local@64 & 42.0 & 4.9 & 67.3 & 2.2 & 27.3 & 6.4 \\
	& Local@32 & 37.5 & 5.8 & 54.7 & 2.9 & 28.1 & 7.4 \\
	& Local@16 & 36.6 & 6.7 & 57.8 & 3.2 & 24.5 & 8.7 \\
	& HPN@(256+32) & 35.7 & 5.2 & 48.8 & 3.3 & 28.3 & 6.2 \\
	& HPN@(256+64) & 38.3 & 4.9 & 56.7 & 2.8 & 27.4 & 6.1 \\
	& HPN@(256+64+32) & 39.7 & 4.7 & 57.6 & 2.6 & 29.8 & 5.7 \\
	& HPN@(256+128+64+32+16) & 42.0 & 4.4 & 61.7 & 2.4 & 30.4 & 5.6 \\ 
	\bottomrule
	\end{tabular}
}
\vspace{1mm}
\caption{Mean F-score (F)
and \colorbox{lightgray}{Chamfer distance (CD)}
for multiple hierarchy levels trained on the category chair and evaluated on all unseen categories. We report the F-score and the CD for the full shape (Full), the visible and the invisible parts of it.}
\label{tbl:analysis_results}
\vspace{3mm}
\end{table}

%% file: 06_conclusion.tex
\section{Conclusion}
\label{sec:conclusion}
In this paper, we introduced a new paradigm for learning single-view reconstruction priors based on multiple locality levels. The decisive advantage of this paradigm over previous global reconstruction approaches is its ability to recombine local shapes. This recombination not only makes much more efficient use of training data, it also enables the generalization to completely unseen shapes or configurations of objects, which has been the key limitation of single-view object reconstruction to-date. Technically, the presented approach is simple yet flexible. While we used Occupancy Networks, any other network that implements an implicit function (even a retrieval approach) could replace that architecture. In this sense, the proposed hierarchy of local networks should not be regarded as an isolated network architecture, but rather as a working principle.
\\ \\
\noindent \textbf{Acknowledgements:} We thank Philipp Schr\"oppel and Evgeny Levinkov for feedback on the manuscript. We also thank Philipp for his help with the infrastructure.

%% file: 11_appendix.tex
\title{Fostering Generalization in Single-view 3D Reconstruction by Learning a Hierarchy of Local and Global Shape Priors \\ - Supplementary Material -}
\maketitle
\date{} 
\author{}

\input{X_qualitative}

\section{Quantitative}
\label{sec:apx_quantitative}
In Tbl.~\ref{tbl:apx_results}, we present the full version of Tbl.~\ref{tbl:results} from the main paper. This contains the full set of evaluation classes (columns). Additionally, we provide the mean for all categories seen during training in the first column. Furthermore, we also report results for networks trained in the \textit{single-class} airplane setting. Best scores are marked in \textbf{bold}. Our method is always best in the generalization settings (black, orange and green numbers) and best for one of the training classes (blue numbers).
In Tbl.~\ref{tbl:apx_results_iou} we report the IoU values for completeness.
\input{X_quantitative_genre13}
\input{X_quantitative_iou}

\section{Local retrieval}

It was shown that single-view reconstruction with shape retrieval is competitive with network approaches~\cite{tatarchenko2019single}. 
The principle of recombination, enabled by the local parts, is also compatible with retrieval. Instead of a local reconstruction network, could  we also use local retrieval for reconstruction? 

\begin{figure}
  \begin{center}
  \begin{tabular}{c c c} 
    \includegraphics[width=0.25\linewidth]{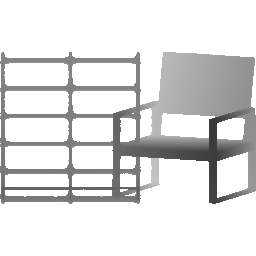} &
    \includegraphics[width=0.21\linewidth]{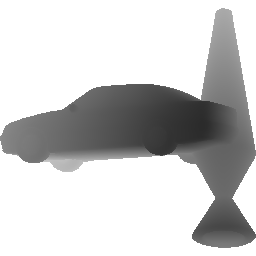} &    \includegraphics[width=0.25\linewidth]{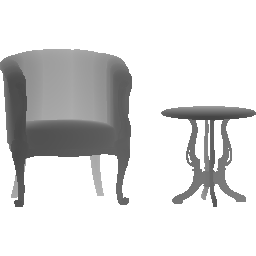}\\ 
    \includegraphics[width=0.25\linewidth]{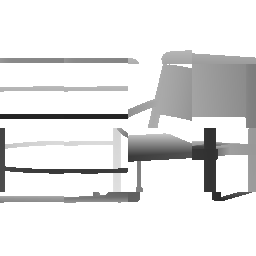} &
    \includegraphics[width=0.21\linewidth]{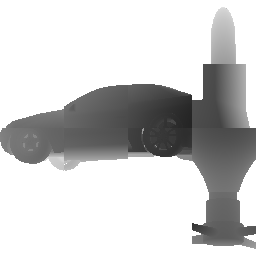} &    \includegraphics[width=0.25\linewidth]{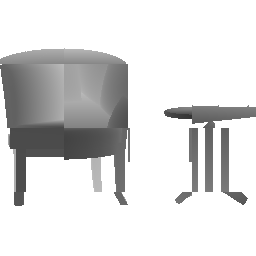}\\ 
    \includegraphics[width=0.25\linewidth]{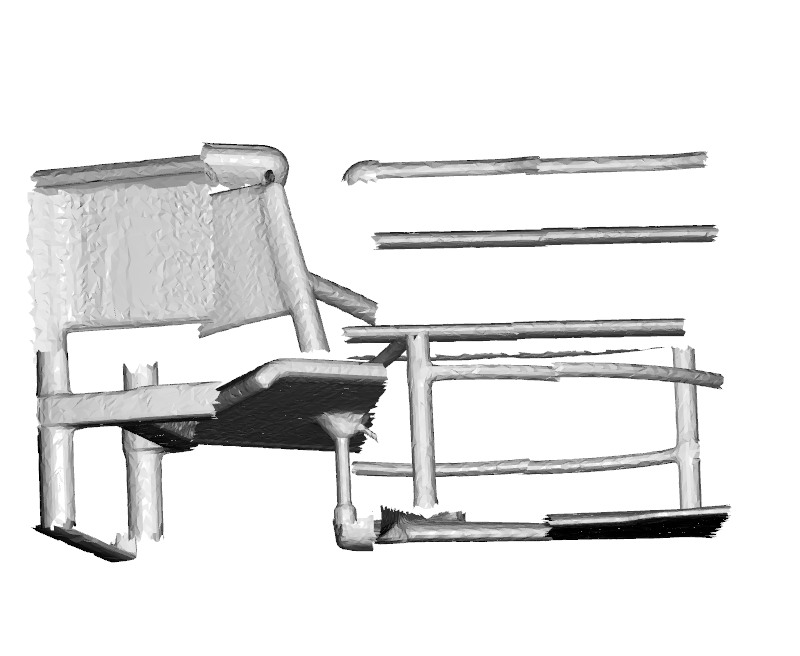} &    \includegraphics[width=0.25\linewidth]{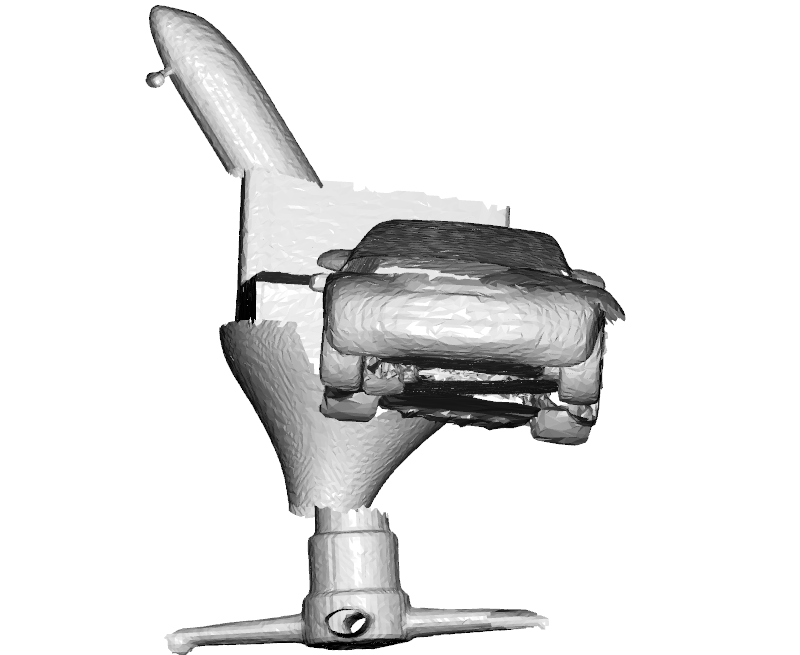} &    \includegraphics[width=0.25\linewidth]{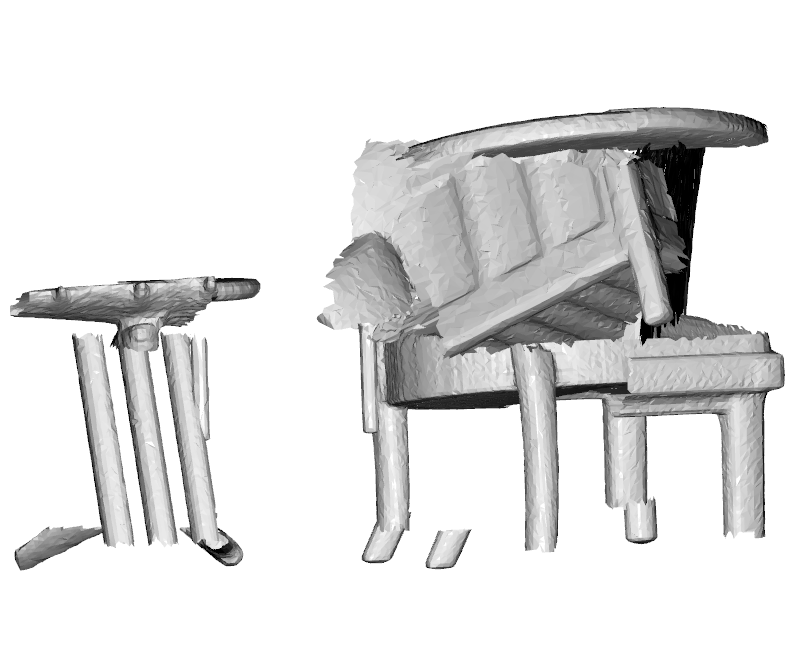} \\ 
    \includegraphics[width=0.25\linewidth]{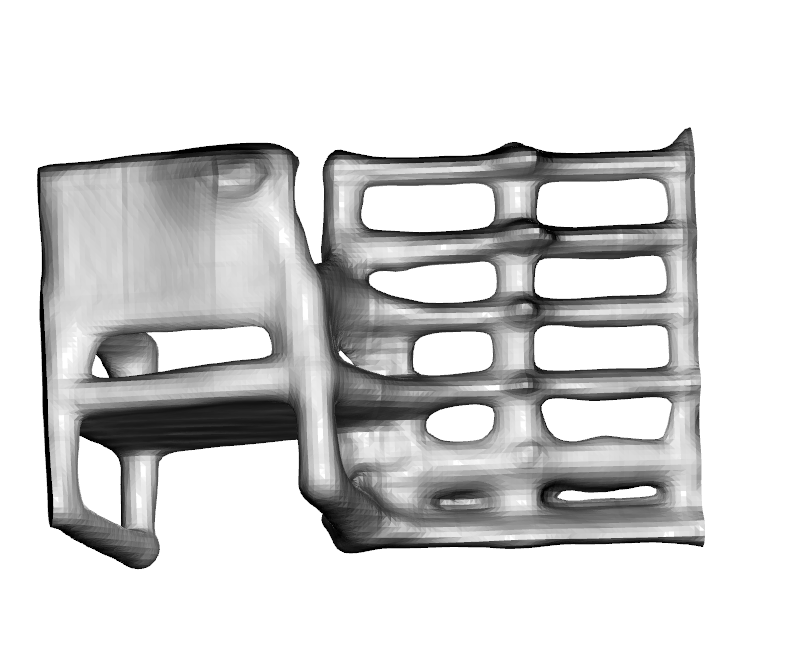} & 
    \includegraphics[width=0.25\linewidth]{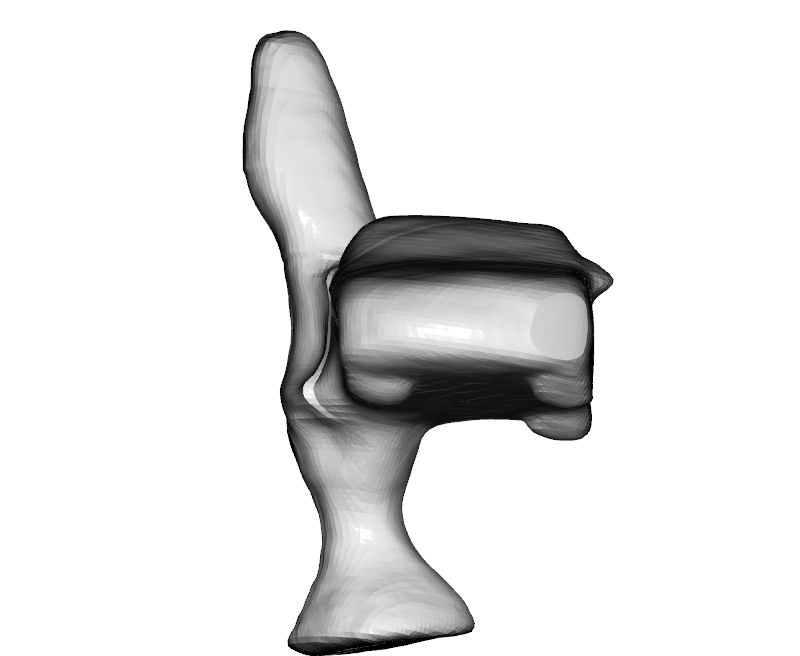} &
    \includegraphics[width=0.25\linewidth]{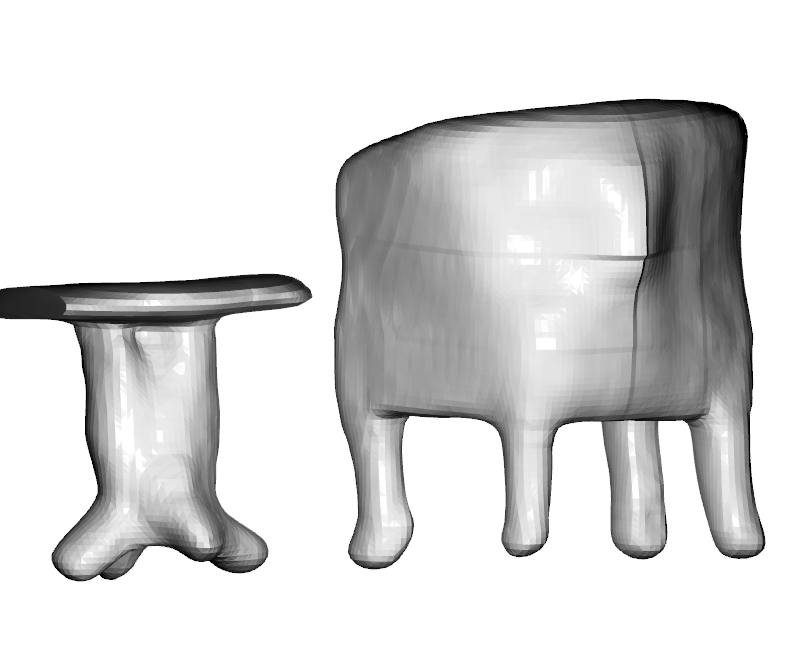}
  \end{tabular}
  \end{center}
\caption{Comparison to a retrieval baseline which locally retrieves the nearest neighbors for depth patches of size $N=64$. \textbf{First row:} Test image. \textbf{Second row:} Image assembled from nearest neighbor patches from the training set. \textbf{Third row:} Reconstruction from the nearest neighbors (opposite viewpoint). \textbf{Fourth row:} Reconstruction from our Local@64 network. 
}
\label{fig:retrieval}
\end{figure}

Fig.~\ref{fig:retrieval} shows a study for patches of size $N=64$. For each patch in the test image (first row), we retrieved the nearest neighbor patch by absolute $L_1$ distance from the multi-class training set. The resulting approximated test image (second row) shows that the silhouette of the nearest neighbors agrees well with the target image, especially for the car and the table. We cropped and assembled the corresponding 3D parts from the ground truth mesh to obtain the reconstruction (third row). The result is roughly right. However, compared to the Local@64 network (forth row), the reconstructed shape is not smooth and shows some strange artifacts. 
Another advantage of networks is the fast inference time with $\sim 4$ seconds per shape versus 3 hours for a naive nearest neighbor search over all parts of the training set. 

That said, the non-smoothness and the runtime could both be mitigated with a more sophisticated retrieval approach. This shows: the key concept to enable generalization in single-view reconstruction across object categories is not a particular choice of network but the recombination and aggregation of local parts. The local retrieval counterpart to our network implementation is a viable alternative, even though the network version is probably more elegant.

%% file: X_qualitative.tex
\section{Qualitative Examples}
\label{sec:apx_qualitative}

Here, we present more qualitative results to illustrate different behaviours of ONet\cite{mescheder2019occupancy} and our Hierarchical Prior Network (HPN). All shapes created by HPN are much closer to the ground truth shape.
For each of the four training settings from Tbl.~\ref{tbl:apx_results} we show one figure. Fig.~\ref{fig:apx_airplane} for \textit{airplane}, Fig.~\ref{fig:apx_lamp} for \textit{lamp}, Fig.~\ref{fig:apx_chair} for \textit{chair} and  Fig.~\ref{fig:apx_multiclass} for \textit{multi-class}. Each figure contains twelve examples. For each sample we show the input image (first row), reconstructions for ONet (second row), reconstructions for Ours (third row) and the ground truth (fourth row) in two views.
The figures are best viewed in PDF, such that one can zoom in.

\input{appendix/airplane/0_plane}

\input{appendix/lamp/0_lamp_figure}

\input{appendix/chair/0_chair_figure}

\input{appendix/multiclass/0_multiclass_figure}

%% file: appendix/airplane/0_plane.tex
\begin{figure*}[ht]
  \renewcommand{\arraystretch}{3}
  \centering
    \begin{tabular}{m{0.5em} c c | c c | c c | c c}
    \raisebox{-0.5\height}{\rotatebox{90}{Input}} & \multicolumn{2}{c}{\raisebox{-0.5\height}{\includegraphics[height=0.075\linewidth]{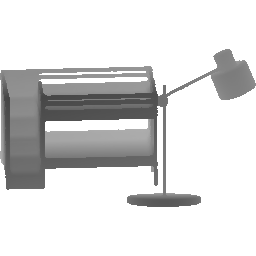}}} &
    \multicolumn{2}{c}{\raisebox{-0.5\height}{\includegraphics[height=0.075\linewidth]{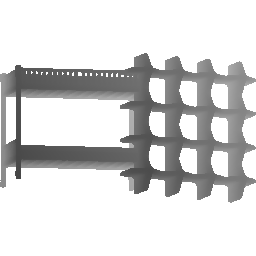}}} &
    \multicolumn{2}{c}{\raisebox{-0.5\height}{\includegraphics[height=0.075\linewidth]{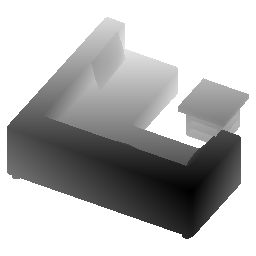}}} &
    \multicolumn{2}{c}{\raisebox{-0.5\height}{\includegraphics[height=0.075\linewidth]{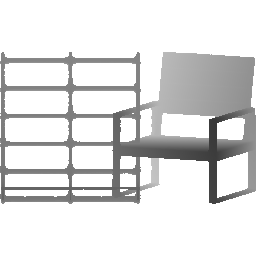}}} \\
    \raisebox{-0.5\height}{\rotatebox{90}{ONet}} &
    \raisebox{-0.5\height}{\includegraphics[height=0.075\linewidth]{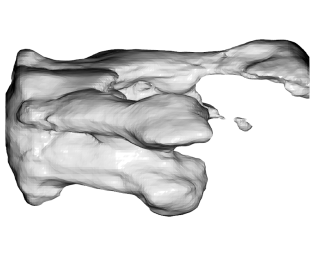}} &
    \raisebox{-0.5\height}{\includegraphics[height=0.075\linewidth]{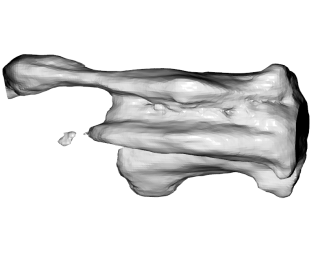}} &
    \raisebox{-0.5\height}{\includegraphics[height=0.075\linewidth]{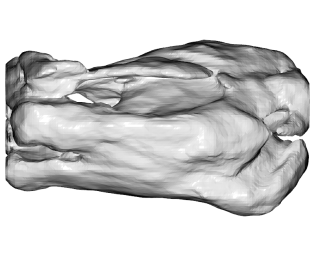}} &
    \raisebox{-0.5\height}{\includegraphics[height=0.075\linewidth]{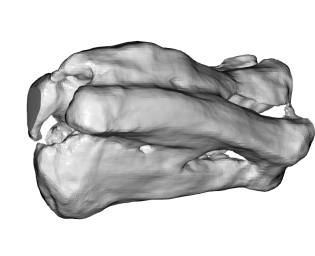}} &
    \raisebox{-0.5\height}{\includegraphics[height=0.075\linewidth]{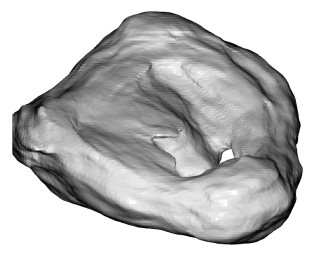}} &
    \raisebox{-0.5\height}{\includegraphics[height=0.075\linewidth]{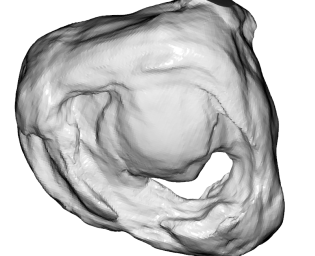}} &
    \raisebox{-0.5\height}{\includegraphics[height=0.075\linewidth]{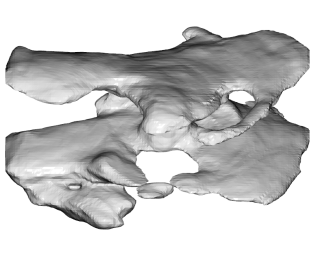}} &
    \raisebox{-0.5\height}{\includegraphics[height=0.075\linewidth]{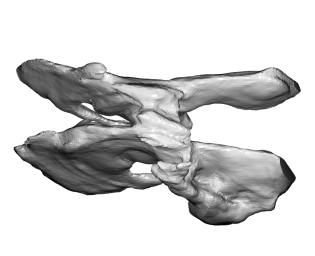}} \\
    \raisebox{-0.5\height}{\rotatebox{90}{HPN}} &
    \raisebox{-0.5\height}{\includegraphics[height=0.075\linewidth]{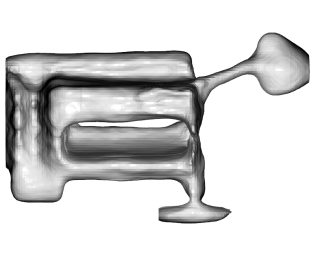}} &
    \raisebox{-0.5\height}{\includegraphics[height=0.075\linewidth]{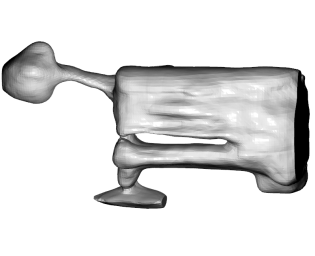}} &
    \raisebox{-0.5\height}{\includegraphics[height=0.075\linewidth]{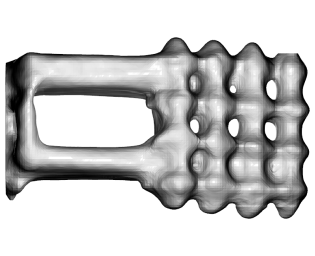}} &
    \raisebox{-0.5\height}{\includegraphics[height=0.075\linewidth]{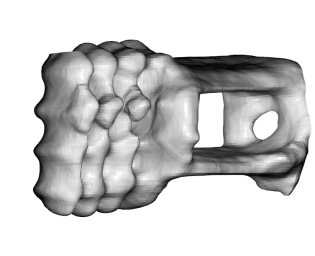}} &
    \raisebox{-0.5\height}{\includegraphics[height=0.075\linewidth]{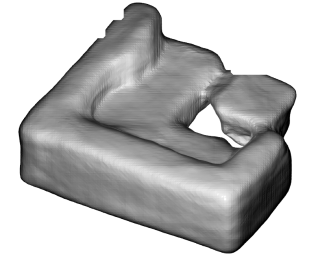}} &
    \raisebox{-0.5\height}{\includegraphics[height=0.075\linewidth]{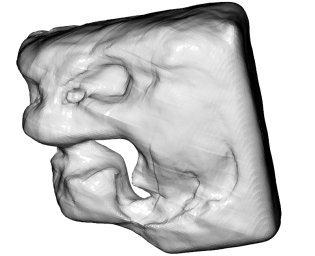}} &
    \raisebox{-0.5\height}{\includegraphics[height=0.075\linewidth]{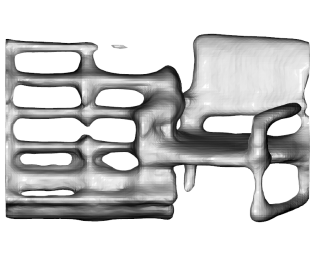}} &
    \raisebox{-0.5\height}{\includegraphics[height=0.075\linewidth]{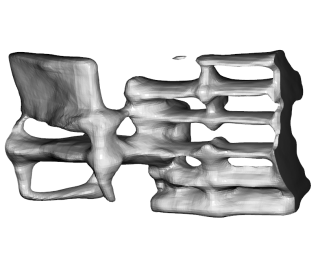}} \\
    \raisebox{-0.5\height}{\rotatebox{90}{GT}} &
    \raisebox{-0.5\height}{\includegraphics[height=0.075\linewidth]{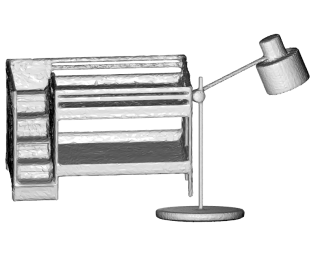}} &
    \raisebox{-0.5\height}{\includegraphics[height=0.075\linewidth]{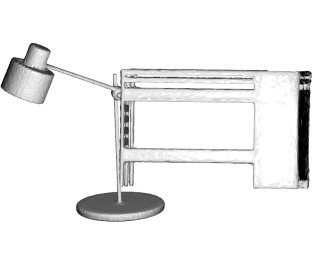}} &
    \raisebox{-0.5\height}{\includegraphics[height=0.075\linewidth]{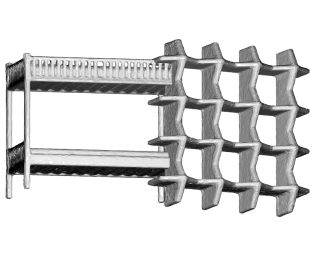}} &
    \raisebox{-0.5\height}{\includegraphics[height=0.075\linewidth]{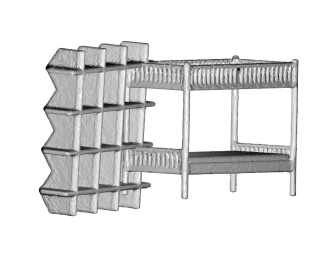}} &
    \raisebox{-0.5\height}{\includegraphics[height=0.075\linewidth]{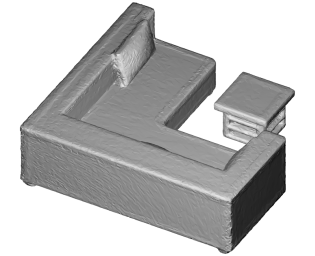}} &
    \raisebox{-0.5\height}{\includegraphics[height=0.075\linewidth]{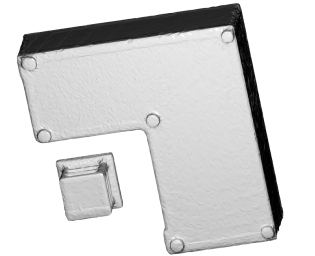}} &
    \raisebox{-0.5\height}{\includegraphics[height=0.075\linewidth]{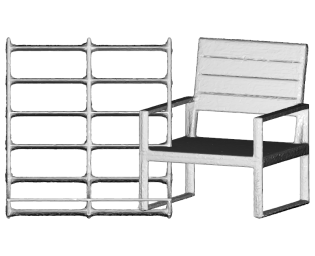}} &
    \raisebox{-0.5\height}{\includegraphics[height=0.075\linewidth]{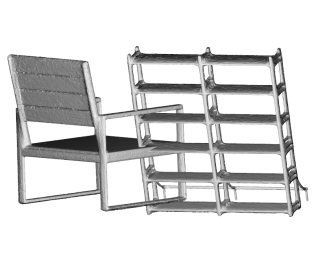}} \\
    \multicolumn{8}{c}{}\\ 
    \raisebox{-0.5\height}{\rotatebox{90}{Input}} & \multicolumn{2}{c}{\raisebox{-0.5\height}{\includegraphics[height=0.075\linewidth]{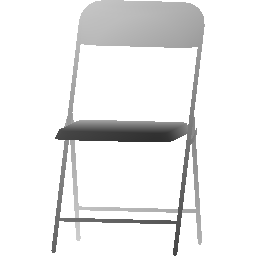}}} &
    \multicolumn{2}{c}{\raisebox{-0.5\height}{\includegraphics[height=0.075\linewidth]{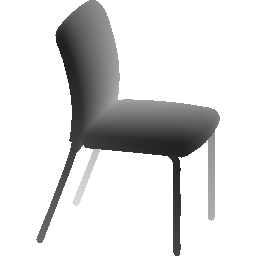}}} &
    \multicolumn{2}{c}{\raisebox{-0.5\height}{\hspace{-1em}\includegraphics[height=0.075\linewidth]{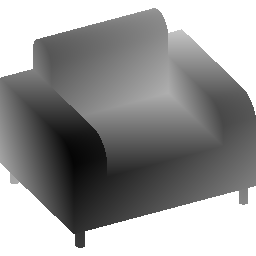}}} &
    \multicolumn{2}{c}{\raisebox{-0.5\height}{\includegraphics[height=0.075\linewidth]{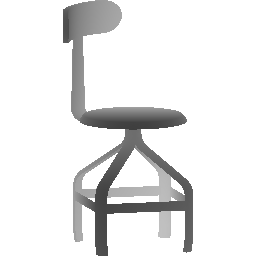}}} \\
    \raisebox{-0.5\height}{\rotatebox{90}{ONet}} &
    \raisebox{-0.5\height}{\includegraphics[height=0.075\linewidth]{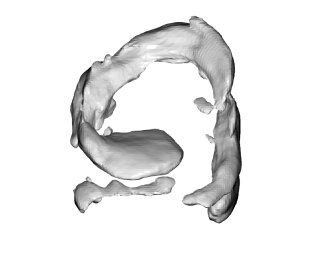}} &
    \raisebox{-0.5\height}{\includegraphics[height=0.075\linewidth]{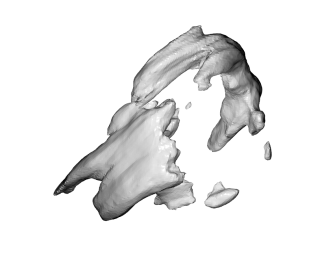}} &
    \raisebox{-0.5\height}{\includegraphics[height=0.075\linewidth]{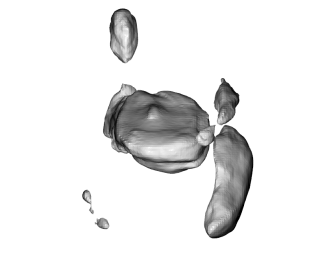}} &
    \raisebox{-0.5\height}{\includegraphics[height=0.075\linewidth]{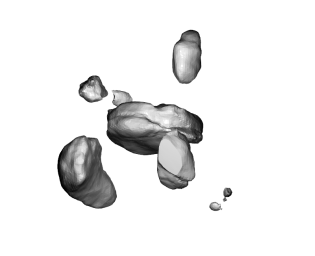}} &
    \raisebox{-0.5\height}{\includegraphics[height=0.075\linewidth]{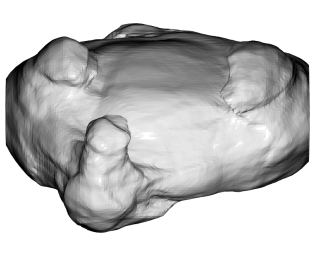}} &
    \hspace{-1em}
    \raisebox{-0.5\height}{\includegraphics[height=0.075\linewidth]{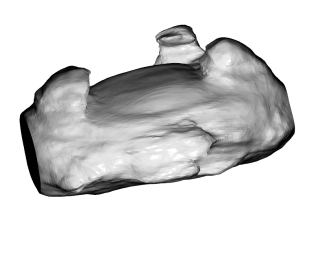}} &
    \raisebox{-0.5\height}{\includegraphics[height=0.075\linewidth]{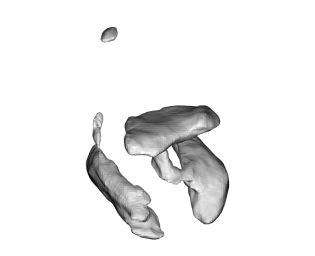}} &
    \hspace{-1em}
    \raisebox{-0.5\height}{\includegraphics[height=0.075\linewidth]{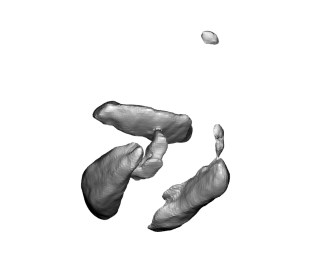}} \\
    \raisebox{-0.5\height}{\rotatebox{90}{HPN}} &
    \raisebox{-0.5\height}{\includegraphics[height=0.075\linewidth]{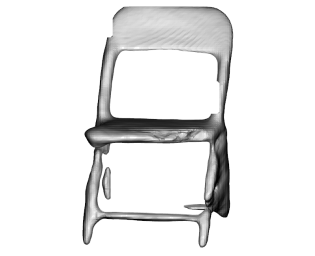}} &
    \raisebox{-0.5\height}{\includegraphics[height=0.075\linewidth]{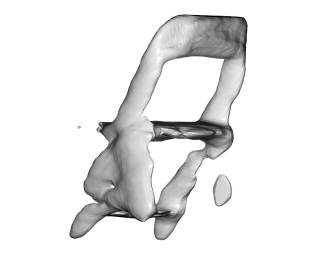}} &
    \raisebox{-0.5\height}{\includegraphics[height=0.075\linewidth]{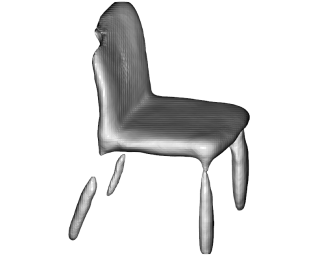}} &
    \raisebox{-0.5\height}{\includegraphics[height=0.075\linewidth]{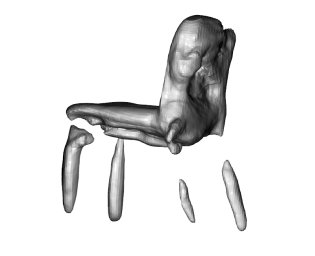}} &
    \raisebox{-0.5\height}{\includegraphics[height=0.075\linewidth]{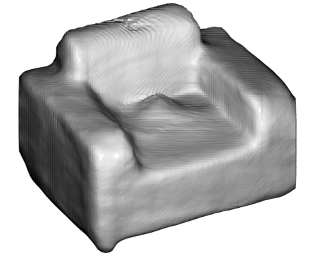}} &
    \raisebox{-0.5\height}{\includegraphics[height=0.075\linewidth]{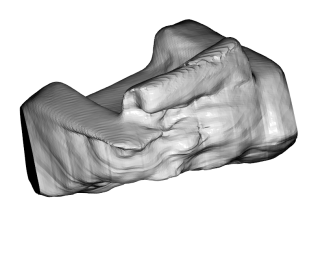}} &
    \raisebox{-0.5\height}{\includegraphics[height=0.075\linewidth]{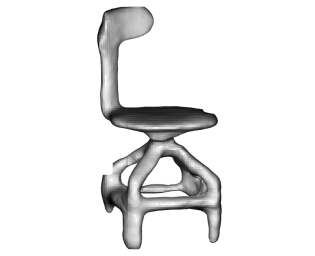}} &
    \raisebox{-0.5\height}{\includegraphics[height=0.075\linewidth]{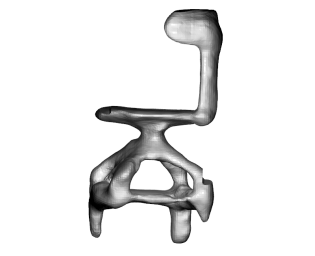}} \\
    \raisebox{-0.5\height}{\rotatebox{90}{GT}} &
    \raisebox{-0.5\height}{\includegraphics[height=0.075\linewidth]{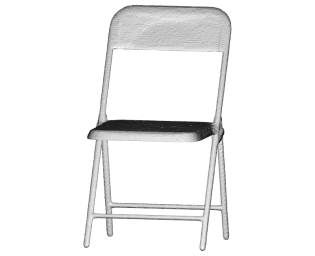}} &
    \raisebox{-0.5\height}{\includegraphics[height=0.075\linewidth]{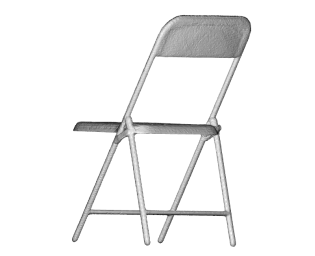}} &
    \raisebox{-0.5\height}{\includegraphics[height=0.075\linewidth]{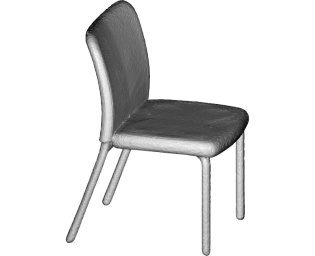}} &
    \raisebox{-0.5\height}{\includegraphics[height=0.075\linewidth]{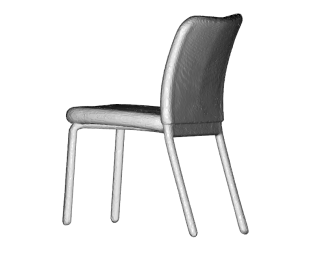}} &
    \raisebox{-0.5\height}{\includegraphics[height=0.075\linewidth]{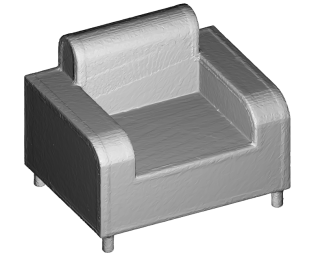}} &
    \raisebox{-0.5\height}{\includegraphics[height=0.075\linewidth]{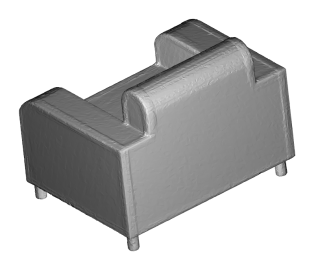}} &
    \raisebox{-0.5\height}{\includegraphics[height=0.075\linewidth]{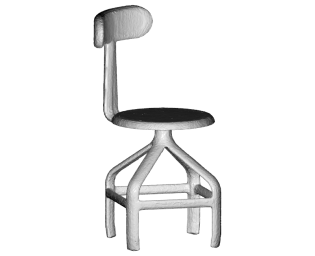}} &
    \raisebox{-0.5\height}{\includegraphics[height=0.075\linewidth]{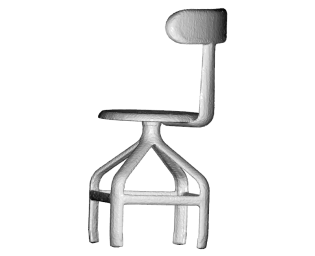}}\\
    \multicolumn{8}{c}{}\\ 
    \raisebox{-0.5\height}{\rotatebox{90}{Input}} &
    \multicolumn{2}{c}{\raisebox{-0.5\height}{\includegraphics[height=0.075\linewidth]{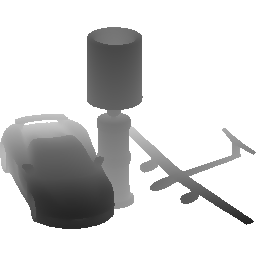}}} & 
    \multicolumn{2}{c}{\raisebox{-0.5\height}{\includegraphics[height=0.075\linewidth]{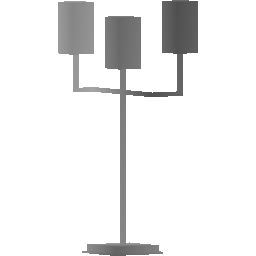}}} &
    \multicolumn{2}{c}{\raisebox{-0.5\height}{\includegraphics[height=0.075\linewidth]{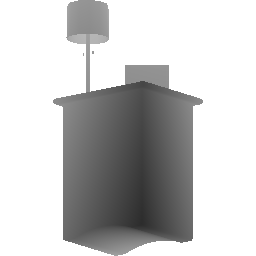}}} & 
    \multicolumn{2}{c}{\raisebox{-0.5\height}{\includegraphics[height=0.075\linewidth]{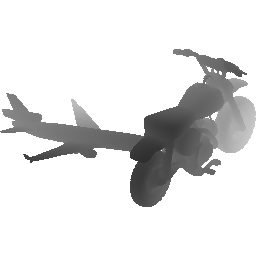}}} \\
    \raisebox{-0.5\height}{\rotatebox{90}{ONet}} &
    \raisebox{-0.5\height}{\includegraphics[height=0.075\linewidth]{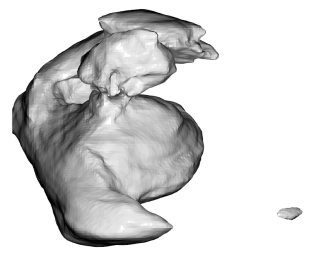}} &
    \raisebox{-0.5\height}{\includegraphics[height=0.075\linewidth]{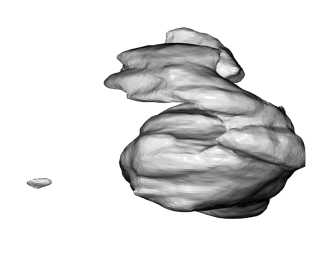}} &
    \raisebox{-0.5\height}{\includegraphics[height=0.075\linewidth]{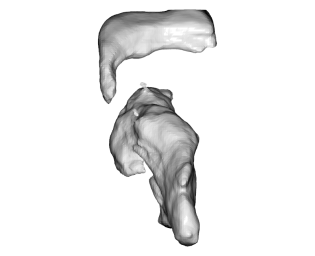}} &
    \raisebox{-0.5\height}{\includegraphics[height=0.075\linewidth]{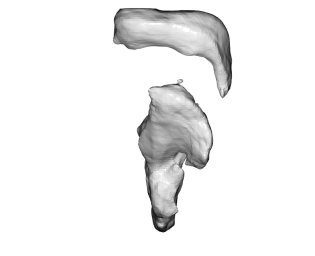}} &
    \raisebox{-0.5\height}{\includegraphics[height=0.075\linewidth]{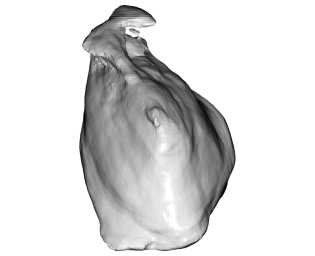}} &
    \raisebox{-0.5\height}{\includegraphics[height=0.075\linewidth]{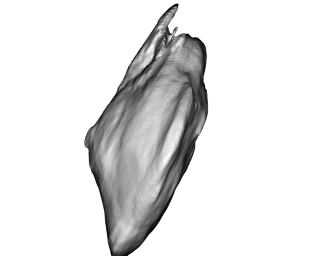}} &
    \raisebox{-0.5\height}{\includegraphics[height=0.075\linewidth]{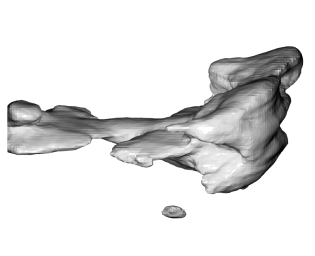}} &
    \raisebox{-0.5\height}{\includegraphics[height=0.075\linewidth]{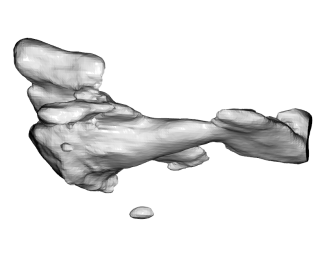}} \\
    \raisebox{-0.5\height}{\rotatebox{90}{HPN}} &
    \raisebox{-0.5\height}{\includegraphics[height=0.075\linewidth]{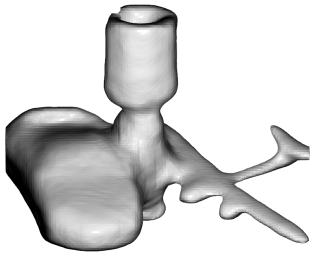}} &
    \raisebox{-0.5\height}{\includegraphics[height=0.075\linewidth]{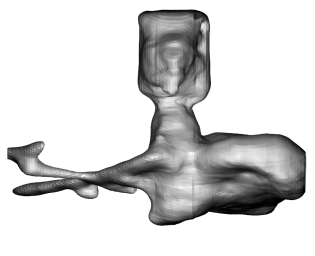}} &
    \raisebox{-0.5\height}{\includegraphics[height=0.075\linewidth]{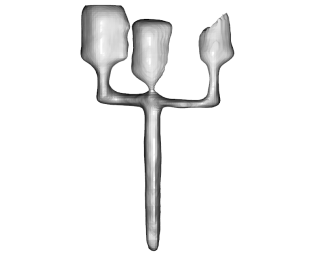}} &
    \raisebox{-0.5\height}{\includegraphics[height=0.075\linewidth]{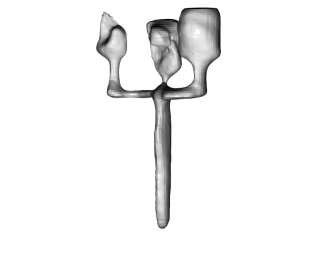}} &
    \raisebox{-0.5\height}{\includegraphics[height=0.075\linewidth]{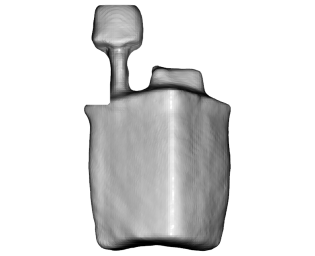}} &
    \raisebox{-0.5\height}{\includegraphics[height=0.075\linewidth]{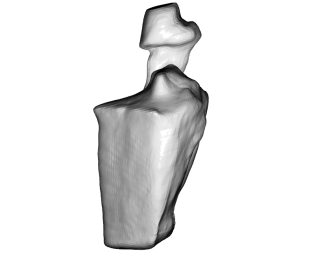}} &
    \raisebox{-0.5\height}{\includegraphics[height=0.075\linewidth]{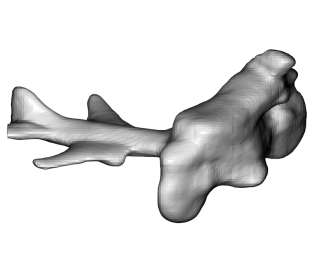}} &
    \raisebox{-0.5\height}{\includegraphics[height=0.075\linewidth]{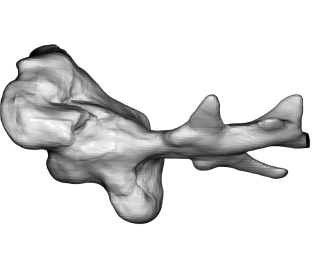}} \\
    \raisebox{-0.5\height}{\rotatebox{90}{GT}} &
    \raisebox{-0.5\height}{\includegraphics[height=0.075\linewidth]{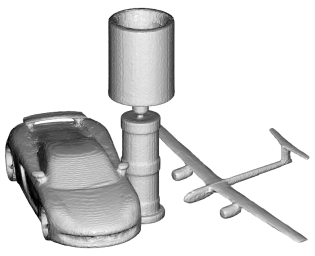}} &
    \raisebox{-0.5\height}{\includegraphics[height=0.075\linewidth]{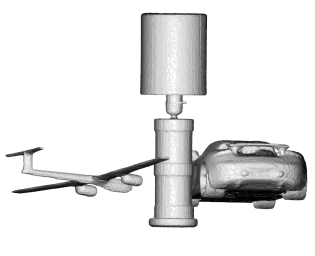}} &
    \raisebox{-0.5\height}{\includegraphics[height=0.075\linewidth]{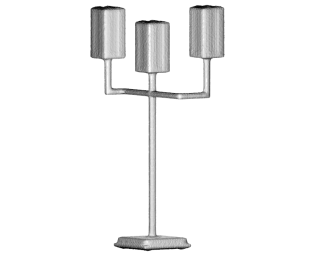}} &
    \raisebox{-0.5\height}{\includegraphics[height=0.075\linewidth]{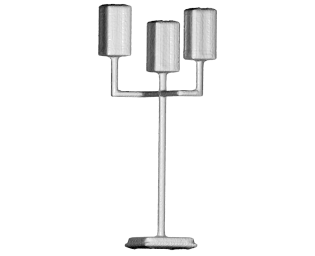}} &
    \raisebox{-0.5\height}{\includegraphics[height=0.075\linewidth]{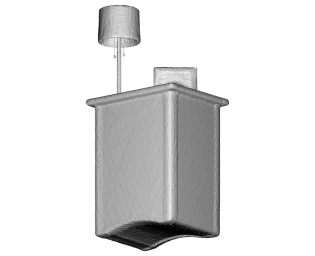}} &
    \raisebox{-0.5\height}{\includegraphics[height=0.075\linewidth]{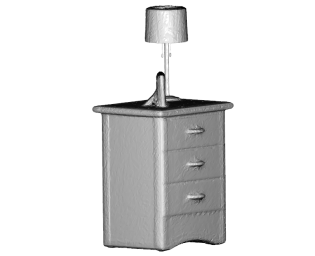}} &
    \raisebox{-0.5\height}{\includegraphics[height=0.075\linewidth]{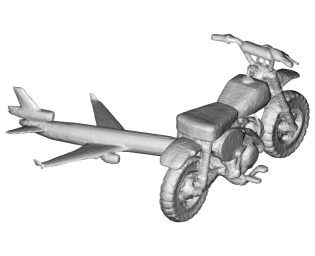}} &
    \raisebox{-0.5\height}{\includegraphics[height=0.075\linewidth]{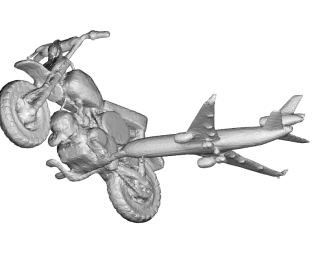}}
    \end{tabular}
\vspace{1.5em}
\caption{Qualitative results for networks trained on the \textit{airplane} class.}
\label{fig:apx_airplane}
\end{figure*}

%% file: appendix/lamp/0_lamp_figure.tex
\begin{figure*}[ht]
\renewcommand{\arraystretch}{3}
  \centering
    \begin{tabular}{m{0.5em} c c | c c | c c | c c}
    \raisebox{-0.5\height}{\rotatebox{90}{Input}} & \multicolumn{2}{c}{\raisebox{-0.5\height}{\includegraphics[height=0.075\linewidth]{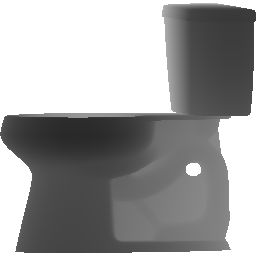}}} &
    \multicolumn{2}{c}{\raisebox{-0.5\height}{\includegraphics[height=0.075\linewidth]{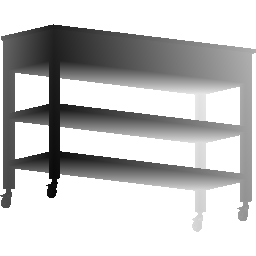}}} &
    \multicolumn{2}{c}{\raisebox{-0.5\height}{\includegraphics[height=0.075\linewidth]{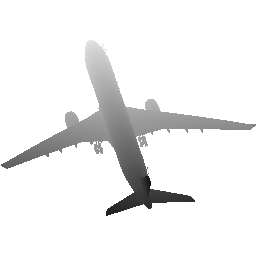}}} &
    \multicolumn{2}{c}{\raisebox{-0.5\height}{\includegraphics[height=0.075\linewidth]{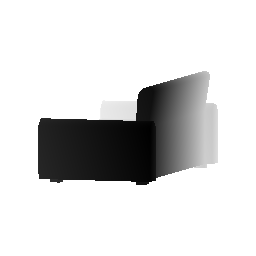}}} \\
    \raisebox{-0.5\height}{\rotatebox{90}{ONet}} &
    \raisebox{-0.5\height}{\includegraphics[height=0.075\linewidth]{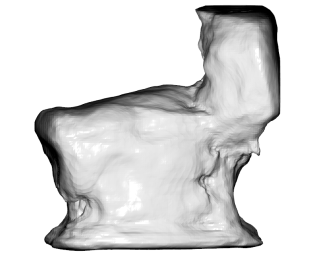}} &
    \raisebox{-0.5\height}{\includegraphics[height=0.075\linewidth]{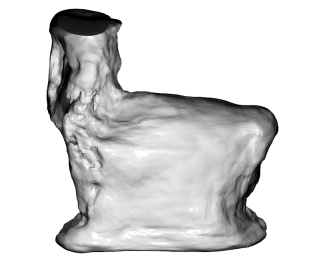}} &
    \raisebox{-0.5\height}{\includegraphics[height=0.075\linewidth]{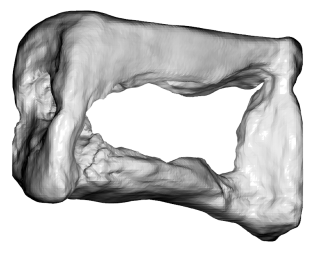}} &
    \raisebox{-0.5\height}{\includegraphics[height=0.075\linewidth]{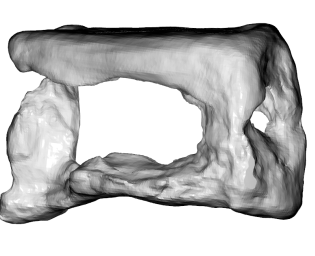}} &
    \raisebox{-0.5\height}{\includegraphics[height=0.075\linewidth]{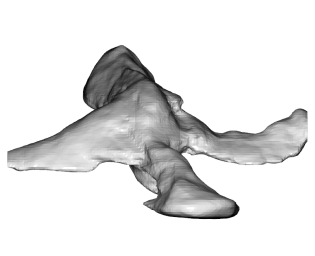}} &
    \raisebox{-0.5\height}{\includegraphics[height=0.075\linewidth]{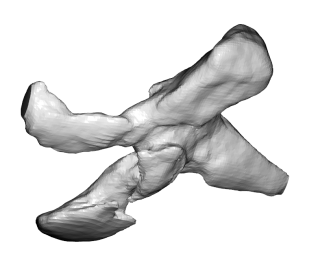}} &
    \raisebox{-0.5\height}{\includegraphics[height=0.075\linewidth]{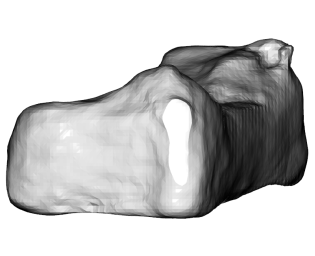}} &
    \raisebox{-0.5\height}{\includegraphics[height=0.075\linewidth]{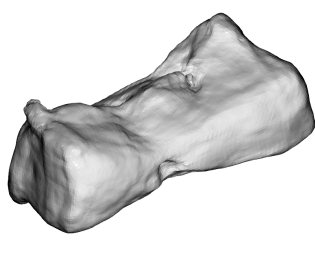}} \\
    \raisebox{-0.5\height}{\rotatebox{90}{HPN}} &
    \raisebox{-0.5\height}{\includegraphics[height=0.075\linewidth]{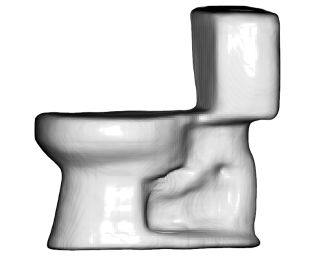}} &
    \raisebox{-0.5\height}{\includegraphics[height=0.075\linewidth]{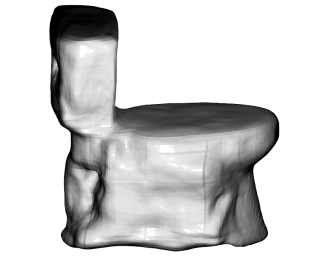}} &
    \raisebox{-0.5\height}{\includegraphics[height=0.075\linewidth]{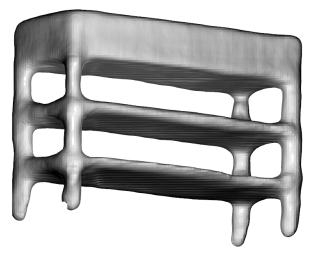}} &
    \raisebox{-0.5\height}{\includegraphics[height=0.075\linewidth]{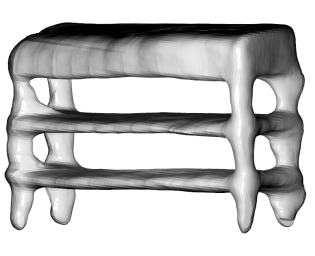}} &
    \raisebox{-0.5\height}{\includegraphics[height=0.075\linewidth]{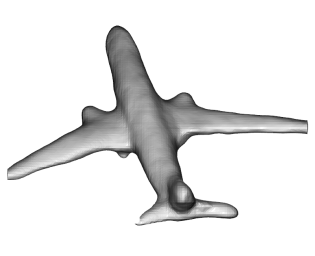}} &
    \raisebox{-0.5\height}{\includegraphics[height=0.075\linewidth]{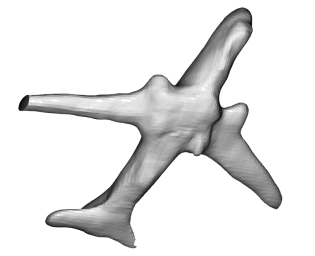}} &
    \raisebox{-0.5\height}{\includegraphics[height=0.075\linewidth]{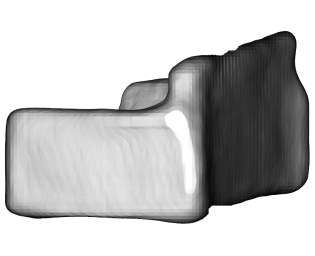}} &
    \raisebox{-0.5\height}{\includegraphics[height=0.075\linewidth]{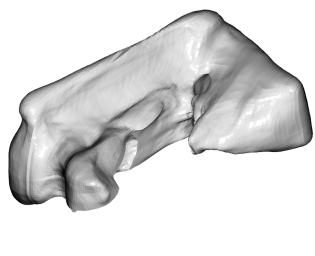}} \\
    \raisebox{-0.5\height}{\rotatebox{90}{GT}} &
    \raisebox{-0.5\height}{\includegraphics[height=0.075\linewidth]{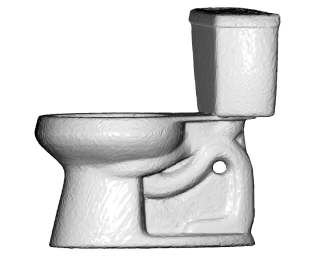}} &
    \raisebox{-0.5\height}{\includegraphics[height=0.075\linewidth]{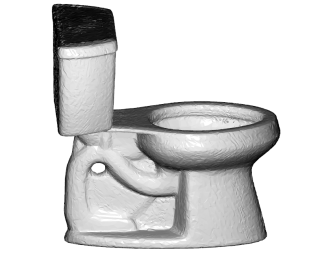}} &
    \raisebox{-0.5\height}{\includegraphics[height=0.075\linewidth]{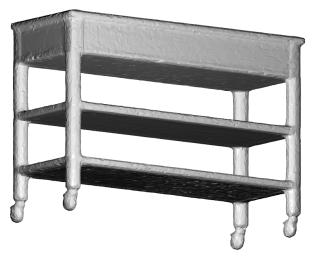}} &
    \raisebox{-0.5\height}{\includegraphics[height=0.075\linewidth]{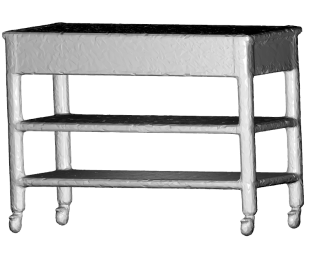}} &
    \raisebox{-0.5\height}{\includegraphics[height=0.075\linewidth]{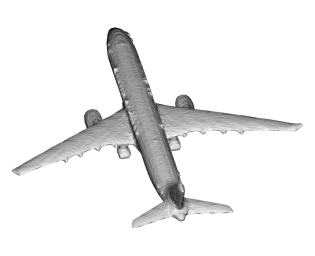}} &
    \raisebox{-0.5\height}{\includegraphics[height=0.075\linewidth]{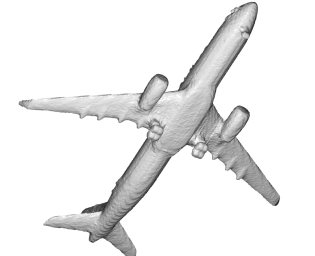}} &
    \raisebox{-0.5\height}{\includegraphics[height=0.075\linewidth]{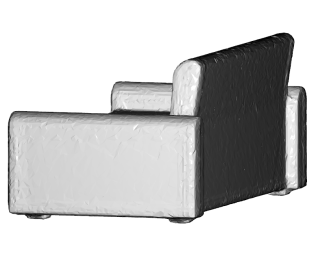}} &
    \raisebox{-0.5\height}{\includegraphics[height=0.075\linewidth]{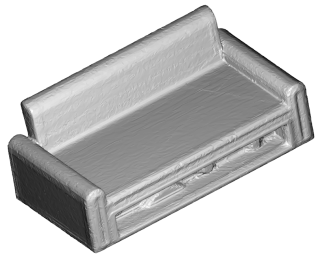}}\\
    \multicolumn{8}{c}{}\\ 
    \raisebox{-0.5\height}{\rotatebox{90}{Input}} & \multicolumn{2}{c}{\raisebox{-0.5\height}{\includegraphics[height=0.075\linewidth]{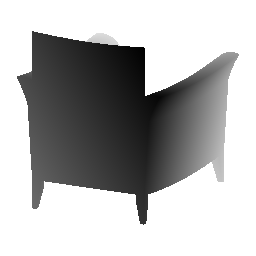}}} &
    \multicolumn{2}{c}{\raisebox{-0.5\height}{\includegraphics[height=0.075\linewidth]{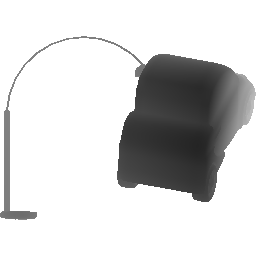}}} &
    \multicolumn{2}{c}{\raisebox{-0.5\height}{\includegraphics[height=0.075\linewidth]{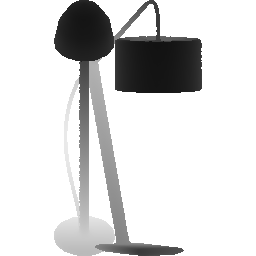}}} &
    \multicolumn{2}{c}{\raisebox{-0.5\height}{\includegraphics[height=0.075\linewidth]{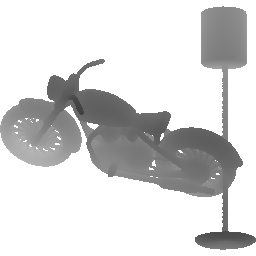}}} \\
    \raisebox{-0.5\height}{\rotatebox{90}{ONet}} &
    \raisebox{-0.5\height}{\includegraphics[height=0.075\linewidth]{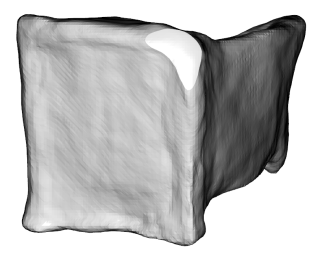}} &
    \raisebox{-0.5\height}{\includegraphics[height=0.075\linewidth]{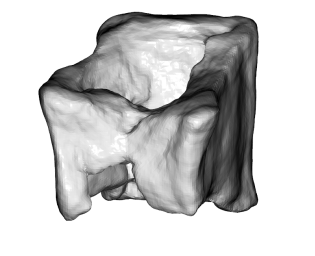}} &
    \raisebox{-0.5\height}{\includegraphics[height=0.075\linewidth]{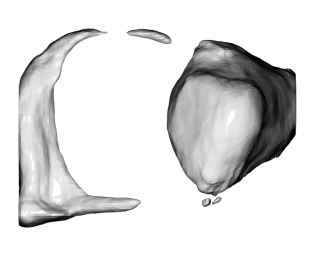}} &
    \raisebox{-0.5\height}{\includegraphics[height=0.075\linewidth]{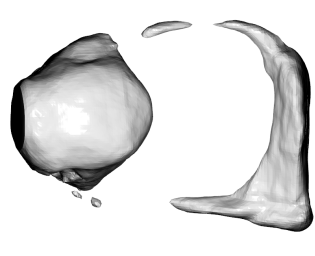}} &
    \raisebox{-0.5\height}{\includegraphics[height=0.075\linewidth]{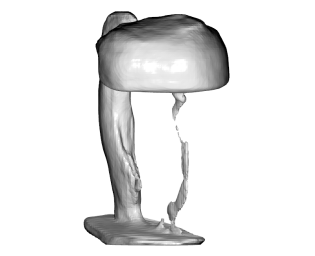}} &
    \raisebox{-0.5\height}{\includegraphics[height=0.075\linewidth]{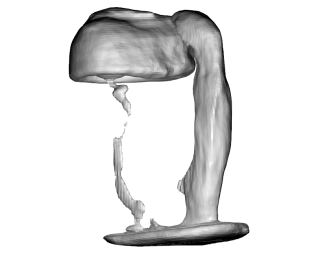}} &
    \raisebox{-0.5\height}{\includegraphics[height=0.075\linewidth]{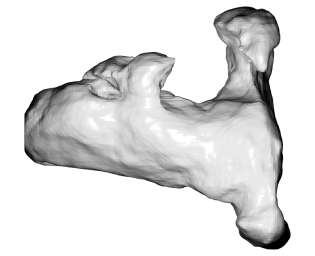}} &
    \raisebox{-0.5\height}{\includegraphics[height=0.075\linewidth]{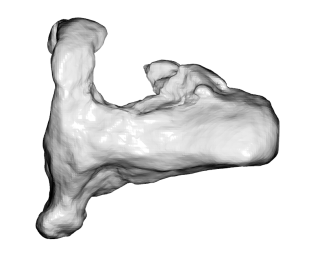}} \\
    \raisebox{-0.5\height}{\rotatebox{90}{HPN}} &
    \raisebox{-0.5\height}{\includegraphics[height=0.075\linewidth]{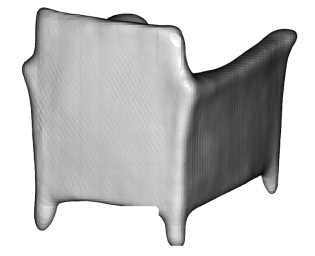}} &
    \raisebox{-0.5\height}{\includegraphics[height=0.075\linewidth]{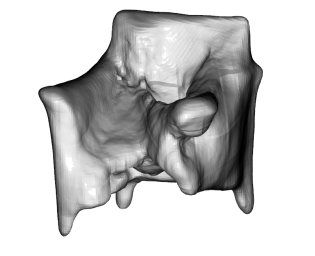}} &
    \raisebox{-0.5\height}{\includegraphics[height=0.075\linewidth]{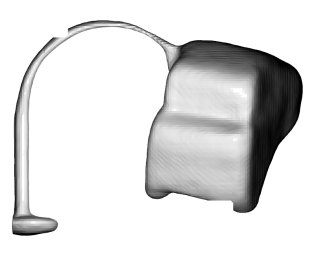}} &
    \raisebox{-0.5\height}{\includegraphics[height=0.075\linewidth]{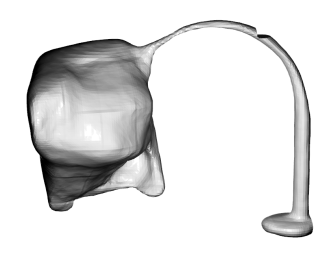}} &
    \raisebox{-0.5\height}{\includegraphics[height=0.075\linewidth]{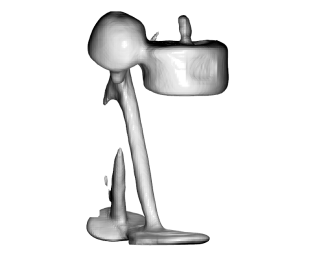}} &
    \raisebox{-0.5\height}{\includegraphics[height=0.075\linewidth]{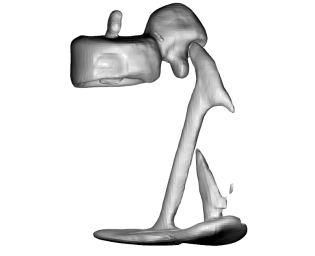}} &
    \raisebox{-0.5\height}{\includegraphics[height=0.075\linewidth]{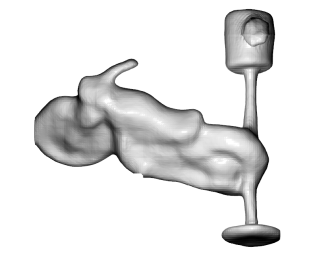}} &
    \raisebox{-0.5\height}{\includegraphics[height=0.075\linewidth]{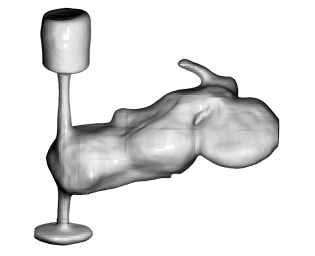}} \\
    \raisebox{-0.5\height}{\rotatebox{90}{GT}} &
    \raisebox{-0.5\height}{\includegraphics[height=0.075\linewidth]{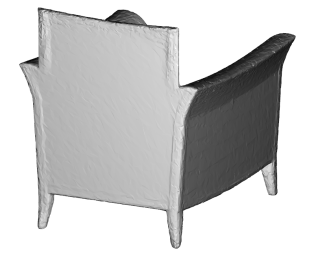}} &
    \raisebox{-0.5\height}{\includegraphics[height=0.075\linewidth]{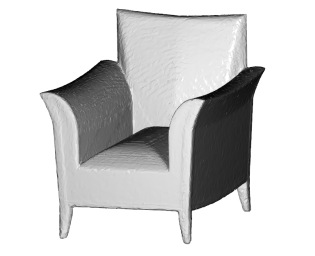}} &
    \raisebox{-0.5\height}{\includegraphics[height=0.075\linewidth]{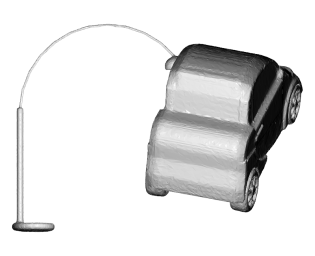}} &
    \raisebox{-0.5\height}{\includegraphics[height=0.075\linewidth]{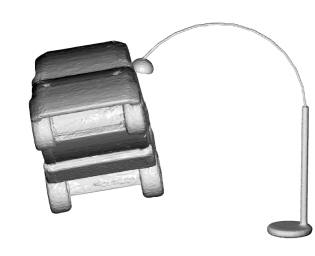}} &
    \raisebox{-0.5\height}{\includegraphics[height=0.075\linewidth]{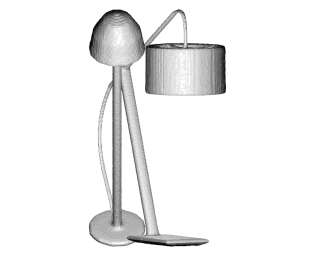}} &
    \raisebox{-0.5\height}{\includegraphics[height=0.075\linewidth]{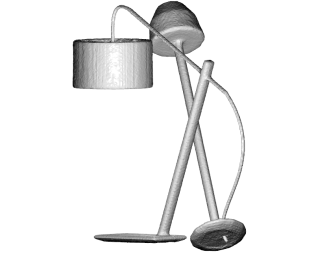}} &
    \raisebox{-0.5\height}{\includegraphics[height=0.075\linewidth]{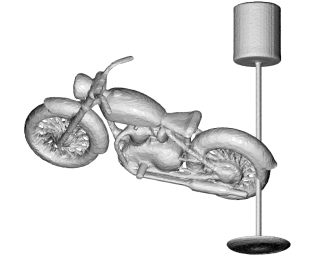}} &
    \raisebox{-0.5\height}{\includegraphics[height=0.075\linewidth]{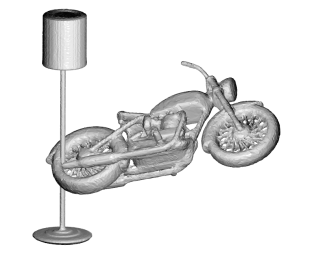}}\\
    \multicolumn{8}{c}{}\\
    \raisebox{-0.5\height}{\rotatebox{90}{Input}} &
    \multicolumn{2}{c}{\raisebox{-0.5\height}{\includegraphics[height=0.075\linewidth]{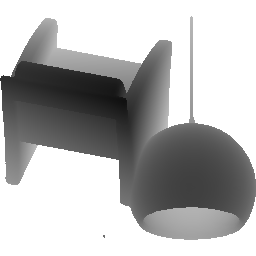}}} &
    \multicolumn{2}{c}{\raisebox{-0.5\height}{\includegraphics[height=0.075\linewidth]{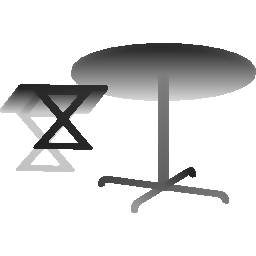}}} &
    \multicolumn{2}{c}{\raisebox{-0.5\height}{\includegraphics[height=0.075\linewidth]{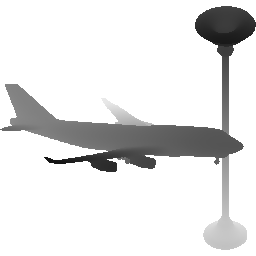}}} &
    \multicolumn{2}{c}{\raisebox{-0.5\height}{\includegraphics[height=0.075\linewidth]{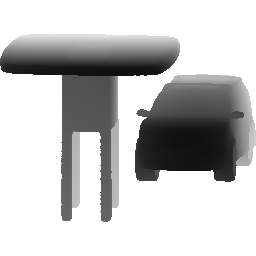}}} \\
    \raisebox{-0.5\height}{\rotatebox{90}{ONet}} &
    \raisebox{-0.5\height}{\includegraphics[height=0.075\linewidth]{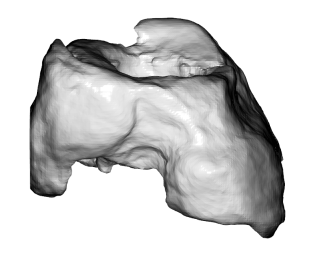}} &
    \raisebox{-0.5\height}{\includegraphics[height=0.075\linewidth]{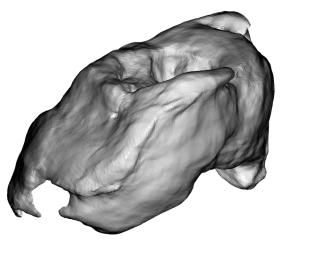}} &
    \raisebox{-0.5\height}{\includegraphics[height=0.075\linewidth]{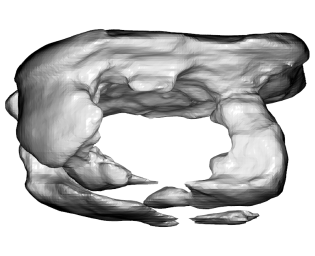}} &
    \raisebox{-0.5\height}{\includegraphics[height=0.075\linewidth]{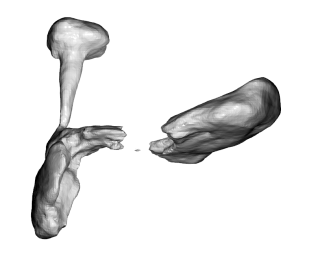}} &
    \raisebox{-0.5\height}{\includegraphics[height=0.075\linewidth]{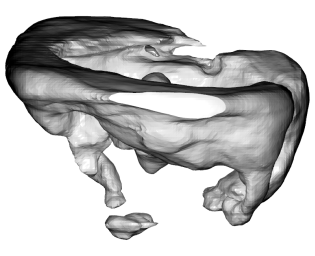}} &
    \raisebox{-0.5\height}{\includegraphics[height=0.075\linewidth]{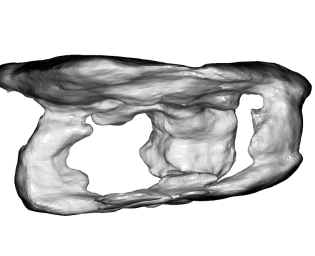}} &
    \raisebox{-0.5\height}{\includegraphics[height=0.075\linewidth]{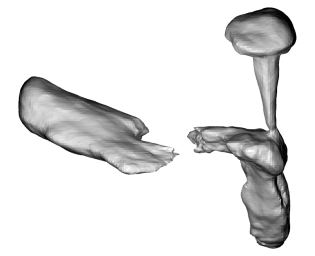}} &
    \raisebox{-0.5\height}{\includegraphics[height=0.075\linewidth]{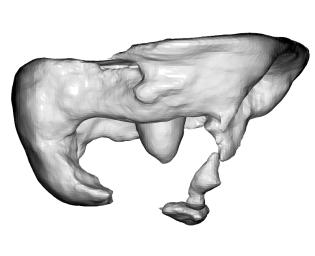}} \\
    \raisebox{-0.5\height}{\rotatebox{90}{HPN}} &
    \raisebox{-0.5\height}{\includegraphics[height=0.075\linewidth]{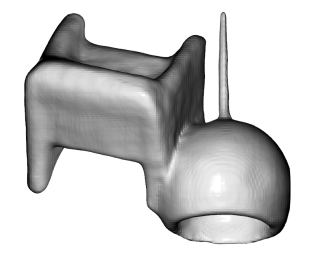}} &
    \raisebox{-0.5\height}{\includegraphics[height=0.075\linewidth]{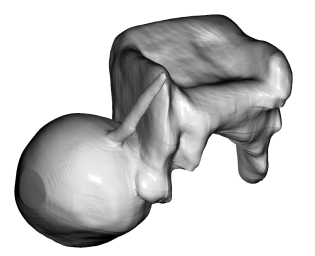}} &
    \raisebox{-0.5\height}{\includegraphics[height=0.075\linewidth]{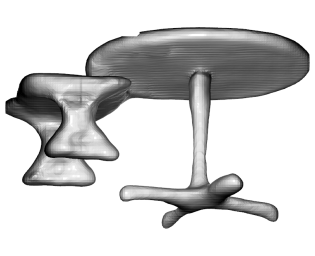}} &
    \raisebox{-0.5\height}{\includegraphics[height=0.075\linewidth]{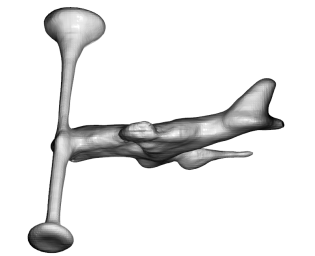}} &
    \raisebox{-0.5\height}{\includegraphics[height=0.075\linewidth]{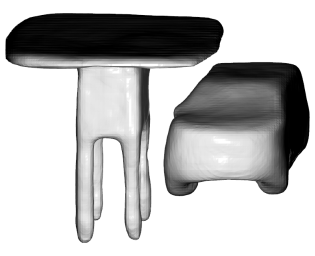}} &
    \raisebox{-0.5\height}{\includegraphics[height=0.075\linewidth]{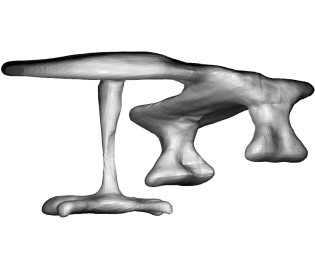}} &
    \raisebox{-0.5\height}{\includegraphics[height=0.075\linewidth]{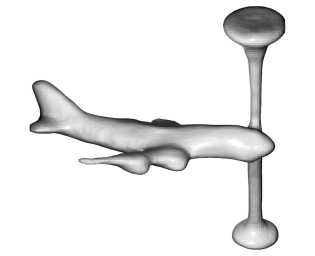}} &
    \raisebox{-0.5\height}{\includegraphics[height=0.075\linewidth]{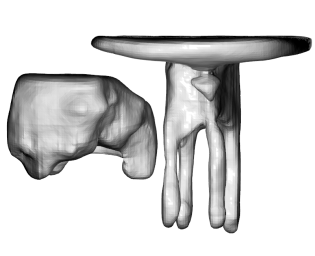}} \\
    \raisebox{-0.5\height}{\rotatebox{90}{GT}} &
    \raisebox{-0.5\height}{\includegraphics[height=0.075\linewidth]{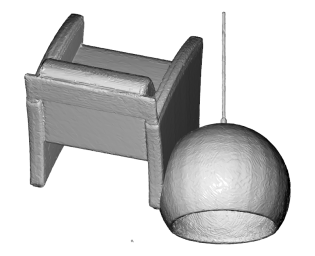}} &
    \raisebox{-0.5\height}{\includegraphics[height=0.075\linewidth]{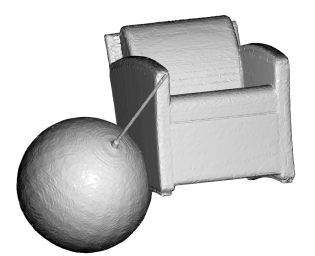}} &
    \raisebox{-0.5\height}{\includegraphics[height=0.075\linewidth]{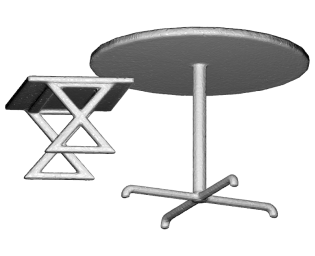}} &
    \raisebox{-0.5\height}{\includegraphics[height=0.075\linewidth]{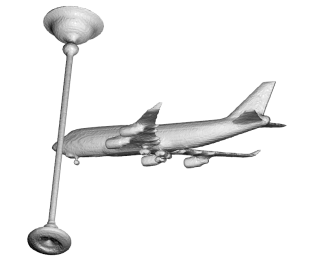}} &
    \raisebox{-0.5\height}{\includegraphics[height=0.075\linewidth]{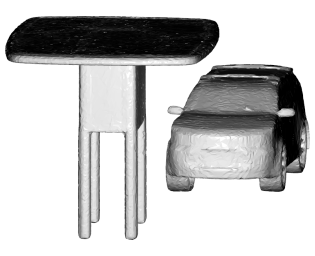}} &
    \raisebox{-0.5\height}{\includegraphics[height=0.075\linewidth]{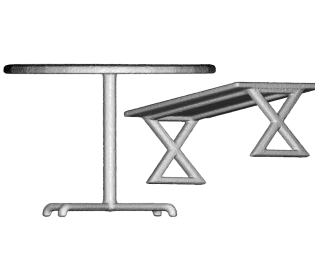}} &
    \raisebox{-0.5\height}{\includegraphics[height=0.075\linewidth]{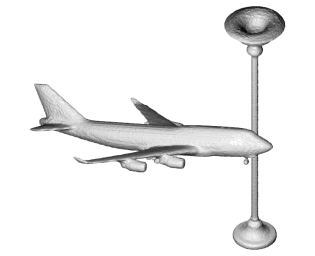}} &
    \raisebox{-0.5\height}{\includegraphics[height=0.075\linewidth]{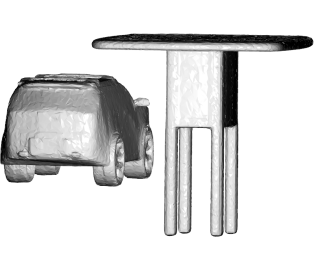}} \\
    \end{tabular}
\vspace{1.5em}
\caption{Qualitative results for networks trained on the \textit{lamp} class.}
\label{fig:apx_lamp}
\end{figure*}

%% file: appendix/chair/0_chair_figure.tex
\begin{figure*}[ht]
\renewcommand{\arraystretch}{3}
  \centering
    \begin{tabular}{m{0.5em} c c | c c | c c | c c}
    \raisebox{-0.5\height}{\rotatebox{90}{Input}} & \multicolumn{2}{c}{\raisebox{-0.5\height}{\includegraphics[height=0.075\linewidth]{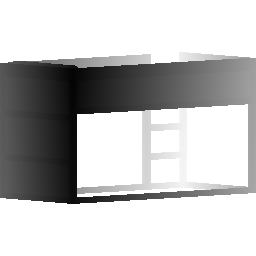}}} &
    \multicolumn{2}{c}{\raisebox{-0.5\height}{\includegraphics[height=0.075\linewidth]{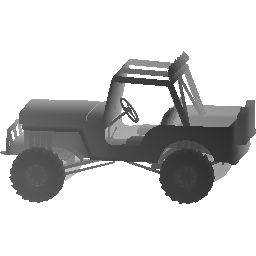}}} &
    \multicolumn{2}{c}{\raisebox{-0.5\height}{\includegraphics[height=0.075\linewidth]{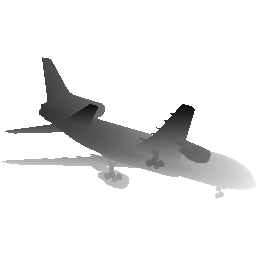}}} &
    \multicolumn{2}{c}{\raisebox{-0.5\height}{\includegraphics[height=0.075\linewidth]{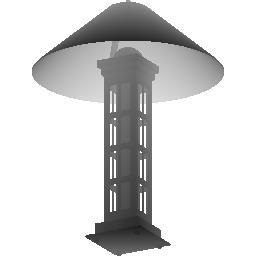}}} \\
    \raisebox{-0.5\height}{\rotatebox{90}{ONet}} &
    \raisebox{-0.5\height}{\includegraphics[height=0.075\linewidth]{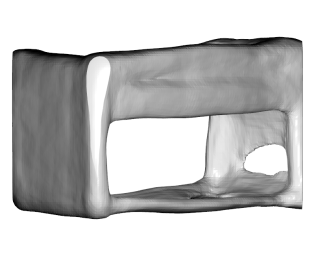}} &
    \raisebox{-0.5\height}{\includegraphics[height=0.075\linewidth]{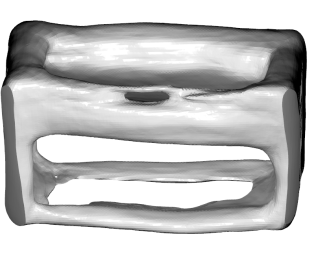}} &
    \raisebox{-0.5\height}{\includegraphics[height=0.075\linewidth]{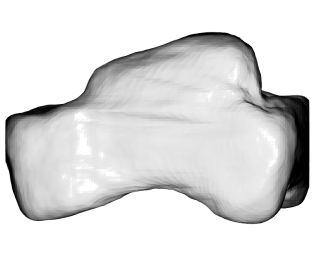}} &
    \raisebox{-0.5\height}{\includegraphics[height=0.075\linewidth]{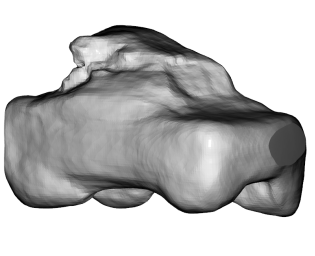}} &
    \raisebox{-0.5\height}{\includegraphics[height=0.075\linewidth]{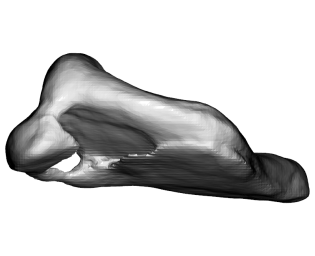}} &
    \raisebox{-0.5\height}{\includegraphics[height=0.075\linewidth]{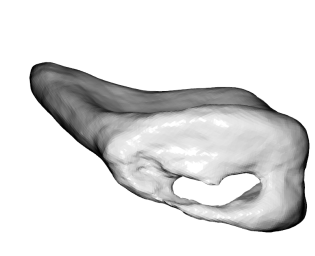}} &
    \raisebox{-0.5\height}{\includegraphics[height=0.075\linewidth]{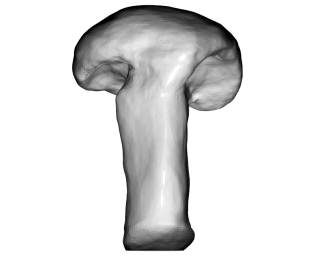}} &
    \raisebox{-0.5\height}{\includegraphics[height=0.075\linewidth]{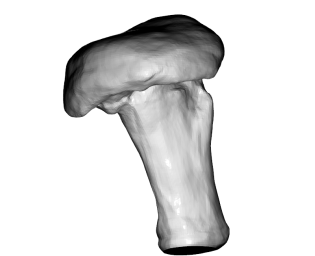}} \\
    \raisebox{-0.5\height}{\rotatebox{90}{HPN}} &
    \raisebox{-0.5\height}{\includegraphics[height=0.075\linewidth]{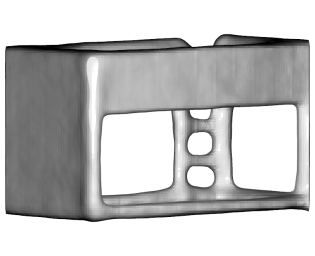}} &
    \raisebox{-0.5\height}{\includegraphics[height=0.075\linewidth]{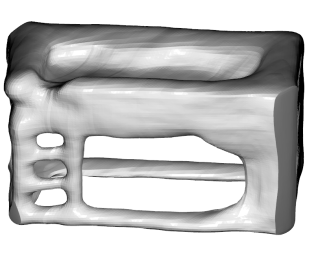}} &
    \raisebox{-0.5\height}{\includegraphics[height=0.075\linewidth]{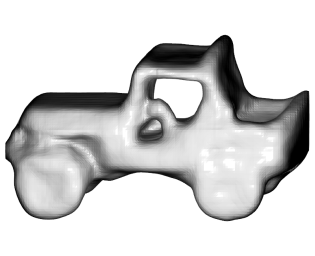}} &
    \raisebox{-0.5\height}{\includegraphics[height=0.075\linewidth]{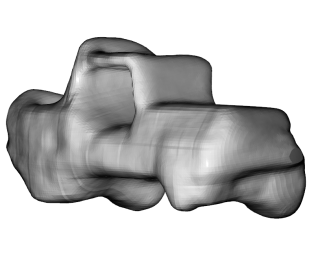}} &
    \raisebox{-0.5\height}{\includegraphics[height=0.075\linewidth]{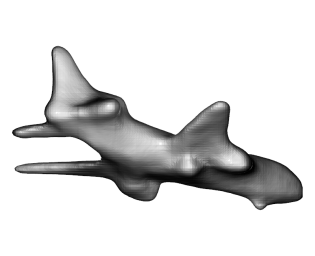}} &
    \raisebox{-0.5\height}{\includegraphics[height=0.075\linewidth]{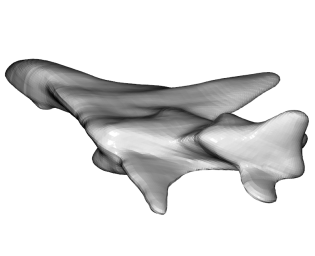}} &
    \raisebox{-0.5\height}{\includegraphics[height=0.075\linewidth]{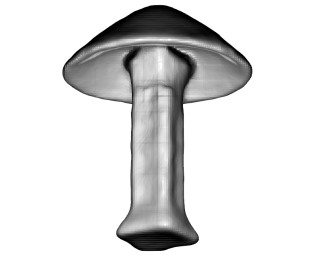}} &
    \raisebox{-0.5\height}{\includegraphics[height=0.075\linewidth]{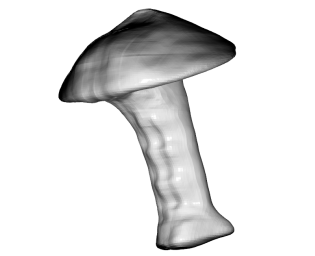}} \\
    \raisebox{-0.5\height}{\rotatebox{90}{GT}} &
    \raisebox{-0.5\height}{\includegraphics[height=0.075\linewidth]{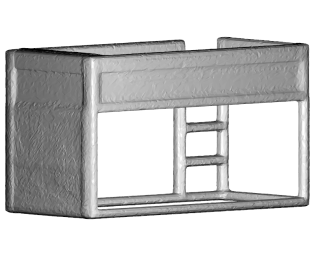}} &
    \raisebox{-0.5\height}{\includegraphics[height=0.075\linewidth]{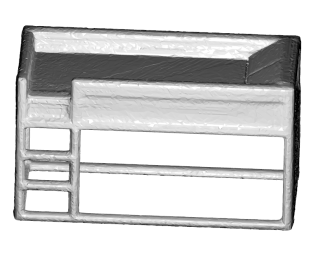}} &
    \raisebox{-0.5\height}{\includegraphics[height=0.075\linewidth]{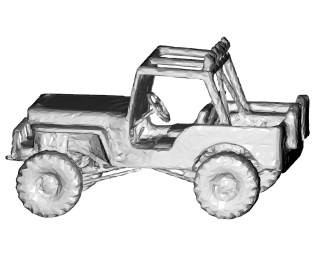}} &
    \raisebox{-0.5\height}{\includegraphics[height=0.075\linewidth]{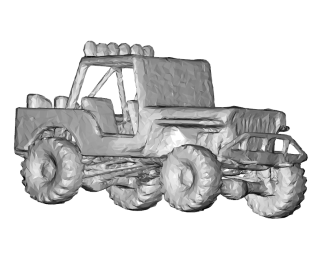}} &
    \raisebox{-0.5\height}{\includegraphics[height=0.075\linewidth]{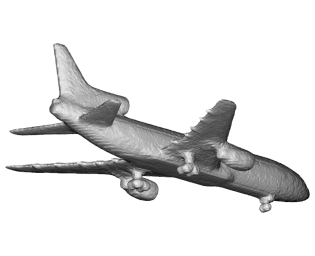}} &
    \raisebox{-0.5\height}{\includegraphics[height=0.075\linewidth]{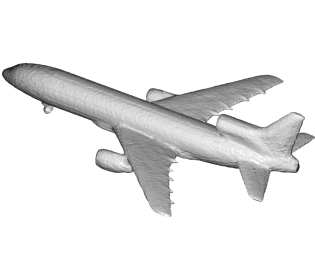}} &
    \raisebox{-0.5\height}{\includegraphics[height=0.075\linewidth]{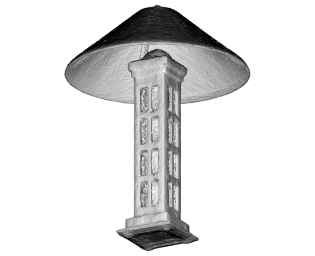}} &
    \raisebox{-0.5\height}{\includegraphics[height=0.075\linewidth]{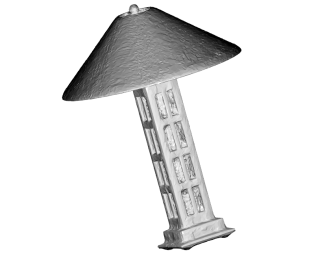}}\\
    \multicolumn{8}{c}{}\\ 
    \raisebox{-0.5\height}{\rotatebox{90}{Input}} & \multicolumn{2}{c}{\raisebox{-0.5\height}{\includegraphics[height=0.075\linewidth]{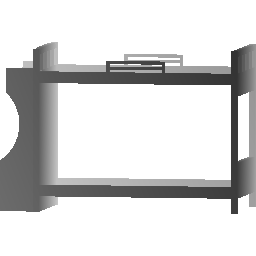}}} &
    \multicolumn{2}{c}{\raisebox{-0.5\height}{\includegraphics[height=0.075\linewidth]{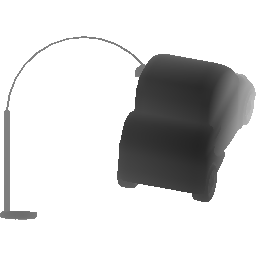}}} &
    \multicolumn{2}{c}{\raisebox{-0.5\height}{\includegraphics[height=0.075\linewidth]{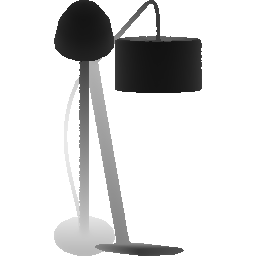}}} &
    \multicolumn{2}{c}{\raisebox{-0.5\height}{\includegraphics[height=0.075\linewidth]{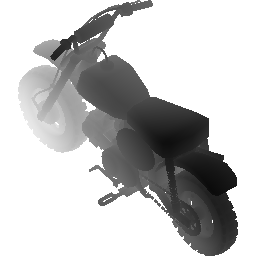}}} \\
    \raisebox{-0.5\height}{\rotatebox{90}{ONet}} &
    \raisebox{-0.5\height}{\includegraphics[height=0.075\linewidth]{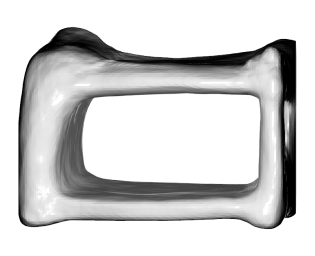}} &
    \raisebox{-0.5\height}{\includegraphics[height=0.075\linewidth]{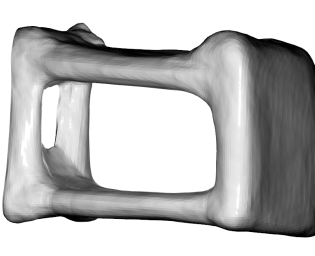}} &
    \raisebox{-0.5\height}{\includegraphics[height=0.075\linewidth]{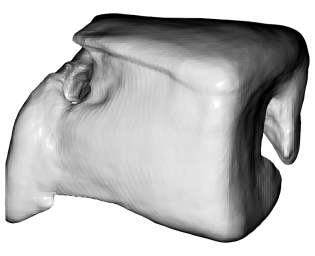}} &
    \raisebox{-0.5\height}{\includegraphics[height=0.075\linewidth]{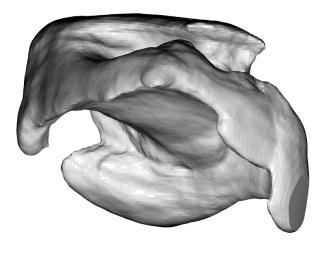}} &
    \raisebox{-0.5\height}{\includegraphics[height=0.075\linewidth]{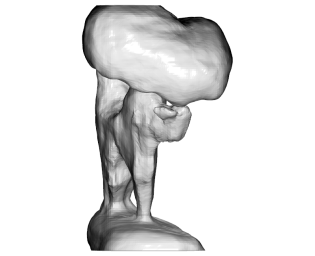}} &
    \raisebox{-0.5\height}{\includegraphics[height=0.075\linewidth]{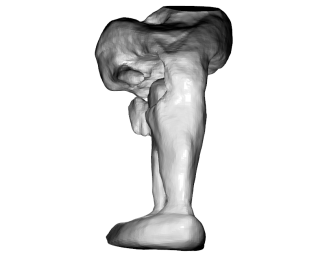}} &
    \raisebox{-0.5\height}{\includegraphics[height=0.075\linewidth]{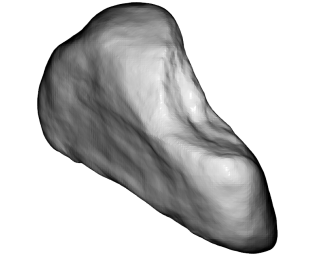}} &
    \raisebox{-0.5\height}{\includegraphics[height=0.075\linewidth]{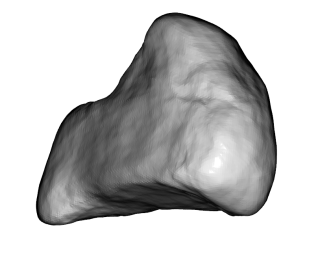}} \\
    \raisebox{-0.5\height}{\rotatebox{90}{HPN}} &
    \raisebox{-0.5\height}{\includegraphics[height=0.075\linewidth]{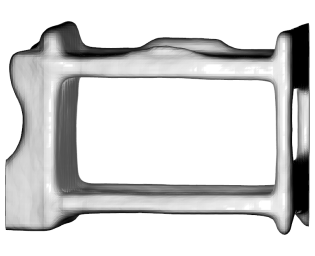}} &
    \raisebox{-0.5\height}{\includegraphics[height=0.075\linewidth]{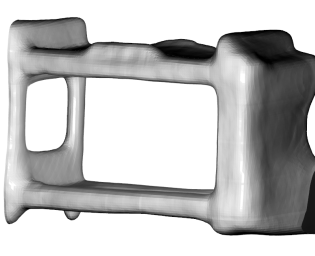}} &
    \raisebox{-0.5\height}{\includegraphics[height=0.075\linewidth]{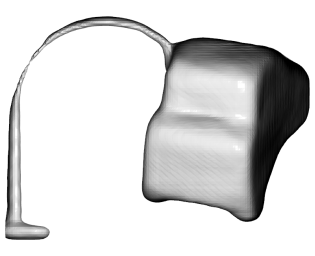}} &
    \raisebox{-0.5\height}{\includegraphics[height=0.075\linewidth]{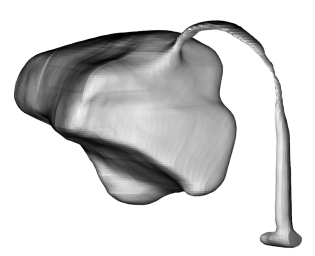}} &
    \raisebox{-0.5\height}{\includegraphics[height=0.075\linewidth]{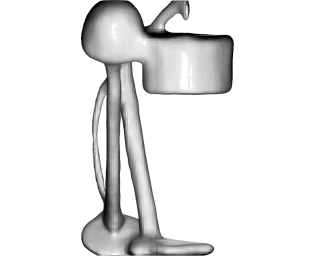}} &
    \raisebox{-0.5\height}{\includegraphics[height=0.075\linewidth]{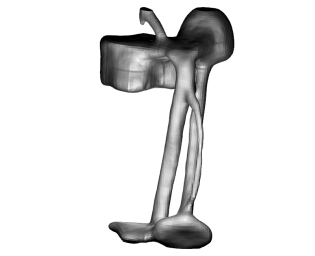}} &
    \raisebox{-0.5\height}{\includegraphics[height=0.075\linewidth]{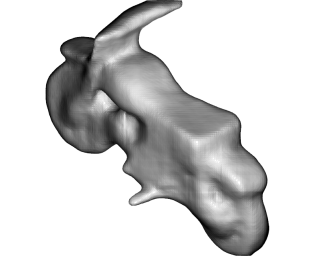}} &
    \raisebox{-0.5\height}{\includegraphics[height=0.075\linewidth]{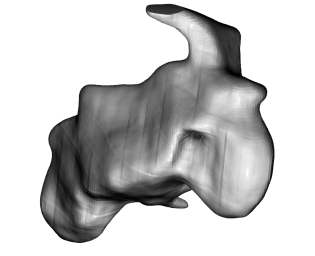}} \\
    \raisebox{-0.5\height}{\rotatebox{90}{GT}} &
    \raisebox{-0.5\height}{\includegraphics[height=0.075\linewidth]{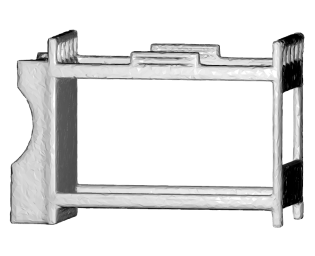}} &
    \raisebox{-0.5\height}{\includegraphics[height=0.075\linewidth]{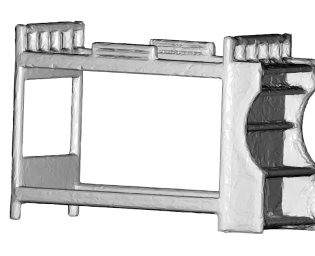}} &
    \raisebox{-0.5\height}{\includegraphics[height=0.075\linewidth]{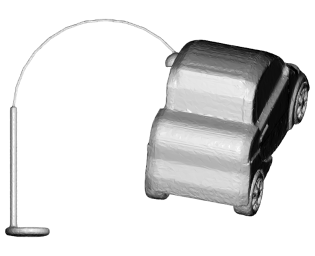}} &
    \raisebox{-0.5\height}{\includegraphics[height=0.075\linewidth]{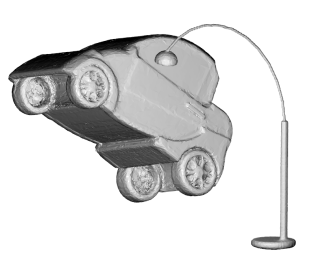}} &
    \raisebox{-0.5\height}{\includegraphics[height=0.075\linewidth]{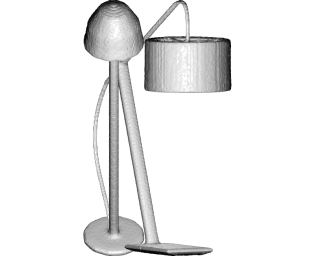}} &
    \raisebox{-0.5\height}{\includegraphics[height=0.075\linewidth]{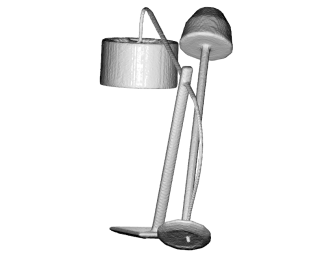}} &
    \raisebox{-0.5\height}{\includegraphics[height=0.075\linewidth]{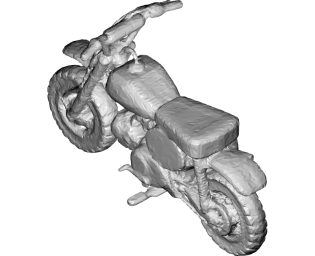}} &
    \raisebox{-0.5\height}{\includegraphics[height=0.075\linewidth]{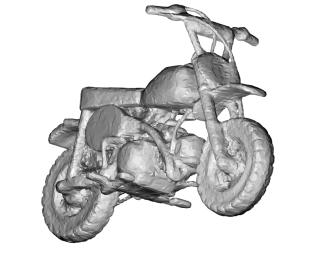}}\\
    \multicolumn{8}{c}{}\\ 
    \raisebox{-0.5\height}{\rotatebox{90}{Input}} &
    \multicolumn{2}{c}{\raisebox{-0.5\height}{\includegraphics[height=0.075\linewidth]{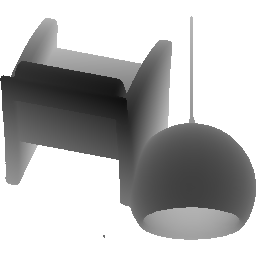}}} & 
    \multicolumn{2}{c}{\raisebox{-0.5\height}{\includegraphics[height=0.075\linewidth]{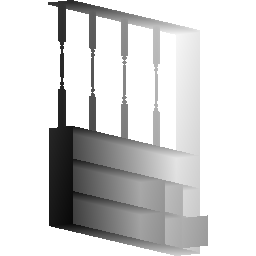}}} &
    \multicolumn{2}{c}{\raisebox{-0.5\height}{\includegraphics[height=0.075\linewidth]{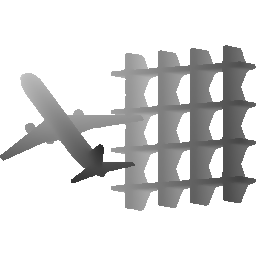}}} & 
    \multicolumn{2}{c}{\raisebox{-0.5\height}{\includegraphics[height=0.075\linewidth]{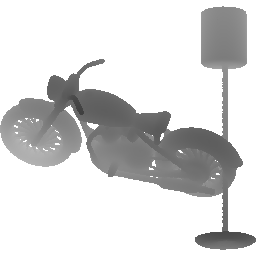}}} \\
    \raisebox{-0.5\height}{\rotatebox{90}{ONet}} &
    \raisebox{-0.5\height}{\includegraphics[height=0.075\linewidth]{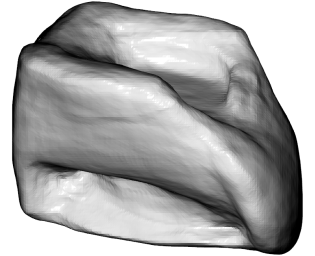}} &
    \raisebox{-0.5\height}{\includegraphics[height=0.075\linewidth]{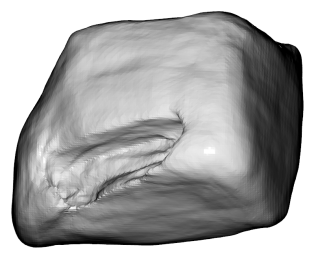}} &
    \raisebox{-0.5\height}{\includegraphics[height=0.075\linewidth]{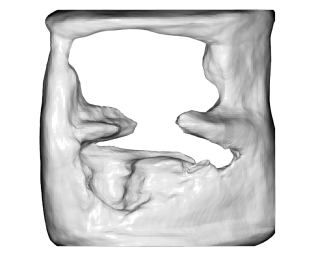}} &
    \raisebox{-0.5\height}{\includegraphics[height=0.075\linewidth]{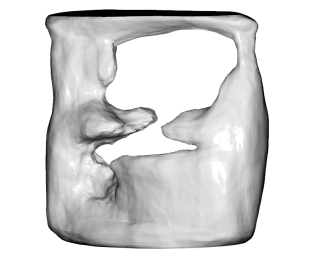}} &
    \raisebox{-0.5\height}{\includegraphics[height=0.075\linewidth]{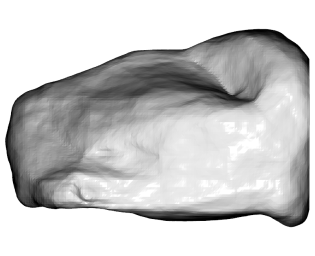}} &
    \raisebox{-0.5\height}{\includegraphics[height=0.075\linewidth]{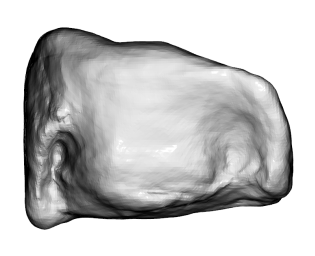}} &
    \raisebox{-0.5\height}{\includegraphics[height=0.075\linewidth]{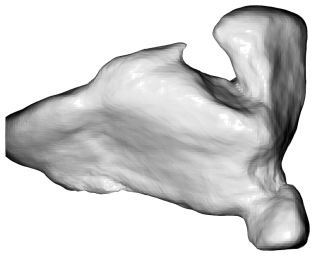}} &
    \raisebox{-0.5\height}{\includegraphics[height=0.075\linewidth]{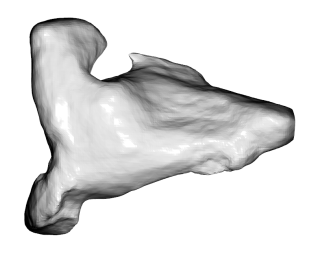}} \\
    \raisebox{-0.5\height}{\rotatebox{90}{HPN}} &
    \raisebox{-0.5\height}{\includegraphics[height=0.075\linewidth]{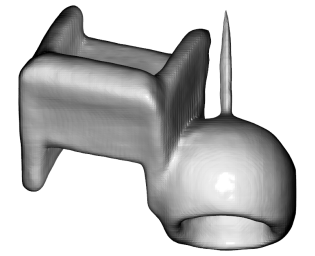}} &
    \raisebox{-0.5\height}{\includegraphics[height=0.075\linewidth]{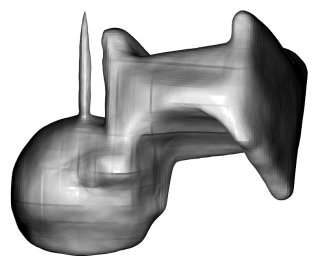}} &
    \raisebox{-0.5\height}{\includegraphics[height=0.075\linewidth]{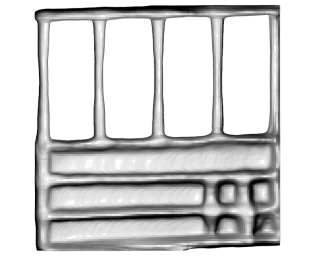}} &
    \raisebox{-0.5\height}{\includegraphics[height=0.075\linewidth]{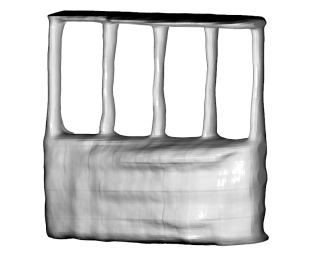}} &
    \raisebox{-0.5\height}{\includegraphics[height=0.075\linewidth]{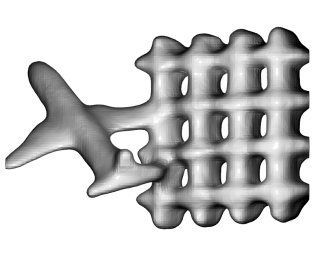}} &
    \raisebox{-0.5\height}{\includegraphics[height=0.075\linewidth]{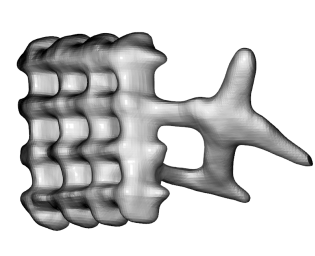}} &
    \raisebox{-0.5\height}{\includegraphics[height=0.075\linewidth]{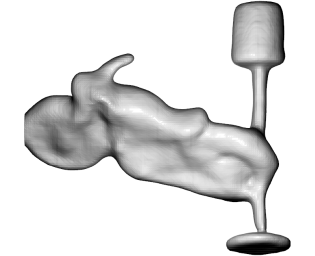}} &
    \raisebox{-0.5\height}{\includegraphics[height=0.075\linewidth]{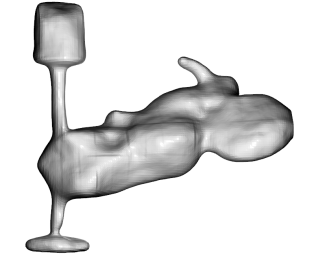}} \\
    \raisebox{-0.5\height}{\rotatebox{90}{GT}} &
    \raisebox{-0.5\height}{\includegraphics[height=0.075\linewidth]{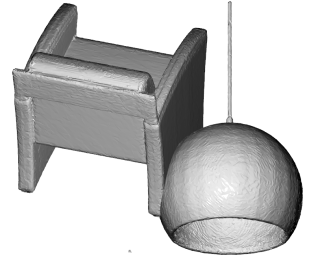}} &
    \raisebox{-0.5\height}{\includegraphics[height=0.075\linewidth]{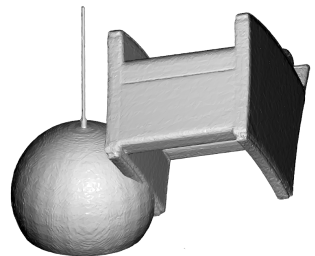}} &
    \raisebox{-0.5\height}{\includegraphics[height=0.075\linewidth]{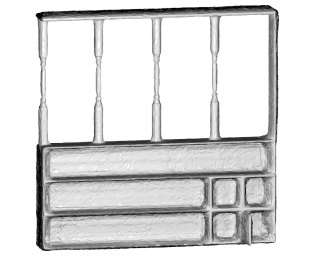}} &
    \raisebox{-0.5\height}{\includegraphics[height=0.075\linewidth]{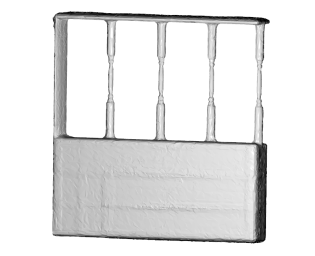}} &
    \raisebox{-0.5\height}{\includegraphics[height=0.075\linewidth]{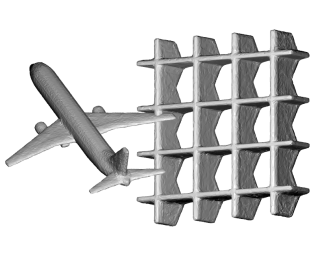}} &
    \raisebox{-0.5\height}{\includegraphics[height=0.075\linewidth]{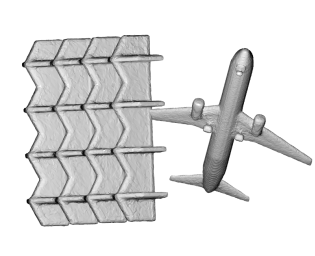}} &
    \raisebox{-0.5\height}{\includegraphics[height=0.075\linewidth]{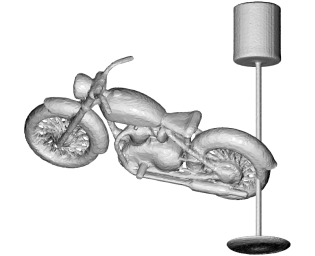}} &
    \raisebox{-0.5\height}{\includegraphics[height=0.075\linewidth]{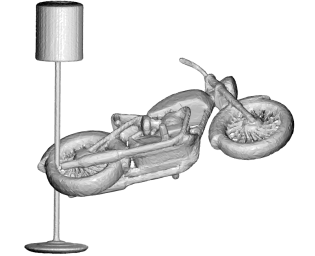}} \\
    \end{tabular}
\vspace{1.5em}
\caption{Qualitative results for networks trained on the \textit{chair} class.}
\label{fig:apx_chair}
\end{figure*}

%% file: appendix/multiclass/0_multiclass_figure.tex
\begin{figure*}[ht]
  \renewcommand{\arraystretch}{3}
  \centering
    \begin{tabular}{m{0.5em} c c | c c | c c | c c}
    \raisebox{-0.5\height}{\rotatebox{90}{Input}} & \multicolumn{2}{c}{\raisebox{-0.5\height}{\includegraphics[height=0.075\linewidth]{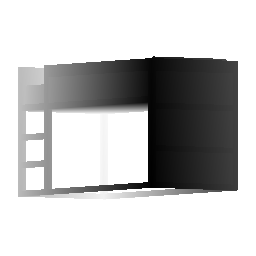}}} &
    \multicolumn{2}{c}{\raisebox{-0.5\height}{\includegraphics[height=0.075\linewidth]{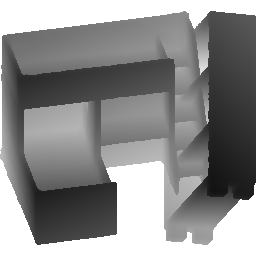}}} &
    \multicolumn{2}{c}{\raisebox{-0.5\height}{\includegraphics[height=0.075\linewidth]{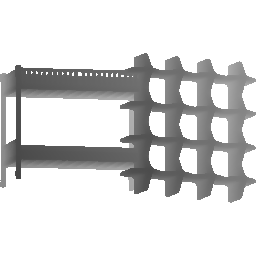}}} &
    \multicolumn{2}{c}{\raisebox{-0.5\height}{\includegraphics[height=0.075\linewidth]{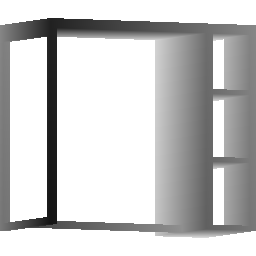}}} \\
    \raisebox{-0.5\height}{\rotatebox{90}{ONet}} &
    \raisebox{-0.5\height}{\includegraphics[height=0.075\linewidth]{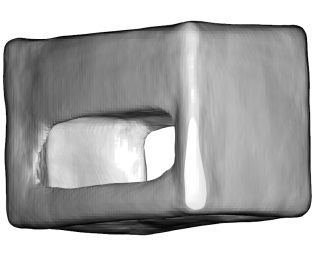}} &
    \raisebox{-0.5\height}{\includegraphics[height=0.075\linewidth]{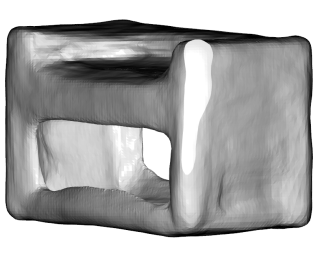}} &
    \raisebox{-0.5\height}{\includegraphics[height=0.075\linewidth]{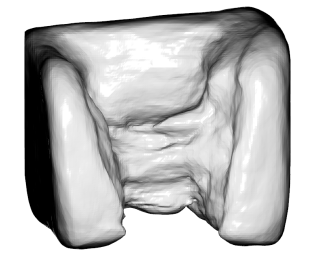}} &
    \raisebox{-0.5\height}{\includegraphics[height=0.075\linewidth]{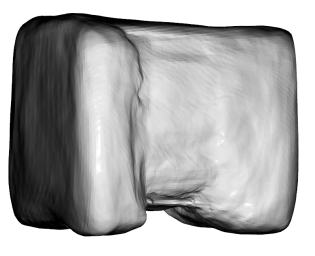}} &
    \raisebox{-0.5\height}{\includegraphics[height=0.075\linewidth]{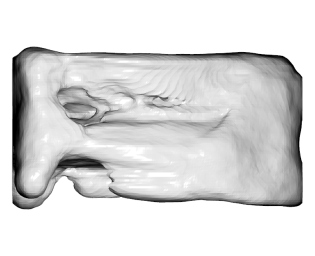}} &
    \raisebox{-0.5\height}{\includegraphics[height=0.075\linewidth]{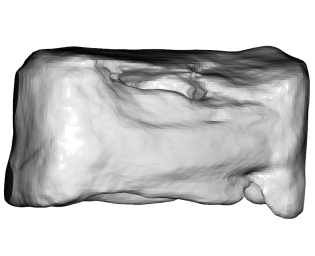}} &
    \raisebox{-0.5\height}{\includegraphics[height=0.075\linewidth]{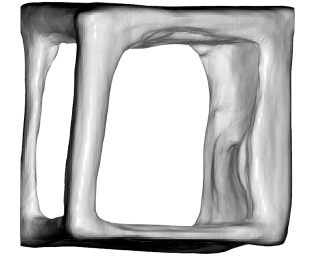}} &
    \raisebox{-0.5\height}{\includegraphics[height=0.075\linewidth]{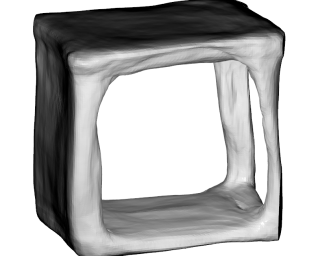}} \\
    \raisebox{-0.5\height}{\rotatebox{90}{HPN-SDF}} &
    \raisebox{-0.5\height}{\includegraphics[height=0.075\linewidth]{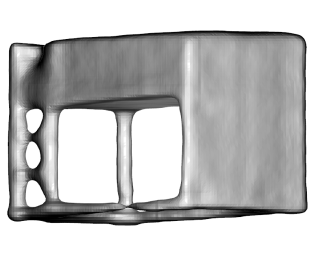}} &
    \raisebox{-0.5\height}{\includegraphics[height=0.075\linewidth]{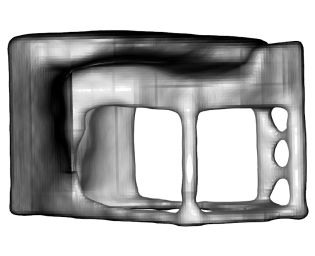}} &
    \raisebox{-0.5\height}{\includegraphics[height=0.075\linewidth]{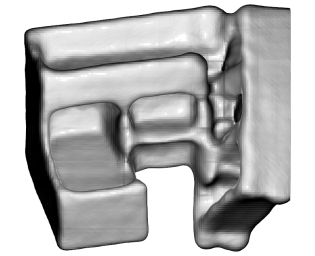}} &
    \raisebox{-0.5\height}{\includegraphics[height=0.075\linewidth]{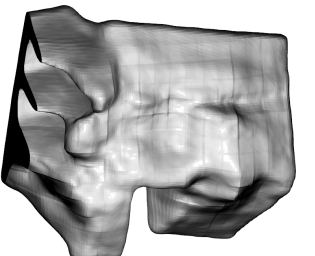}} &
    \raisebox{-0.5\height}{\includegraphics[height=0.075\linewidth]{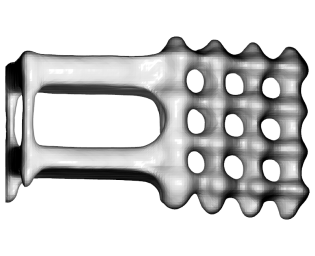}} &
    \raisebox{-0.5\height}{\includegraphics[height=0.075\linewidth]{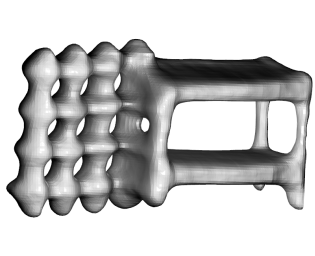}} &
    \raisebox{-0.5\height}{\includegraphics[height=0.075\linewidth]{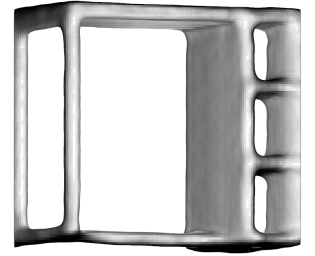}} &
    \raisebox{-0.5\height}{\includegraphics[height=0.075\linewidth]{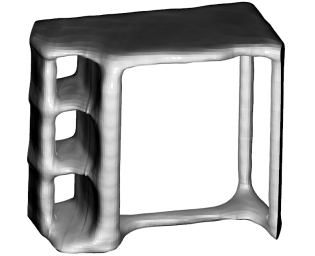}} \\
    \raisebox{-0.5\height}{\rotatebox{90}{GT}} &
    \raisebox{-0.5\height}{\includegraphics[height=0.075\linewidth]{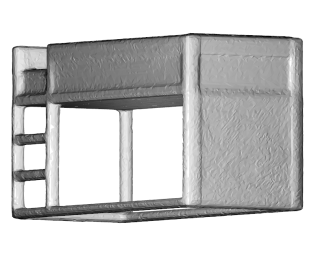}} &
    \raisebox{-0.5\height}{\includegraphics[height=0.075\linewidth]{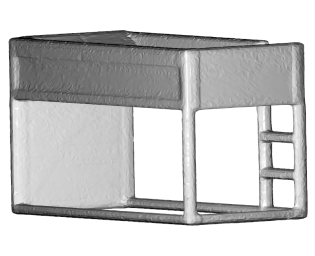}} &
    \raisebox{-0.5\height}{\includegraphics[height=0.075\linewidth]{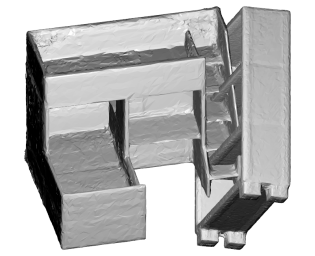}} &
    \raisebox{-0.5\height}{\includegraphics[height=0.075\linewidth]{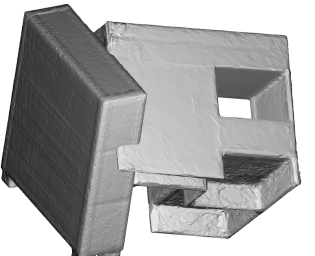}} &
    \raisebox{-0.5\height}{\includegraphics[height=0.075\linewidth]{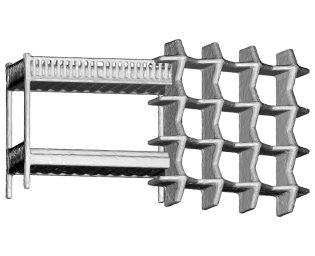}} &
    \raisebox{-0.5\height}{\includegraphics[height=0.075\linewidth]{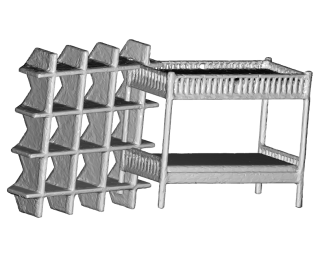}} &
    \raisebox{-0.5\height}{\includegraphics[height=0.075\linewidth]{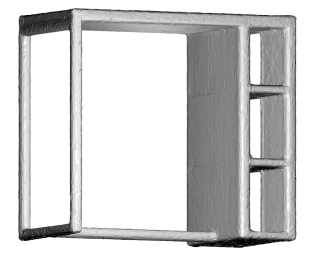}} &
    \raisebox{-0.5\height}{\includegraphics[height=0.075\linewidth]{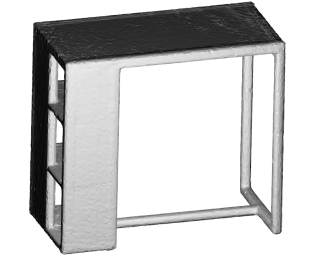}} \\
    \multicolumn{8}{c}{}\\ 
    \raisebox{-0.5\height}{\rotatebox{90}{Input}} & \multicolumn{2}{c}{\raisebox{-0.5\height}{\includegraphics[height=0.075\linewidth]{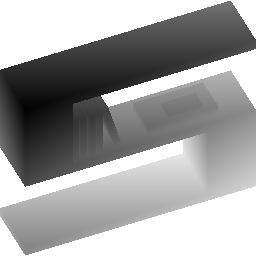}}} &
    \multicolumn{2}{c}{\raisebox{-0.5\height}{\includegraphics[height=0.075\linewidth]{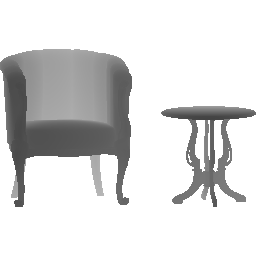}}} &
    \multicolumn{2}{c}{\raisebox{-0.5\height}{\includegraphics[height=0.075\linewidth]{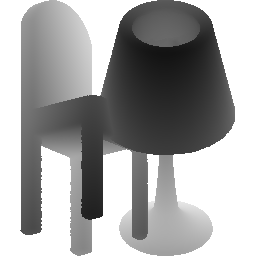}}} &
    \multicolumn{2}{c}{\raisebox{-0.5\height}{\includegraphics[height=0.075\linewidth]{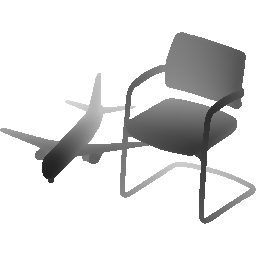}}} \\
    \raisebox{-0.5\height}{\rotatebox{90}{ONet}} &
    \raisebox{-0.5\height}{\includegraphics[height=0.075\linewidth]{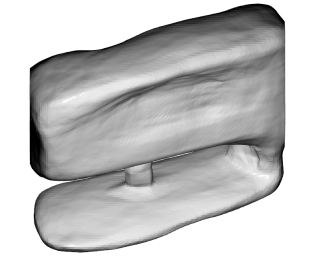}} &
    \raisebox{-0.5\height}{\includegraphics[height=0.075\linewidth]{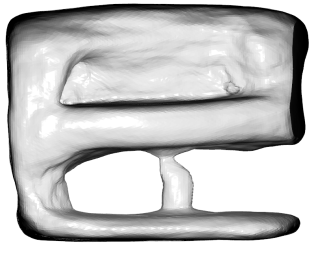}} &
    \raisebox{-0.5\height}{\includegraphics[height=0.075\linewidth]{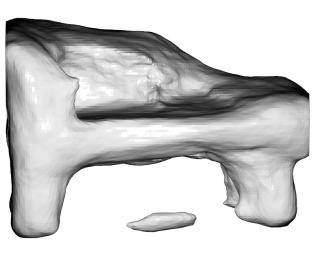}} &
    \raisebox{-0.5\height}{\includegraphics[height=0.075\linewidth]{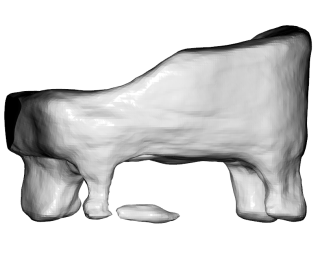}} &
    \raisebox{-0.5\height}{\includegraphics[height=0.075\linewidth]{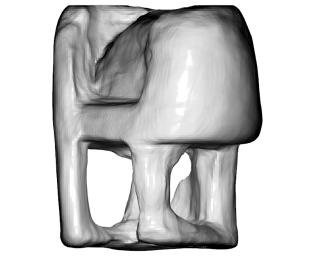}} &
    \raisebox{-0.5\height}{\includegraphics[height=0.075\linewidth]{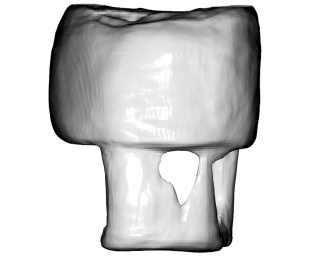}} &
    \raisebox{-0.5\height}{\includegraphics[height=0.075\linewidth]{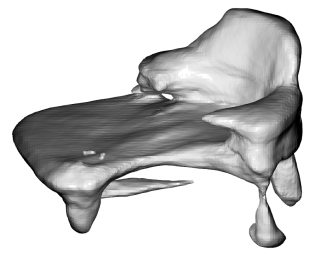}} &
    \raisebox{-0.5\height}{\includegraphics[height=0.075\linewidth]{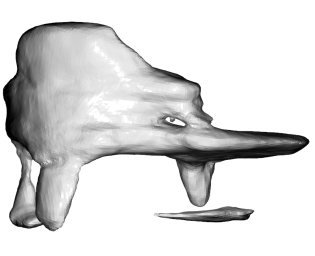}} \\
    \raisebox{-0.5\height}{\rotatebox{90}{HPN-SDF}} &
    \raisebox{-0.5\height}{\includegraphics[height=0.075\linewidth]{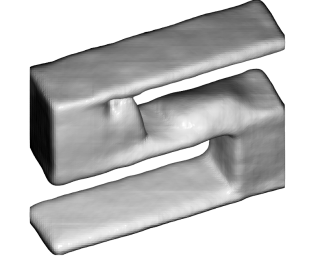}} &
    \raisebox{-0.5\height}{\includegraphics[height=0.075\linewidth]{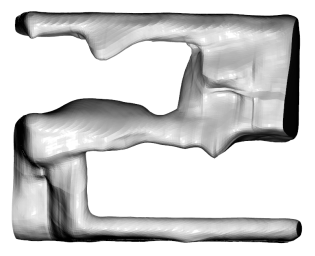}} &
    \raisebox{-0.5\height}{\includegraphics[height=0.075\linewidth]{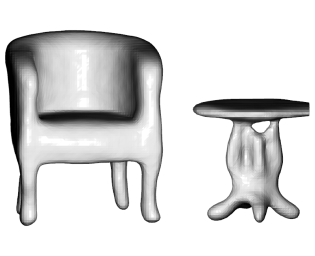}} &
    \raisebox{-0.5\height}{\includegraphics[height=0.075\linewidth]{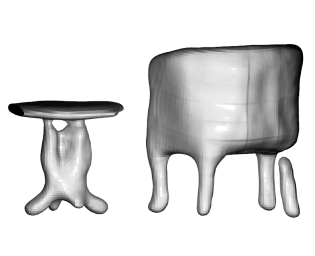}} &
    \raisebox{-0.5\height}{\includegraphics[height=0.075\linewidth]{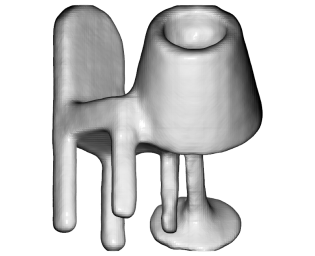}} &
    \raisebox{-0.5\height}{\includegraphics[height=0.075\linewidth]{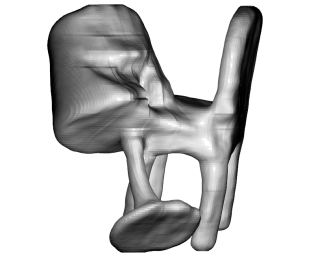}} &
    \raisebox{-0.5\height}{\includegraphics[height=0.075\linewidth]{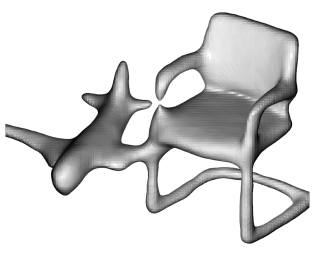}} &
    \raisebox{-0.5\height}{\includegraphics[height=0.075\linewidth]{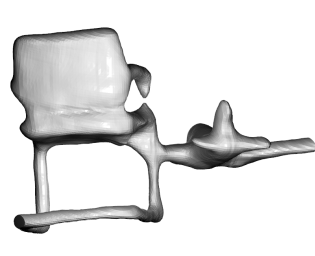}} \\
    \raisebox{-0.5\height}{\rotatebox{90}{GT}} &
    \raisebox{-0.5\height}{\includegraphics[height=0.075\linewidth]{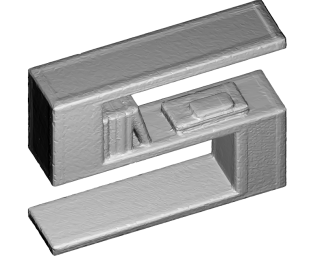}} &
    \raisebox{-0.5\height}{\includegraphics[height=0.075\linewidth]{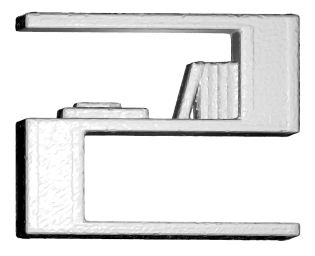}} &
    \raisebox{-0.5\height}{\includegraphics[height=0.075\linewidth]{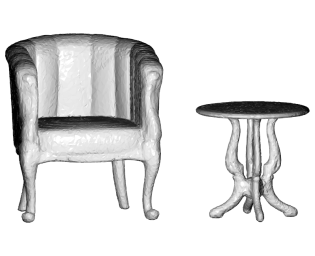}} &
    \raisebox{-0.5\height}{\includegraphics[height=0.075\linewidth]{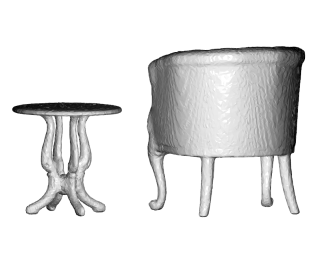}} &
    \raisebox{-0.5\height}{\includegraphics[height=0.075\linewidth]{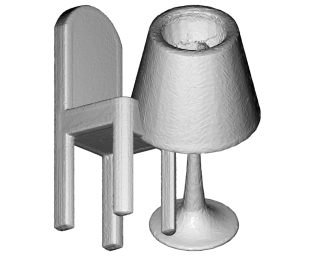}} &
    \raisebox{-0.5\height}{\includegraphics[height=0.075\linewidth]{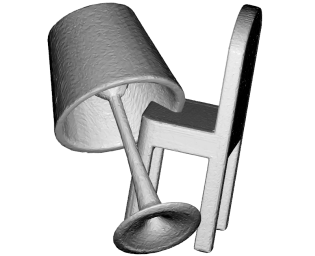}} &
    \raisebox{-0.5\height}{\includegraphics[height=0.075\linewidth]{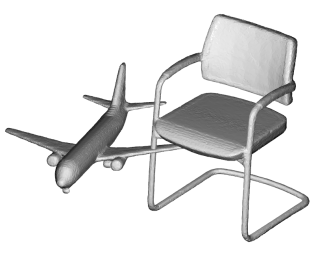}} &
    \raisebox{-0.5\height}{\includegraphics[height=0.075\linewidth]{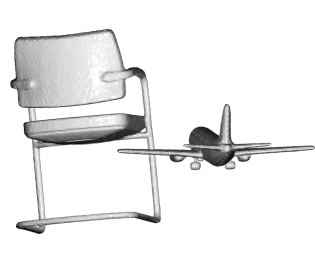}}\\
    \multicolumn{8}{c}{}\\ 
    \raisebox{-0.5\height}{\rotatebox{90}{Input}} &
    \multicolumn{2}{c}{\raisebox{-0.5\height}{\includegraphics[height=0.075\linewidth]{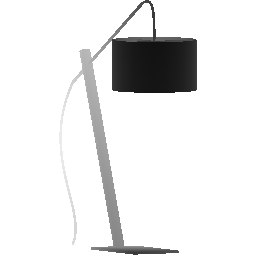}}} & 
    \multicolumn{2}{c}{\raisebox{-0.5\height}{\includegraphics[height=0.075\linewidth]{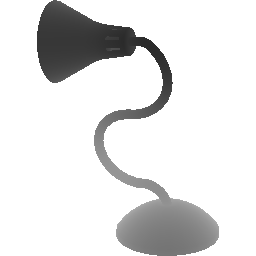}}} &
    \multicolumn{2}{c}{\raisebox{-0.5\height}{\includegraphics[height=0.075\linewidth]{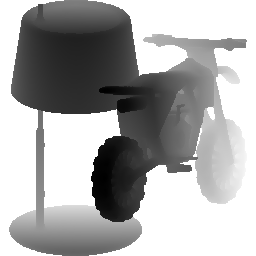}}} & 
    \multicolumn{2}{c}{\raisebox{-0.5\height}{\includegraphics[height=0.075\linewidth]{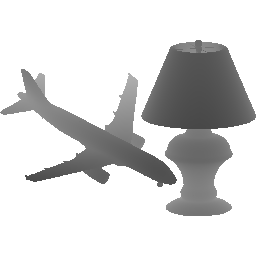}}} \\
    \raisebox{-0.5\height}{\rotatebox{90}{ONet}} &
    \raisebox{-0.5\height}{\includegraphics[height=0.075\linewidth]{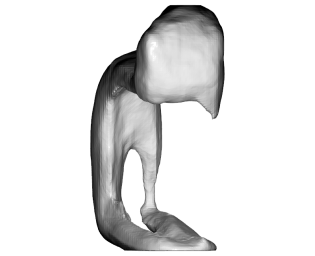}} &
    \raisebox{-0.5\height}{\includegraphics[height=0.075\linewidth]{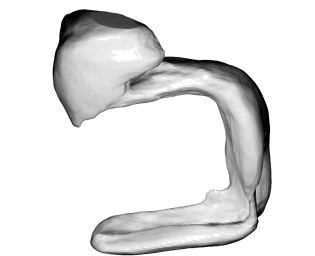}} &
    \raisebox{-0.5\height}{\includegraphics[height=0.075\linewidth]{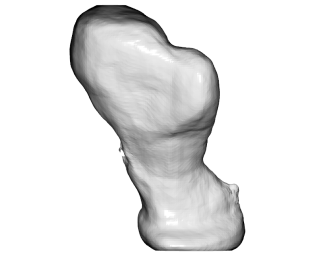}} &
    \raisebox{-0.5\height}{\includegraphics[height=0.075\linewidth]{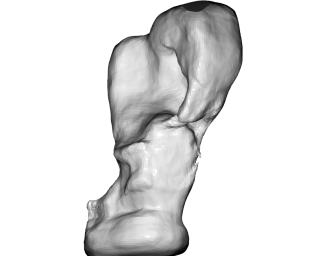}} &
    \raisebox{-0.5\height}{\includegraphics[height=0.075\linewidth]{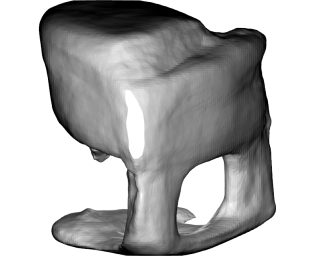}} &
    \raisebox{-0.5\height}{\includegraphics[height=0.075\linewidth]{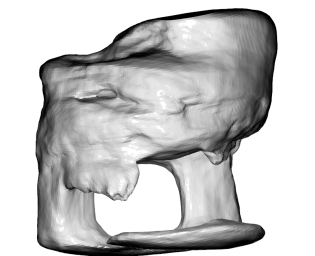}} &
    \raisebox{-0.5\height}{\includegraphics[height=0.075\linewidth]{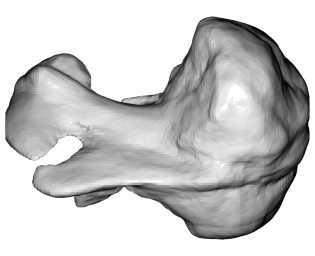}} &
    \raisebox{-0.5\height}{\includegraphics[height=0.075\linewidth]{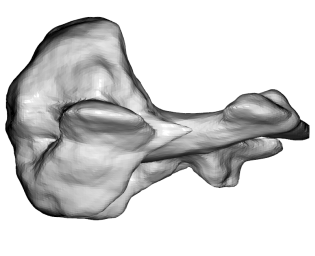}} \\
    \raisebox{-0.5\height}{\rotatebox{90}{HPN-SDF}} &
    \raisebox{-0.5\height}{\includegraphics[height=0.075\linewidth]{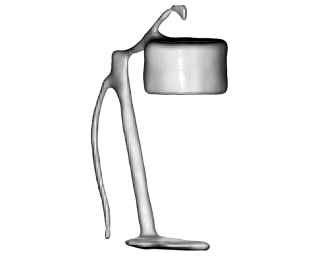}} &
    \raisebox{-0.5\height}{\includegraphics[height=0.075\linewidth]{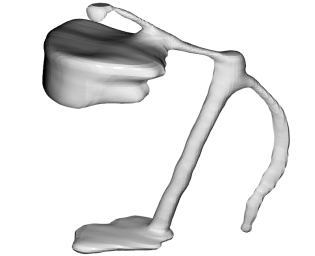}} &
    \raisebox{-0.5\height}{\includegraphics[height=0.075\linewidth]{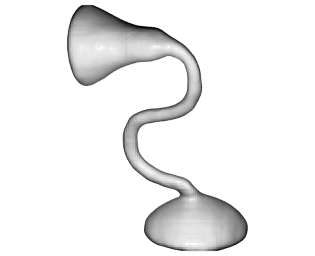}} &
    \raisebox{-0.5\height}{\includegraphics[height=0.075\linewidth]{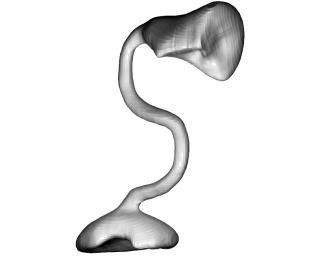}} &
    \raisebox{-0.5\height}{\includegraphics[height=0.075\linewidth]{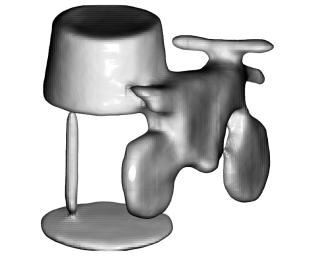}} &
    \raisebox{-0.5\height}{\includegraphics[height=0.075\linewidth]{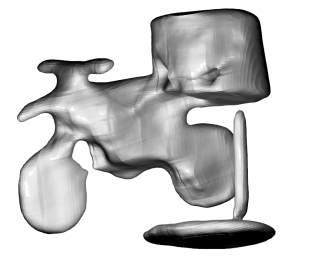}} &
    \raisebox{-0.5\height}{\includegraphics[height=0.075\linewidth]{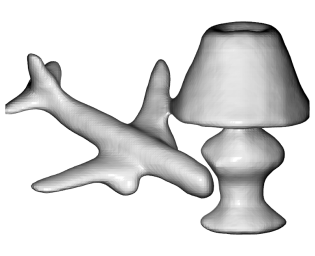}} &
    \raisebox{-0.5\height}{\includegraphics[height=0.075\linewidth]{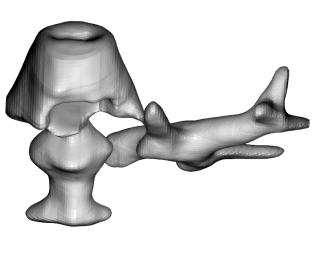}} \\
    \raisebox{-0.5\height}{\rotatebox{90}{GT}} &
    \raisebox{-0.5\height}{\includegraphics[height=0.075\linewidth]{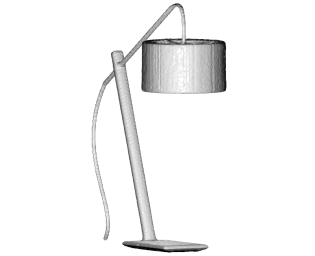}} &
    \raisebox{-0.5\height}{\includegraphics[height=0.075\linewidth]{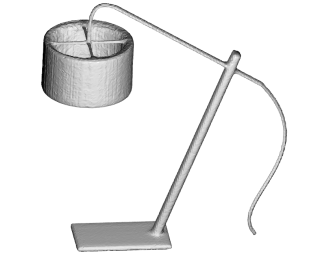}} &
    \raisebox{-0.5\height}{\includegraphics[height=0.075\linewidth]{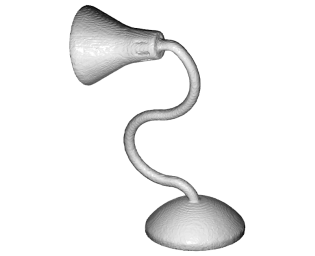}} &
    \raisebox{-0.5\height}{\includegraphics[height=0.075\linewidth]{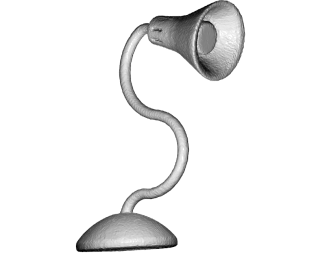}} &
    \raisebox{-0.5\height}{\includegraphics[height=0.075\linewidth]{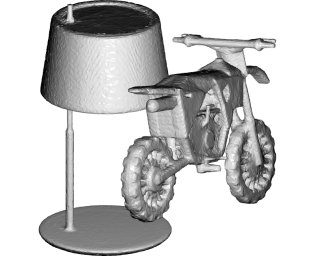}} &
    \raisebox{-0.5\height}{\includegraphics[height=0.075\linewidth]{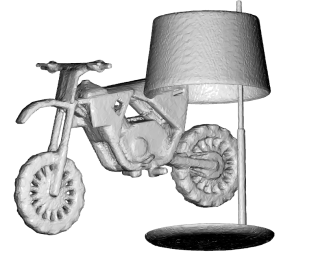}} &
    \raisebox{-0.5\height}{\includegraphics[height=0.075\linewidth]{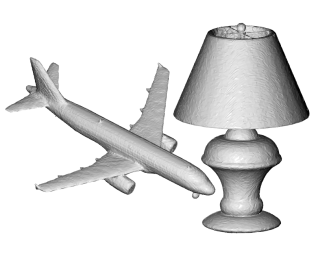}} &
    \raisebox{-0.5\height}{\includegraphics[height=0.075\linewidth]{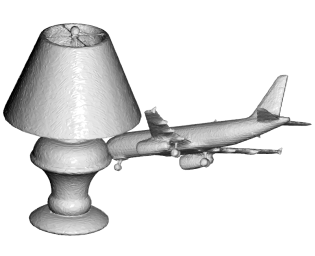}}
    \end{tabular}
\vspace{1.5em}
\caption{Qualitative results for networks trained in the \textit{multi-class} setting on \textit{airplanes}, \textit{cars} and \textit{chairs}.}
\label{fig:apx_multiclass}
\end{figure*}

%% file: X_quantitative_genre13.tex
\begin{landscape}

\begin{table}[htb!]
\centering
\resizebox*{\hsize}{!}{
	\setlength{\tabcolsep}{2mm}
	\begin{tabular}{l c
	c >{\columncolor{lightgray}}c || 
	c >{\columncolor{lightgray}}c 
	c >{\columncolor{lightgray}}c 
	c >{\columncolor{lightgray}}c 
	c >{\columncolor{lightgray}}c 
	c >{\columncolor{lightgray}}c 
	c >{\columncolor{lightgray}}c 
	c >{\columncolor{lightgray}}c 
	c >{\columncolor{lightgray}}c 
	c >{\columncolor{lightgray}}c 
	c >{\columncolor{lightgray}}c 
	c >{\columncolor{lightgray}}c 
	c >{\columncolor{lightgray}}c 
	c >{\columncolor{lightgray}}c 
	|| c >{\columncolor{lightgray}}c 
	|| c >{\columncolor{lightgray}}c} 
	\toprule
	& &	
	\multicolumn{2}{c||}{Mean (seen)} &
	\multicolumn{2}{c}{Airplane} &
	\multicolumn{2}{c}{Car} &
	\multicolumn{2}{c}{Chair} &
	\multicolumn{2}{c}{Lamp} &
	\multicolumn{2}{c}{Bench} &
	\multicolumn{2}{c}{Cabinet} &
	\multicolumn{2}{c}{Display} &
	\multicolumn{2}{c}{Speaker} &
	\multicolumn{2}{c}{Rifle} &
	\multicolumn{2}{c}{Phone} &
	\multicolumn{2}{c}{Vessel} &
	\multicolumn{2}{c}{Sofa} &
	\multicolumn{2}{c||}{Table} &
	\multicolumn{2}{c||}{Mean (unseen)} & \multicolumn{2}{c}{Composition}\\
	& &
	F$\uparrow$ & CD$\downarrow$ & 
	F$\uparrow$ & CD$\downarrow$ & 
	F$\uparrow$ & CD$\downarrow$ & 
	F$\uparrow$ & CD$\downarrow$ & 
	F$\uparrow$ & CD$\downarrow$ & 
	F$\uparrow$ & CD$\downarrow$ & 
	F$\uparrow$ & CD$\downarrow$ & 
	F$\uparrow$ & CD$\downarrow$ & 
	F$\uparrow$ & CD$\downarrow$ & 
	F$\uparrow$ & CD$\downarrow$ & 
	F$\uparrow$ & CD$\downarrow$ & 
	F$\uparrow$ & CD$\downarrow$ & 
	F$\uparrow$ & CD$\downarrow$ & 
	F$\uparrow$ & CD$\downarrow$ & 
	F$\uparrow$ & CD$\downarrow$ & 
	F$\uparrow$ & CD$\downarrow$\\ 
	\midrule
	\multirow{6}{*}{\rotatebox{90}{\textit{plane,car,chair}}} &
	ONet~\cite{mescheder2019occupancy} &
	\textcolor{blue_our}{44.4} & \textcolor{blue_our}{3.8} &
	\textcolor{blue_our}{34.7} & \textcolor{blue_our}{4.1} &
	\textcolor{blue_our}{57.7} & \textcolor{blue_our}{\textbf{3.2}} &
	\textcolor{blue_our}{40.8} & \textcolor{blue_our}{4.1} &
	18.8 & 9.3 &
	31.7 & 5.2 & 
	46.9 & 4.8 & 
	19.5 & 9.0 & 
	38.4 & 6.0 & 
	13.3 & 9.0 & 
	19.8 & 8.0 & 
	26.5 & 6.7 & 
	43.2 & 4.7 &
	35.2 & 5.3 &
	\textcolor{orange_our}{29.3} & \textcolor{orange_our}{6.8} &
	\textcolor{green_our}{18.3} & \textcolor{green_our}{8.7}\\
	& ONet-SDF~\cite{mescheder2019occupancy} &
	\textcolor{blue_our}{37.2} & \textcolor{blue_our}{4.5} &
	\textcolor{blue_our}{29.0} & \textcolor{blue_our}{4.9} &
	\textcolor{blue_our}{46.7} & \textcolor{blue_our}{3.9} &
	\textcolor{blue_our}{35.9} & \textcolor{blue_our}{4.6} &
	19.9 & 8.5 &
	28.4 & 5.6 &
	41.7 & 5.2 &
	23.0 & 8.0 &
	37.6 & 5.8 &
	14.7 & 8.5 &
	23.2 & 7.2 &
	26.7 & 6.5 &
	38.3 & 5.1 &
	33.0 & 5.6 &
	\textcolor{orange_our}{28.6} & \textcolor{orange_our}{6.6} &
	\textcolor{green_our}{19.3} & \textcolor{green_our}{8.0}\\
	& GenRe~\cite{zhang2018learninggenre} &
	\textcolor{blue_our}{-} & \textcolor{blue_our}{4.5*} &
	\textcolor{blue_our}{-} & \textcolor{blue_our}{-} &
	\textcolor{blue_our}{-} & \textcolor{blue_our}{-} &
	\textcolor{blue_our}{-} & \textcolor{blue_our}{-} &
	- & 6.0* &
	- & 5.0* &
	- & 7.6* &
	- & 6.0* &
	- & 7.7* &
	- & 3.1* &
	- & 5.4* &
	- & 4.8* &
	- & 5.9* &
	- & 5.7* &
	\textcolor{orange_our}{-} & \textcolor{orange_our}{5.7}* &
	\textcolor{green_our}{-} & \textcolor{green_our}{-}\\
	& LDIF$_{svim1d}$~\cite{genova2020local} & 
	\textcolor{blue_our}{\textbf{76.7}} & \textcolor{blue_our}{\textbf{0.5}} &
	\textcolor{blue_our}{\textbf{87.9}} & \textcolor{blue_our}{\textbf{0.2}} &
	\textcolor{blue_our}{\textbf{80.0}} & \textcolor{blue_our}{\textbf{0.3}} &
	\textcolor{blue_our}{\textbf{62.1}} & \textcolor{blue_our}{\textbf{0.9}} &
	20.8 & 9.4 & 
	\textbf{48.6} & \textbf{1.2} &
	26.2 & \textbf{3.4} &
	15.8 & \textbf{5.7} &
	22.9 & 5.2 &
	32.2 & \textbf{1.5} &
	20.6 & \textbf{2.3} &
	\textbf{48.1} & \textbf{1.4} &
	52.7 & \textbf{1.3} &
	33.0 & \textbf{3.3} &
	\textcolor{orange_our}{32.1} & \textcolor{orange_our}{\textbf{3.5}} &
	\textcolor{green_our}{16.4} & \textcolor{green_our}{10.9}\\
	\cmidrule{2-34} 
	& HPN (ours) &
	\textcolor{blue_our}{45.4} & \textcolor{blue_our}{3.8} &
	\textcolor{blue_our}{36.8} & \textcolor{blue_our}{3.9} &
	\textcolor{blue_our}{55.1} & \textcolor{blue_our}{3.5} &
	\textcolor{blue_our}{44.3} & \textcolor{blue_our}{3.8} &
	38.4 & 4.8 &
	37.4 & 4.3 &
	\textbf{54.0} & \textbf{4.3} &
	43.2 & 6.3 &
	\textbf{49.7} & \textbf{4.8} &
	33.1 & 5.0 &
	45.9 & 5.3 &
	37.3 & 5.4 &
	46.6 & 4.5 &
	43.7 & 4.4 &
	\textcolor{orange_our}{42.9} & \textcolor{orange_our}{4.9} &
	\textcolor{green_our}{30.2} & \textcolor{green_our}{5.7}\\
	& HPN-SDF (ours) &
	\textcolor{blue_our}{54.9} & \textcolor{blue_our}{3.2} &
	\textcolor{blue_our}{52.8} & \textcolor{blue_our}{2.9} &
	\textcolor{blue_our}{58.3} & \textcolor{blue_our}{3.5} &
	\textcolor{blue_our}{53.6} & \textcolor{blue_our}{3.3} &
	\textbf{56.5} & \textbf{3.5} &
	36.9 & 4.4 &
	50.7 & 5.0 &
	\textbf{47.1} & 6.0 &
	49.4 & 5.0 &
	\textbf{39.9} & 4.3 &
	\textbf{53.8} & 4.9 &
	40.1 & 5.7 &
	\textbf{54.4} & 3.9 &
	\textbf{53.1} & 3.7 &
	\textcolor{orange_our}{\textbf{48.2}} &
	\textcolor{orange_our}{4.6} &
	\textcolor{green_our}{\textbf{42.4}} &
	\textcolor{green_our}{\textbf{3.9}}\\
	\midrule
	\midrule
	\multirow{5}{*}{\rotatebox{90}{\textit{plane}}} &
	ONet~\cite{mescheder2019occupancy} &
    \textcolor{blue_our}{48.1} & \textcolor{blue_our}{2.8} &
	\textcolor{blue_our}{48.1} & \textcolor{blue_our}{2.8} &
	27.7 & 5.7 &
	12.7 & 13.0 &
	7.6 & 22.6 &
	24.1 & 7.2 &
	13.9 & 11.3 &
	12.4 & 13.0 &
	11.9 & 13.4 &
	19.6 & 7.1 &
	8.3 & 16.6 &
	27.1 & 7.0 &
	22.4 & 7.6 &
	19.9 & 11.6 &
	\textcolor{orange_our}{17.3} & \textcolor{orange_our}{11.3} &
	\textcolor{green_our}{16.5} & \textcolor{green_our}{10.0}\\
	& ONet-SDF~\cite{mescheder2019occupancy} &
    \textcolor{blue_our}{30.5} & \textcolor{blue_our}{4.7} &
	\textcolor{blue_our}{30.5} & \textcolor{blue_our}{4.7} &
	23.0 & 6.7 &
	11.4 & 12.5 &
	8.0 & 21.5 &
	19.4 & 8.1 &
	14.1 & 11.1 &
	11.0 & 13.1 &
	12.2 & 13.1 &
	16.1 & 8.0 &
	9.0 & 14.5 &
	22.4 & 7.6 &
	20.3 & 8.1 &
	19.0 & 11.2 &
	\textcolor{orange_our}{15.5} & \textcolor{orange_our}{11.3} &
	\textcolor{green_our}{13.8} & \textcolor{green_our}{10.4}\\
	& LDIF$_{svim1d}$~\cite{genova2020local} &
	\textcolor{blue_our}{\textbf{86.1}} & \textcolor{blue_our}{\textbf{0.2}} &
	\textcolor{blue_our}{\textbf{86.1}} & \textcolor{blue_our}{\textbf{0.2}} &
	33.1 & \textbf{1.8} &
	12.9 & 22.7 &
	10.0 & 50.4 &
	26.9 & 5.4 &
	10.4 & 28.5 &
	11.6 & 27.9 &
	8.3	 & 39.1 &
	30.6 & \textbf{1.8} &
	9.9	 & 47.3 &
	\textbf{41.1} & \textbf{3.4} &
	18.2 & 5.9 &
	18.7 & 16.1 &
	\textcolor{orange_our}{19.3} & \textcolor{orange_our}{20.9} &
	\textcolor{green_our}{15.9} & \textcolor{green_our}{16.7} \\
	& HPN (ours) &
	\textcolor{blue_our}{46.1} & \textcolor{blue_our}{3.1} &
	\textcolor{blue_our}{46.1} & \textcolor{blue_our}{3.1} &
	\textbf{45.3} & 4.7 &
	\textbf{35.0} & \textbf{6.2} &
	30.7 & 18.4 &
	\textbf{35.5} & \textbf{4.9} &
	39.8 & 7.2 &
	39.3 & 7.5 &
	35.4 & 7.9 &
	35.4 & 4.7 &
	36.0 & 8.5 &
	40.7 & 5.4 &
	\textbf{39.9} & \textbf{5.5} &
	\textbf{42.7} & 5.5 &
	\textcolor{orange_our}{38.0} & \textcolor{orange_our}{7.2} &
	\textcolor{green_our}{\textbf{30.1}} & \textcolor{green_our}{\textbf{5.9}}\\
	& HPN-SDF (ours) &
	\textcolor{blue_our}{39.0} & \textcolor{blue_our}{3.7} &
	\textcolor{blue_our}{39.0} & \textcolor{blue_our}{3.7} &
	43.7 & 5.1 &
	33.2 & \textbf{6.2} &
	\textbf{36.0} & \textbf{10.8} &
	31.7 & 5.3 &
	\textbf{46.4} & \textbf{6.1} &
	\textbf{43.5} & \textbf{6.3} &
	\textbf{42.0} & \textbf{6.8} &
	\textbf{36.9} & 4.8 &
	\textbf{47.7} & \textbf{5.9} &
	40.1 & 5.5 &
	39.7 & 5.7 &
	41.4 & \textbf{5.3} &
	\textcolor{orange_our}{\textbf{40.2}} & \textcolor{orange_our}{\textbf{6.1}} &
	\textcolor{green_our}{27.5} & \textcolor{green_our}{6.3}\\
	\midrule
	\multirow{5}{*}{\rotatebox{90}{\textit{chair}}} &
	ONet~\cite{mescheder2019occupancy} &
	\textcolor{blue_our}{36.2} & \textcolor{blue_our}{4.6} &
	15.9 & 9.1 &
	29.6 & 5.8 &
	\textcolor{blue_our}{36.2} & \textcolor{blue_our}{4.6} &
	16.6 & 10.3 &
	26.6 & 6.0 &
	37.6 & 5.7 &
	18.7 & 9.6 &
	34.2 & 6.5 &
	8.6 & 12.1 &
	17.7 & 9.1 &
	19.1 & 8.8 &
	35.3 & 5.4 &
	31.6 & 5.9 &
	\textcolor{orange_our}{24.3} & \textcolor{orange_our}{7.9} &
	\textcolor{green_our}{16.5} & \textcolor{green_our}{9.3}\\
	& ONet-SDF~\cite{mescheder2019occupancy} &
	\textcolor{blue_our}{37.9} & \textcolor{blue_our}{4.4} &
	19.2 & 7.6 &
	31.5 & 5.6 &
	\textcolor{blue_our}{37.9} & \textcolor{blue_our}{4.4} &
    21.5 & 8.3 &
	29.2 & 5.6 &
	39.2 & 5.5 &
	21.5 & 8.4 &
	37.4 & 6.0 &
	10.9 & 10.7 &
	21.6 & 7.5 &
	20.8 & 8.2 &
	37.3 & 5.0 &
	34.0 & 5.5 &
	\textcolor{orange_our}{27.0} & \textcolor{orange_our}{7.0} &
	\textcolor{green_our}{18.4} & \textcolor{green_our}{8.8} \\
	& LDIF$_{svim1d}$~\cite{genova2020local} &
	\textcolor{blue_our}{\textbf{59.2}} & \textcolor{blue_our}{\textbf{1.0}} &
	24.5 & 6.7 &
	31.9 & \textbf{1.8} &
	\textcolor{blue_our}{\textbf{59.2}} & \textcolor{blue_our}{\textbf{1.0}} &
    17.8 & 10.6 &
	\textbf{42.5} & \textbf{1.4} &
	28.5 & \textbf{3.7} &
	15.7 & 6.2 &
	21.6 & 5.6 &
	18.7 & 3.5 &
	23.7 & \textbf{3.4} &
	32.3 & \textbf{2.2} &
	\textbf{44.4} & \textbf{1.4} &
	31.4 & \textbf{3.9} &
	\textcolor{orange_our}{27.7} & \textcolor{orange_our}{\textbf{4.2}} &
	\textcolor{green_our}{14.9} & \textcolor{green_our}{13.0} \\
	& HPN (ours) &
	\textcolor{blue_our}{43.0} & \textcolor{blue_our}{3.9} &
	37.2 & 4.4 &
	\textbf{48.0} & 4.6 &
	\textcolor{blue_our}{43.0} & \textcolor{blue_our}{3.9} &
	40.2 & 4.6 &
	37.3 & 4.1 &
	51.3 & 4.7 &
	44.9 & 6.0 &
	48.6 & \textbf{4.8} &
	40.6 & 3.8 &
	38.6 & 5.5 &
	41.9 & 5.3 &
	\textbf{44.4} & 4.6 &
	\textbf{44.2} & 4.3 &
	\textcolor{orange_our}{43.1} &
	\textcolor{orange_our}{4.7} &
	\textcolor{green_our}{31.2} &
	\textcolor{green_our}{5.3}\\
	& HPN-SDF (ours) &
	\textcolor{blue_our}{41.2} & \textcolor{blue_our}{4.2} &
	\textbf{40.9} & \textbf{4.0} &
	47.6 & 5.0 &
	\textcolor{blue_our}{41.2} & \textcolor{blue_our}{4.2} &
    \textbf{43.6} & \textbf{4.3} &
	38.0 & 4.1 &
	\textbf{51.4} & 5.2 &
	\textbf{46.6} & \textbf{5.7} &
	\textbf{48.8} & 5.0 &
	\textbf{46.7} & \textbf{3.3} &
	\textbf{47.5} & 4.7 &
	\textbf{44.7} & 5.3 &
	43.8 & 5.0 &
	\textbf{44.2} & 4.5 &
	\textcolor{orange_our}{\textbf{45.3}} & \textcolor{orange_our}{4.7} &
	\textcolor{green_our}{\textbf{31.7}} & \textcolor{green_our}{\textbf{5.2}} \\
	\midrule
	\multirow{5}{*}{\rotatebox{90}{\textit{lamp}}} &
	ONet~\cite{mescheder2019occupancy} &
	\textcolor{blue_our}{42.0} & \textcolor{blue_our}{4.7} &
	20.8 & 7.7 &
	26.8 & 6.4 &
	20.4 & 8.1 &
	\textcolor{blue_our}{42.0} & \textcolor{blue_our}{4.7} &
	23.7 & 7.1 &
	35.1 & 5.5 &
	24.9 & 7.0 &
	37.8 & 5.6 &
	21.0 & 6.7 &
	30.8 & 6.8 &
	27.2 & 6.4 &
	24.2 & 7.2 &
	29.1 & 7.1 &
	\textcolor{orange_our}{26.8} & \textcolor{orange_our}{6.8} &
	\textcolor{green_our}{18.1} & \textcolor{green_our}{8.5}\\
	& ONet-SDF~\cite{mescheder2019occupancy} &
	\textcolor{blue_our}{31.6} & \textcolor{blue_our}{5.5} &
	17.2 & 8.1 &
	24.0 & 6.9 &
	18.3 & 8.3 &
    \textcolor{blue_our}{31.6} & \textcolor{blue_our}{5.5} &
	18.5 & 8.3 &
	34.5 & 5.6 &
	23.5 & 7.3 &
	34.3 & 6.0 &
	17.4 & 7.3 &
	28.3 & 6.9 &
	21.9 & 7.6 &
	21.5 & 7.9 &
	23.4 & 7.9 &
	\textcolor{orange_our}{23.6} & \textcolor{orange_our}{7.3} &
	\textcolor{green_our}{16.1} & \textcolor{green_our}{8.9} \\
	& LDIF$_{svim1d}$~\cite{genova2020local} & 
	\textcolor{blue_our}{48.1} & \textcolor{blue_our}{\textbf{2.5}} &
	18.1 & 5.4 &
	22.4 & \textbf{2.4} &
	12.4 & 12.2 &
	\textcolor{blue_our}{48.1} & \textcolor{blue_our}{\textbf{2.5}} &
	14.6 & 7.1 &
	23.4 & \textbf{3.4} &
	14.1 & 6.8 &
	21.6 & 5.1 &
	\textbf{48.6} & \textbf{1.3} &
	15.5 & \textbf{3.6} &
	34.1 & \textbf{2.0} &
	11.8 & 7.4 &
	17.1 & 10.5 &
	\textcolor{orange_our}{21.1} & \textcolor{orange_our}{5.6} &
	\textcolor{green_our}{12.5} & \textcolor{green_our}{14.0}\\
	& HPN (ours) &
	\textcolor{blue_our}{\textbf{50.3}} & \textcolor{blue_our}{3.6} &
	\textbf{43.0} & \textbf{4.2} &
	\textbf{46.5} & 5.1 &
	\textbf{42.4} & \textbf{4.7} &
	\textcolor{blue_our}{\textbf{50.3}} & \textcolor{blue_our}{3.6} &
	\textbf{41.0} & \textbf{4.3} &
	\textbf{54.5} & 4.5 &
	48.6 & 5.4 &
	\textbf{53.2} & \textbf{4.6} &
	43.8 & 4.0 &
	54.0 & 5.4 &
	\textbf{45.7} & 5.0 &
	\textbf{45.2} & \textbf{5.0} &
	\textbf{47.1} & \textbf{4.7} &
	\textcolor{orange_our}{\textbf{47.1}} &
	\textcolor{orange_our}{\textbf{4.7}} &
	\textcolor{green_our}{\textbf{35.8}} &
	\textcolor{green_our}{\textbf{5.0}}\\
	& HPN-SDF (ours) &
	\textcolor{blue_our}{48.4} & \textcolor{blue_our}{3.6} &
	41.6 & \textbf{4.2} &
	44.5 & 5.1 &
	41.1 & 4.8 &
    \textcolor{blue_our}{48.4} & \textcolor{blue_our}{3.6} &
	38.3 & 4.5 &
	53.2 & 4.4 &
	\textbf{49.8} & \textbf{5.2} &
	51.5 & \textbf{4.6} &
	43.0 & 4.0 &
	\textbf{56.7} & 5.4 &
	44.4 & 5.1 &
	44.7 & \textbf{5.0} &
	44.8 & 4.8 &
	\textcolor{orange_our}{46.1} & \textcolor{orange_our}{4.8} &
	\textcolor{green_our}{33.9} & \textcolor{green_our}{5.2} \\
	\bottomrule
	\end{tabular}
}
\vspace{1mm}
\caption{Comparison of the hierarchical prior network (HPN) to the state of the art in terms of generalization. The top part of the table shows training in the \textit{multi-class} setting, the lower part shows training on a single class. We report two metrics: F-score (F, shown in \%) and \colorbox{lightgray}{Chamfer distance} (CD, multiplied by 100 for better readability). * denotes results taken from the original paper. \textcolor{blue_our}{Results on categories seen during training are marked in blue.}  \textcolor{orange_our}{\emph{Mean (unseen)} shows the average of per-class scores over unseen categories.} \textcolor{green_our}{\emph{Composition} shows results on the composition of two objects per image.} On \textcolor{green_our}{compositions}, HPN is more than twice as accurate as the state of the art and generally better on \textcolor{orange_our}{unseen
classes}, while LDIF is better on \textcolor{blue_our}{seen classes}. Best viewed in color.}
\label{tbl:apx_results}
\vspace{3mm}
\end{table}

\end{landscape}

%% file: X_quantitative_iou.tex
\begin{landscape}

\begin{table}[htb!]
\centering
\resizebox*{\hsize}{!}{
	\setlength{\tabcolsep}{2mm}
	\begin{tabular}{l c
	c|| 
	>{\columncolor{lightgray}}c 
	c 
	>{\columncolor{lightgray}}c 
	c 
	>{\columncolor{lightgray}}c 
	c 
	>{\columncolor{lightgray}}c 
	c 
	>{\columncolor{lightgray}}c 
	c 
	>{\columncolor{lightgray}}c 
	c 
	>{\columncolor{lightgray}}c 
	||c 
	||c} 
	\toprule
	& &	
	Mean (seen) &
	Airplane &
	Car &
	Chair &
	Lamp &
	Bench &
	Cabinet &
	Display &
	Speaker &
	Rifle &
	Phone &
	Vessel &
	Sofa &
	Table &
	Mean (unseen) &
	Composition\\
	\midrule
	\multirow{6}{*}{\rotatebox{90}{\textit{plane,car,chair}}} &
	ONet~\cite{mescheder2019occupancy} &
	\textcolor{blue_our}{77.5} & 
	\textcolor{blue_our}{77.2} & 
	\textcolor{blue_our}{\textbf{83.6}} & 
	\textcolor{blue_our}{71.7} & 
	51.6 &
	60.3 & 
	73.3 & 
	50.8 & 
	68.5 & 
	62.4 & 
	58.5 & 
	65.7 & 
	74.4 &
	57.9 &
	\textcolor{orange_our}{62.3} &
	\textcolor{green_our}{46.1}\\
	& ONet-SDF~\cite{mescheder2019occupancy} &
	\textcolor{blue_our}{75.1} & 
	\textcolor{blue_our}{74.5} & 
	\textcolor{blue_our}{81.2} & 
	\textcolor{blue_our}{69.8} & 
	55.5 &
	60.3 & 
	73.0 & 
	53.8 & 
	69.6 & 
	65.0 & 
	62.8 & 
	66.2 & 
	73.6 &
	56.6 &
	\textcolor{orange_our}{63.6} &
	\textcolor{green_our}{49.4}\\
	& LDIF$_{svim1d}$~\cite{genova2020local} &	
	\textcolor{blue_our}{72.6} & 
	\textcolor{blue_our}{74.6} & 
	\textcolor{blue_our}{82.3} & 
	\textcolor{blue_our}{60.8} & 
	18.6 &
	32.1 & 
	49.5 & 
	12.8 & 
	44.0 & 
	28.9 & 
	19.7 & 
	50.2 & 
	66.5 &
	27.4 &
	\textcolor{orange_our}{35.0} &
	\textcolor{green_our}{16.2}\\
	\cmidrule{2-18} 
	& HPN (ours) &
	\textcolor{blue_our}{\textbf{79.9}} & 
	\textcolor{blue_our}{\textbf{80.6}} & 
	\textcolor{blue_our}{82.9} & 
	\textcolor{blue_our}{\textbf{76.1}} & 
	\textbf{72.1} &
	\textbf{70.6} & 
	\textbf{77.7} & 
	63.9 & 
	\textbf{75.6} & 
	80.0 & 
	73.5 & 
	\textbf{72.2} & 
	\textbf{77.5} &
	\textbf{67.8} &
	\textcolor{orange_our}{\textbf{73.1}} &
	\textcolor{green_our}{64.2}\\
	& HPN-SDF (ours) &
	\textcolor{blue_our}{76.5} & 
	\textcolor{blue_our}{78.9} & 
	\textcolor{blue_our}{78.1} & 
	\textcolor{blue_our}{72.5} & 
	65.3 &
	70.0 & 
	75.2 & 
	\textbf{64.9} & 
	74.8 & 
	\textbf{82.6} & 
	\textbf{74.7} & 
	71.2 & 
	72.5 &
	66.5 &
	\textcolor{orange_our}{71.8} &
	\textcolor{green_our}{70.1}\\
	\midrule
	\midrule
	\multirow{5}{*}{\rotatebox{90}{\textit{plane}}} &
	ONet~\cite{mescheder2019occupancy} &
	\textcolor{blue_our}{80.0} & 
	\textcolor{blue_our}{80.0} & 
	66.1 & 
	28.3 & 
	20.1 &
	36.4 & 
	50.7 & 
	36.6 & 
	42.7 & 
	62.6 & 
	33.4 & 
	59.3 & 
	53.9 &
	23.4 &
	\textcolor{orange_our}{42.8} &
	\textcolor{green_our}{35.0}\\
	& ONet-SDF~\cite{mescheder2019occupancy} &
	\textcolor{blue_our}{75.4} & 
	\textcolor{blue_our}{75.4} & 
	67.3 & 
	34.7 & 
	24.2 &
	43.8 & 
	51.3 & 
	37.7 & 
	44.3 & 
	65.3 & 
	37.4 & 
	61.6 & 
	58.8 &
	30.1 &
	\textcolor{orange_our}{46.4} &
	\textcolor{green_our}{40.9}\\
	& LDIF$_{svim1d}$~\cite{genova2020local} &
	\textcolor{blue_our}{73.4} & 
	\textcolor{blue_our}{73.4} & 
	51.6 & 
	12.6 & 
	6.4 &
	11.6 & 
	18.3 & 
	8.4 & 
	15.4 & 
	23.5 & 
	6.7 & 
	42.7 & 
	21.4 &
	11.6 &
	\textcolor{orange_our}{19.2} &
	\textcolor{green_our}{14.3}\\
	& HPN (ours) &
	\textcolor{blue_our}{\textbf{82.9}} & 
	\textcolor{blue_our}{\textbf{82.9}} & 
	71.7 & 
	55.9 & 
	42.6 &
	61.2 & 
	62.0 & 
	54.4 & 
	57.3 & 
	77.8 & 
	52.5 & 
	67.6 & 
	67.2 &
	51.8 &
	\textcolor{orange_our}{60.2} &
	\textcolor{green_our}{59.2}\\
	& HPN-SDF (ours) &
	\textcolor{blue_our}{81.4} & 
	\textcolor{blue_our}{81.4} & 
	\textbf{71.8} & 
	\textbf{58.8} & 
	\textbf{53.1} &
	\textbf{63.7} & 
	\textbf{64.8} & 
	\textbf{60.3} & 
	\textbf{61.8} & 
	\textbf{80.4} & 
	\textbf{64.6} & 
	\textbf{69.6} & 
	\textbf{68.6} &
	\textbf{58.8} &
	\textcolor{orange_our}{\textbf{64.7}} &
	\textcolor{green_our}{\textbf{59.7}}\\
	\midrule
	\multirow{5}{*}{\rotatebox{90}{\textit{chair}}} &
	ONet~\cite{mescheder2019occupancy} &
	\textcolor{blue_our}{69.0} & 
	46.0 & 
	68.8 & 
	\textcolor{blue_our}{69.0} & 
	49.2 &
	56.9 & 
	69.9 & 
	45.9 & 
	66.8 & 
	46.6 & 
	54.2 & 
	54.8 & 
	71.5 &
	54.8 &
	\textcolor{orange_our}{57.1} &
	\textcolor{green_our}{42.5}\\
	& ONet-SDF~\cite{mescheder2019occupancy} &
	\textcolor{blue_our}{70.5} & 
	49.6 & 
	70.5 & 
	\textcolor{blue_our}{70.5} & 
	51.4 &
	58.6 & 
	70.1 & 
	48.0 & 
	67.8 & 
	44.4 & 
	54.0 & 
	55.9 & 
	72.9 &
	55.0 &
	\textcolor{orange_our}{58.2} &
	\textcolor{green_our}{45.5}\\
	& LDIF$_{svim1d}$~\cite{genova2020local} &
	\textcolor{blue_our}{59.5} & 
	15.3 & 
	49.2 & 
	\textcolor{blue_our}{59.5} & 
	13.3 &
	28.0 & 
	46.7 & 
	13.7 & 
	39.6 & 
	9.1 & 
	13.3 & 
	28.2 & 
	63.7 &
	24.7 &
	\textcolor{orange_our}{28.7} &
	\textcolor{green_our}{12.9}\\
	& HPN (ours) &
	\textcolor{blue_our}{\textbf{76.3}} & 
	72.3 & 
	\textbf{76.2} & 
	\textcolor{blue_our}{\textbf{76.3}} & 
	72.6 &
	\textbf{72.3} & 
	\textbf{75.8} & 
	65.1 & 
	\textbf{75.6} & 
	82.5 & 
	73.2 & 
	\textbf{71.1} & 
	\textbf{77.2} &
	\textbf{68.8} &
	\textcolor{orange_our}{\textbf{73.6}} &
	\textcolor{green_our}{66.0}\\
	& HPN-SDF (ours) &
	\textcolor{blue_our}{74.4} & 
	\textbf{74.3} & 
	74.6 & 
	\textcolor{blue_our}{74.4} & 
	\textbf{72.7} &
	72.2 & 
	73.6 & 
	\textbf{66.7} & 
	74.4 & 
	\textbf{84.0} & 
	\textbf{75.8} & 
	70.5 & 
	75.0 &
	67.6 &
	\textcolor{orange_our}{73.4} &
	\textcolor{green_our}{\textbf{67.2}}\\
	\midrule
	\multirow{5}{*}{\rotatebox{90}{\textit{lamp}}} &
	ONet~\cite{mescheder2019occupancy} &
	\textcolor{blue_our}{69.8} & 
	44.2 & 
	64.0 & 
	40.6 & 
    \textcolor{blue_our}{69.8} &
	39.9 & 
	69.0 & 
	53.6 & 
	67.8 & 
	64.4 & 
	66.2 & 
	58.5 & 
	54.8 &
	43.9 &
	\textcolor{orange_our}{55.6} &
	\textcolor{green_our}{38.0}\\
	& ONet-SDF~\cite{mescheder2019occupancy} &
	\textcolor{blue_our}{70.5} & 
	48.8 & 
	66.1 & 
	44.3 & 
    \textcolor{blue_our}{70.5} &
	44.4 & 
	70.8 & 
	54.8 & 
	69.4 & 
	67.8 & 
	67.1 & 
	60.8 & 
	57.9 &
	44.9 &
	\textcolor{orange_our}{58.1} &
	\textcolor{green_our}{43.2}\\
	& LDIF$_{svim1d}$~\cite{genova2020local} & 
	\textcolor{blue_our}{41.8} & 
	9.9 & 
	42.6 & 
	9.6 & 
    \textcolor{blue_our}{41.8} &
	7.3 & 
	45.6 & 
	25.3 & 
	48.8 & 
	37.0 & 
	35.0 & 
	38.0 & 
	10.3 &
	13.2 &
	\textcolor{orange_our}{26.9} &
	\textcolor{green_our}{10.3}\\
	& HPN (ours) &
	\textcolor{blue_our}{78.0} & 
	68.3 & 
	72.1 & 
	\textbf{65.0} & 
    \textcolor{blue_our}{78.0} &
	64.4 & 
	74.8 & 
	65.1 & 
	74.3 & 
	80.5 & 
	\textbf{72.1} & 
	68.8 & 
	68.4 &
	\textbf{62.4} &
	\textcolor{orange_our}{69.7} &
	\textcolor{green_our}{\textbf{64.1}}\\
	& HPN-SDF (ours) &
	\textcolor{blue_our}{\textbf{78.2}} & 
	\textbf{69.6} & 
	\textbf{72.2} & 
	64.9 & 
    \textcolor{blue_our}{\textbf{78.2}} &
	\textbf{65.7} & 
	\textbf{75.7} & 
	\textbf{65.5} & 
	\textbf{75.0} & 
	\textbf{81.4} & 
	70.5 & 
	\textbf{68.9} & 
	\textbf{69.3} &
	\textbf{62.4} &
	\textcolor{orange_our}{\textbf{70.1}} &
	\textcolor{green_our}{63.5}\\
	\bottomrule
	\end{tabular}
}
\vspace{1mm}
\caption{Comparison of the hierarchical prior network (HPN) to the state of the art in terms of generalization. The top part of the table shows training in the \textit{multi-class} setting, the lower part shows training on a single class. This table reports the intersection over union (IoU) values in \%. }
\label{tbl:apx_results_iou}
\vspace{3mm}
\end{table}

\end{landscape}

%% file: cvpr.bbl
\begin{thebibliography}{10}\itemsep=-1pt

\bibitem{Badki20meshlet}
Abhishek Badki, Orazio Gallo, Jan Kautz, and Pradeep Sen.
\newblock Meshlet priors for 3d mesh reconstruction.
\newblock In {\em CVPR}, 2020.

\bibitem{barrow1977chamfer}
Harry~G. Barrow, Jay~M. Tenenbaum, Robert~C. Bolles, and Helen~C. Wolf.
\newblock Parametric correspondence and chamfer matching: Two new techniques
  for image matching.
\newblock In {\em IJCAI}, 1977.

\bibitem{bautista2020generalization}
Miguel~{\'{A}}ngel Bautista, Walter Talbott, Shuangfei Zhai, Nitish Srivastava,
  and Joshua~M. Susskind.
\newblock On the generalization of learning-based 3d reconstruction.
\newblock {\em arXiv preprint arXiv:2006.15427}, 2020.

\bibitem{Chabra20deeplocalsdf}
Rohan Chabra, Jan~Eric Lenssen, Eddy Ilg, Tanner Schmidt, Julian Straub, Steven
  Lovegrove, and Richard~A. Newcombe.
\newblock Deep local shapes: Learning local {SDF} priors for detailed 3d
  reconstruction.
\newblock In {\em ECCV}, 2020.

\bibitem{chang15shapenet}
Angel~X. Chang, Thomas~A. Funkhouser, Leonidas~J. Guibas, Pat Hanrahan,
  Qi{-}Xing Huang, Zimo Li, Silvio Savarese, Manolis Savva, Shuran Song, Hao
  Su, Jianxiong Xiao, Li Yi, and Fisher Yu.
\newblock {ShapeNet}: An information-rich 3{D} model repository.
\newblock {\em arXiv:1512.03012}, 2015.

\bibitem{chibane2020implicit}
Julian Chibane, Thiemo Alldieck, and Gerard Pons{-}Moll.
\newblock Implicit functions in feature space for 3d shape reconstruction and
  completion.
\newblock In {\em CVPR}, 2020.

\bibitem{choy20163d}
Christopher~B. Choy, Danfei Xu, JunYoung Gwak, Kevin Chen, and Silvio Savarese.
\newblock 3d-r2n2: A unified approach for single and multi-view 3d object
  reconstruction.
\newblock In {\em ECCV}, 2016.

\bibitem{dai2017scannet}
Angela Dai, Angel~X. Chang, Manolis Savva, Maciej Halber, Thomas~A. Funkhouser,
  and Matthias Nie{\ss}ner.
\newblock Scannet: Richly-annotated 3d reconstructions of indoor scenes.
\newblock In {\em CVPR}, 2017.

\bibitem{deng2020cvxnet}
Boyang Deng, Kyle Genova, Soroosh Yazdani, Sofien Bouaziz, Geoffrey~E. Hinton,
  and Andrea Tagliasacchi.
\newblock Cvxnet: Learnable convex decomposition.
\newblock In {\em CVPR}, 2020.

\bibitem{fan17cvpr}
Haoqiang Fan, Hao Su, and Leonidas~J. Guibas.
\newblock A point set generation network for 3d object reconstruction from a
  single image.
\newblock In {\em CVPR}, 2017.

\bibitem{genova2020local}
Kyle Genova, Forrester Cole, Avneesh Sud, Aaron Sarna, and Thomas~A.
  Funkhouser.
\newblock Local deep implicit functions for 3d shape.
\newblock In {\em CVPR}, 2020.

\bibitem{groueix2018papier}
Thibault Groueix, Matthew Fisher, Vladimir~G. Kim, Bryan~C. Russell, and
  Mathieu Aubry.
\newblock A papier-m{\^a}ch{\'e} approach to learning 3d surface generation.
\newblock In {\em CVPR}, 2018.

\bibitem{Jiang20locimplgrid}
Chiyu~Max Jiang, Avneesh Sud, Ameesh Makadia, Jingwei Huang, Matthias
  Nie{\ss}ner, and Thomas~A. Funkhouser.
\newblock Local implicit grid representations for 3d scenes.
\newblock In {\em CVPR}, 2020.

\bibitem{li2019learning}
Li Jun, Niu Chengjie, and Xu Kai.
\newblock Learning part generation and assembly for structure-aware shape
  synthesis.
\newblock In {\em AAAI}, 2020.

\bibitem{kingma2014adam}
Diederik~P. Kingma and Jimmy Ba.
\newblock Adam: A method for stochastic optimization.
\newblock In {\em ICLR}, 2015.

\bibitem{Knapitsch2017fscore}
Arno Knapitsch, Jaesik Park, Qian-Yi Zhou, and Vladlen Koltun.
\newblock Tanks and temples: Benchmarking large-scale scene reconstruction.
\newblock {\em ACM Transactions on Graphics}, 36(4), 2017.

\bibitem{li17grass}
Jun Li, Kai Xu, Siddhartha Chaudhuri, Ersin Yumer, Hao~(Richard) Zhang, and
  Leonidas~J. Guibas.
\newblock {GRASS:} generative recursive autoencoders for shape structures.
\newblock {\em {ACM} Trans. Graph.}, 36(4):52:1--52:14, 2017.

\bibitem{lorensen1987marching}
William~E. Lorensen and Harvey~E. Cline.
\newblock Marching cubes: A high resolution 3d surface construction algorithm.
\newblock {\em ACM siggraph computer graphics}, 21(4):163--169, 1987.

\bibitem{mescheder2019occupancy}
Lars~M. Mescheder, Michael Oechsle, Michael Niemeyer, Sebastian Nowozin, and
  Andreas Geiger.
\newblock Occupancy networks: Learning 3d reconstruction in function space.
\newblock In {\em CVPR}, 2019.

\bibitem{mo2019partnet}
Kaichun Mo, Shilin Zhu, Angel~X. Chang, Li Yi, Subarna Tripathi, Leonidas~J.
  Guibas, and Hao Su.
\newblock Partnet: A large-scale benchmark for fine-grained and hierarchical
  part-level 3d object understanding.
\newblock In {\em CVPR}, 2019.

\bibitem{niu2018im2struct}
Chengjie Niu, Jun Li, and Kai Xu.
\newblock Im2struct: Recovering 3d shape structure from a single rgb image.
\newblock In {\em CVPR}, 2018.

\bibitem{paschalidou2020learning}
Despoina Paschalidou, Luc~Van Gool, and Andreas Geiger.
\newblock Learning unsupervised hierarchical part decomposition of 3d objects
  from a single rgb image.
\newblock In {\em CVPR}, 2020.

\bibitem{paschalidou2019superquadrics}
Despoina Paschalidou, Ali~Osman Ulusoy, and Andreas Geiger.
\newblock Superquadrics revisited: Learning 3d shape parsing beyond cuboids.
\newblock In {\em CVPR}, 2019.

\bibitem{NEURIPS2019_9015}
Adam Paszke, Sam Gross, Francisco Massa, Adam Lerer, James Bradbury, Gregory
  Chanan, Trevor Killeen, Zeming Lin, Natalia Gimelshein, Luca Antiga, Alban
  Desmaison, Andreas K{\"{o}}pf, Edward Yang, Zachary DeVito, Martin Raison,
  Alykhan Tejani, Sasank Chilamkurthy, Benoit Steiner, Lu Fang, Junjie Bai, and
  Soumith Chintala.
\newblock Pytorch: An imperative style, high-performance deep learning library.
\newblock In {\em NeurIPS}, 2019.

\bibitem{peng2020convolutional}
Songyou Peng, Michael Niemeyer, Lars~M. Mescheder, Marc Pollefeys, and Andreas
  Geiger.
\newblock Convolutional occupancy networks.
\newblock In {\em ECCV}, 2020.

\bibitem{ravi2020pytorch3d}
Nikhila Ravi, Jeremy Reizenstein, David Novotny, Taylor Gordon, Wan-Yen Lo,
  Justin Johnson, and Georgia Gkioxari.
\newblock Accelerating 3d deep learning with pytorch3d.
\newblock {\em arXiv:2007.08501}, 2020.

\bibitem{richter18cvpr}
Stephan~R. Richter and Stefan Roth.
\newblock Matryoshka networks: Predicting 3d geometry via nested shape layers.
\newblock In {\em CVPR}, 2018.

\bibitem{saito2019pifu}
Shunsuke Saito, Zeng Huang, Ryota Natsume, Shigeo Morishima, Hao Li, and Angjoo
  Kanazawa.
\newblock Pifu: Pixel-aligned implicit function for high-resolution clothed
  human digitization.
\newblock In {\em CVPR}, 2019.

\bibitem{shin2018pixels}
Daeyun Shin, Charless~C. Fowlkes, and Derek Hoiem.
\newblock Pixels, voxels, and views: A study of shape representations for
  single view 3d object shape prediction.
\newblock In {\em CVPR}, 2018.

\bibitem{tatarchenko2017octree}
Maxim Tatarchenko, Alexey Dosovitskiy, and Thomas Brox.
\newblock Octree generating networks: Efficient convolutional architectures for
  high-resolution 3d outputs.
\newblock In {\em ICCV}, 2017.

\bibitem{tatarchenko2019single}
Maxim Tatarchenko, Stephan~R. Richter, Ren{\'{e}} Ranftl, Zhuwen Li, Vladlen
  Koltun, and Thomas Brox.
\newblock What do single-view 3d reconstruction networks learn?
\newblock In {\em CVPR}, 2019.

\bibitem{thai20203dsdfnet}
Anh Thai, Stefan Stojanov, Vijay Upadhya, and James~M. Rehg.
\newblock 3d reconstruction of novel object shapes from single images.
\newblock {\em arXiv:2006.07752}, 2020.

\bibitem{abstractionTulsiani17}
Shubham Tulsiani, Hao Su, Leonidas~J. Guibas, Alexei~A. Efros, and Jitendra
  Malik.
\newblock Learning shape abstractions by assembling volumetric primitives.
\newblock In {\em CVPR}, 2017.

\bibitem{wang18eccv}
Nanyang Wang, Yinda Zhang, Zhuwen Li, Yanwei Fu, Wei Liu, and Yu{-}Gang Jiang.
\newblock Pixel2mesh: Generating 3d mesh models from single {RGB} images.
\newblock In {\em ECCV}, 2018.

\bibitem{shapehd}
Jiajun Wu, Chengkai Zhang, Xiuming Zhang, Zhoutong Zhang, William~T. Freeman,
  and Joshua~B. Tenenbaum.
\newblock {Learning 3D Shape Priors for Shape Completion and Reconstruction}.
\newblock In {\em ECCV}, 2018.

\bibitem{Wu_2020_CVPR}
Rundi Wu, Yixin Zhuang, Kai Xu, Hao Zhang, and Baoquan Chen.
\newblock Pq-net: A generative part seq2seq network for 3d shapes.
\newblock In {\em CVPR}, 2020.

\bibitem{xu2019disn}
Qiangeng Xu, Weiyue Wang, Duygu Ceylan, Radom{\'{\i}}r Mech, and Ulrich
  Neumann.
\newblock Disn: Deep implicit surface network for high-quality single-view 3d
  reconstruction.
\newblock In {\em NeurIPS}, 2019.

\bibitem{zhang2018learninggenre}
Xiuming Zhang, Zhoutong Zhang, Chengkai Zhang, Josh Tenenbaum, Bill Freeman,
  and Jiajun Wu.
\newblock Learning to reconstruct shapes from unseen classes.
\newblock In {\em NeurIPS}, 2018.

\bibitem{Zhou2018open3d}
Qian-Yi Zhou, Jaesik Park, and Vladlen Koltun.
\newblock {Open3D}: {A} modern library for {3D} data processing.
\newblock {\em arXiv:1801.09847}, 2018.

\bibitem{zou17iccv}
Chuhang Zou, Ersin Yumer, Jimei Yang, Duygu Ceylan, and Derek Hoiem.
\newblock 3d-prnn: Generating shape primitives with recurrent neural networks.
\newblock In {\em ICCV}, 2017.

\end{thebibliography}
